\documentclass[10pt,twocolumn,letterpaper]{article}

\usepackage{3dv}
\usepackage{times}
\usepackage{epsfig}
\usepackage{graphicx}
\usepackage{amsmath}
\usepackage{amssymb}

\usepackage{booktabs}
\usepackage{multirow}
\usepackage{tabularx}
\usepackage{paralist}
\usepackage{pifont}
\usepackage{placeins}

\usepackage[pagebackref=true,breaklinks=true,colorlinks,bookmarks=false]{hyperref}

\threedvfinalcopy 

\def\threedvPaperID{83} 

\begin{document}

\title{HybridSDF: Combining Deep Implicit Shapes and Geometric Primitives for~3D~Shape Representation and Manipulation}



\author{
	\textbf{Subeesh Vasu\thanks{Equal contributions\newline Subeesh Vasu: {\tt subeeshvasu@gmail.com}\newline Project page: \url{https://ntalabot.github.io/hybridsdf/}}$^{\phantom{*},1}$ \,\,\, Nicolas Talabot$^{*,1}$ \,\,\, Artem Lukoianov$^{1,2}$ \,\,\,
	Pierre Baqu\'e$^{2}$} \\ \textbf{Jonathan Donier$^{2}$ \,\,\, Pascal Fua$^{1}$} \\
	{\small $^{1}$CVLab, EPFL, {\tt\footnotesize \{firstname.lastname\}@epfl.ch}} \\
	{\small $^{2}$Neural Concept, {\tt\footnotesize \{firstname.lastname\}@neuralconcept.com}}
}

\maketitle


\newif\ifdraft
\draftfalse

\definecolor{orange}{rgb}{1,0.5,0}
\definecolor{green0}{rgb}{0.1,0.7,0.1}
\newcommand{\sparag}[1]{\vspace{-3mm}\subparagraph{#1}}
\newcommand{\parag}[1]{\vspace{-1mm}\paragraph{#1}}

\ifdraft
 \newcommand{\PF}[1]{{\color{red}{\bf PF: #1}}}
 \newcommand{\pf}[1]{{\color{red} #1}}
 \newcommand{\SV}[1]{{\color{blue}{\bf SV: #1}}}
 \newcommand{\sv}[1]{{\color{blue} #1}}
 \newcommand{\NT}[1]{{\color{green0}{\bf NT: #1}}}
 \newcommand{\nt}[1]{{\color{green0} #1}}
\else
 \newcommand{\PF}[1]{{\color{red}{}}}
 \newcommand{\pf}[1]{ #1 }
 \newcommand{\SV}[1]{{\color{blue}{}}}
 \newcommand{\sv}[1]{ #1 }
 \newcommand{\NT}[1]{{\color{green0}{}}}
 \newcommand{\nt}[1]{ #1 }
\fi

\newif\ifarxiv
\arxivfalse
\newcommand{\refappendix}[2]{\ifarxiv\ref{#1}\else#2\fi}

\renewcommand{\topfraction}{1}
\renewcommand{\dbltopfraction}{1}
\renewcommand{\bottomfraction}{1}
\renewcommand{\textfraction}{.0}
\renewcommand{\floatpagefraction}{1}
\renewcommand{\dblfloatpagefraction}{1}

\newcommand{\bL}[0]{\mathcal{L}}
\newcommand{\bd}[0]{\mathbf{d}}
\newcommand{\by}[0]{\mathbf{y}}
\newcommand{\hby}[0]{\hat{\mathbf{y}}}

\newcommand{\real}{\mathbb{R}}

\newcommand{\PP}{\mathrm{PP}}
\newcommand{\AP}{\mathrm{AP}}
\newcommand{\TP}{\mathrm{TP}}
\newcommand{\IoU}{\mathrm{IoU}}

\newcommand{\ha}[0]{\hat{a}}
\newcommand{\hb}[0]{\hat{b}}


\newcommand{\CCQ}{{\it CCQ}}
\newcommand{\TLTS}{{\it TLTS}}
\newcommand{\APLS}{{\it APLS}}
\newcommand{\HM}{{\it H\&M}}

\newcommand{\bp}[0]{\textbf{p}}
\newcommand{\mbp}[0]{\mathbf{p}}
\newcommand{\SDF}[0]{\mathbf{SDF}}
\newcommand{\LC}[0]{\mathbf{LC}}
\newcommand{\LV}[0]{\mathbf{LV}}

\newcommand{\lcp}{{$\mathbf{LC}_{\text{part}}$}}
\newcommand{\bR}[0]{\textbf{R}}
\newcommand{\bT}[0]{\textbf{T}}
\newcommand{\bS}[0]{\textbf{S}}

\newcommand{\DeepS}{{\it DeepSDF}}
\newcommand{\DeepSV}{{\it DeepSDF-V}}
\newcommand{\HS}{{\it HybridSDF}}
\newcommand{\HSV}{{\it HybridSDF-V}}
\newcommand{\HSs}{{\it HybridSDF$_{s}$}}
\newcommand{\HSd}{{\it HybridSDF$_{d}$}}
\newcommand{\HSVs}{{\it HybridSDF-V$_{s}$}}
\newcommand{\HSVd}{{\it HybridSDF-V$_{d}$}}
\newcommand{\HSb}{{\it HybridSDF$_{base}$}}
\newcommand{\DualS}{{\it DualSDF}}
\newcommand{\NP}{{\it NeuralParts}}
\newcommand{\HSQ}{{\it HierSQ}}
\arxivtrue


\begin{abstract}
	
Deep implicit surfaces excel at modeling generic shapes but do not always capture the regularities present in manufactured objects, which is something simple geometric primitives are particularly good at. In this paper, we propose a representation combining latent and explicit parameters that can be decoded into a set of deep implicit and geometric shapes that are consistent with each other. As a result, we can effectively model both complex and highly regular shapes that coexist in manufactured objects. This enables our approach to manipulate 3D shapes in an efficient and precise manner.



\end{abstract}



\section{Introduction}
\label{sec:intro}

Implicit surface representations have a long history~\cite{Sethian99} and have recently re-emerged in the form of signed distance functions~\cite{Park19c} or occupancy fields~\cite{Mescheder19, Chen19c} that can be implemented by deep networks mapping 3D points to implicit function values. These representations are lightweight, high-fidelity, unlimited in resolution, naturally able to handle topology changes and, consequently, extremely popular in both computer vision and computer graphics, for example to explore a shape space for design purposes. However, because they represent objects as arbitrary, or generic, surfaces, they can easily fail to capture the regularities that are prevalent in manufactured objects and that are best represented by simple geometric shapes. As a result, the geometry and topology of complex objects such as those of Fig.~\ref{fig:teaser} are not always faithfully preserved. On the other hand, approaches that rely purely on a collection of simple primitives often lack accuracy~\cite{Paschalidou20}.


\newlength{\teaserfigheight}
\setlength{\teaserfigheight}{2.4cm}

\begin{figure}[t]
  \centering
  \small  
\begin{tabular}{cc}
   \hspace{-3mm}\includegraphics[height=\teaserfigheight]{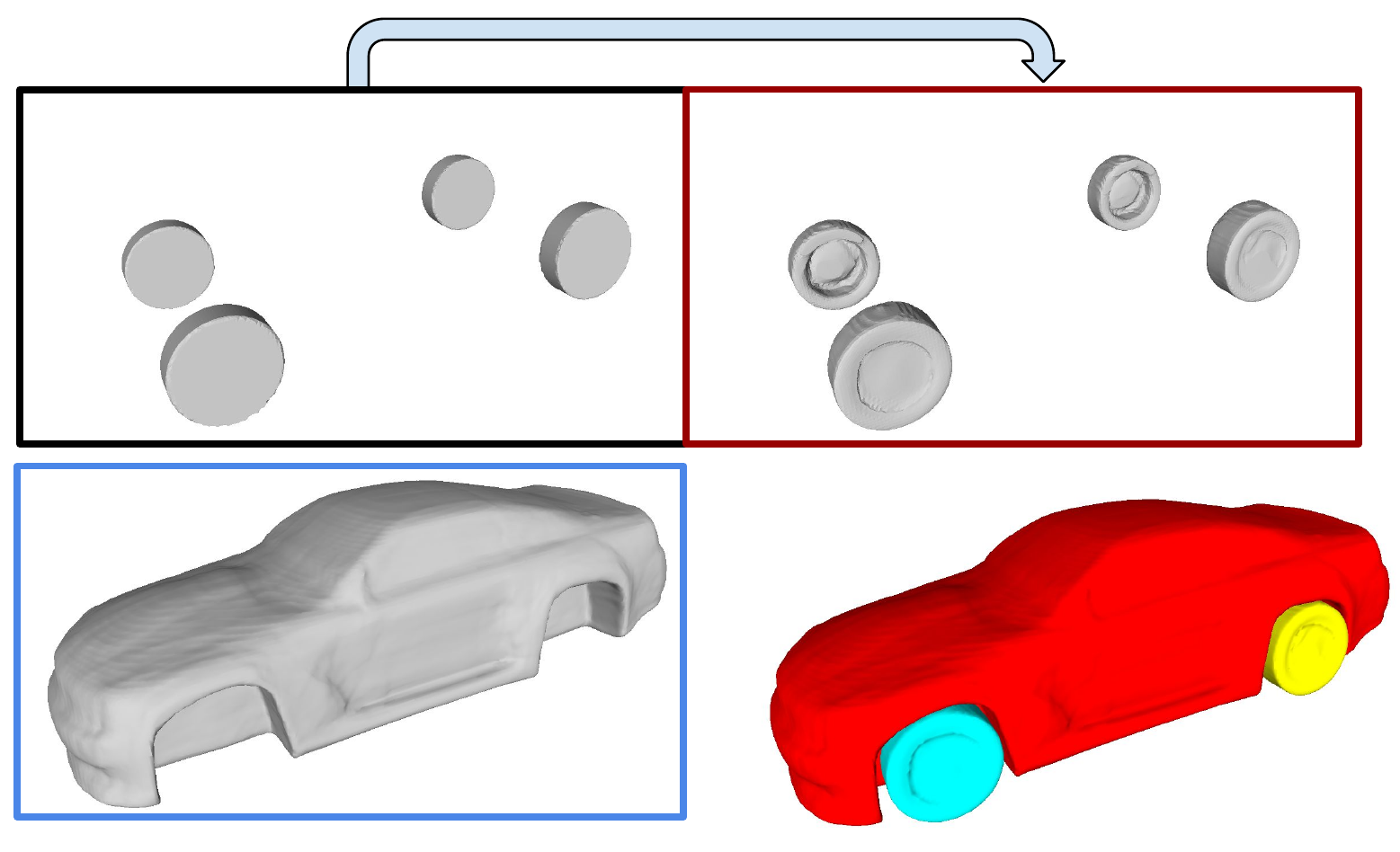}&\hspace{-3mm}\includegraphics[height=\teaserfigheight]{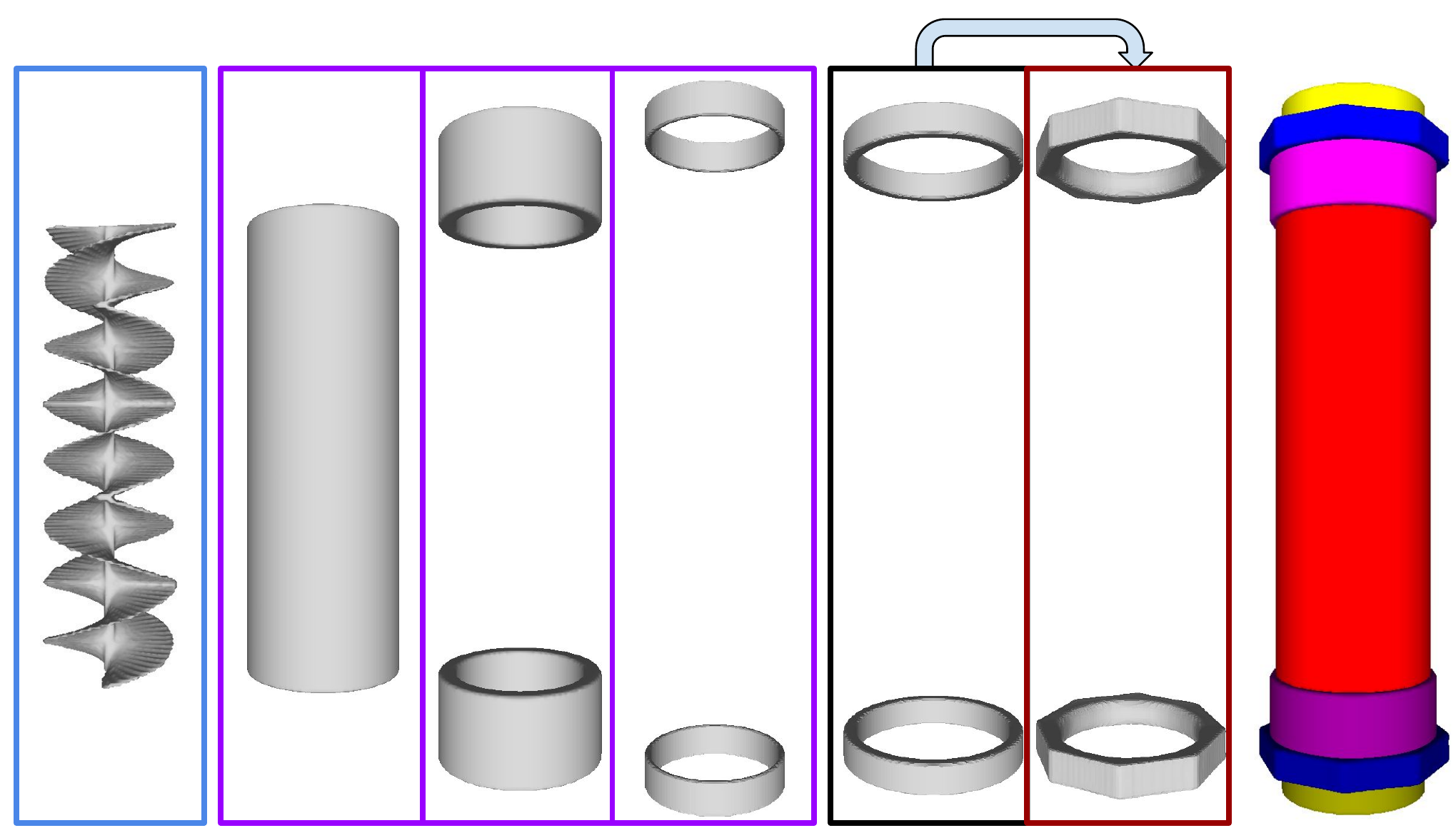}\\
  \hspace{-3mm}(a) & \hspace{-3mm}\hfill(b) \hfill (c) \hfill (d) \hfill (e) \hfill (f) \hfill (g) \hfill (h) \hfill
  \end{tabular}
  \caption{\textit{Modeling complex objects as a collection of primitives.} { (a) We depict cars as the combination of a main body and the wheels. The car body is represented by a {\it Generic-SDF} (red) whose shape can change significantly. By contrast, the wheels are modeled by {\it  Geometry-assisted-SDFs} (cyan and yellow) whose shapes are constrained not to differ excessively from that of the cylindrical {\it  Geometric-SDFs} shown in the upper left corner.  (b-h) This depicts a static mixer. It comprises a spiral helix (b) whose shape can also change significantly and is represented by a  {\it Generic-SDF}. It is encased in a tube (c, red in h) with screws at both ends. The center tube along with the tube-like parts associated with the screw (d, e, yellow and violet in h) are represented by {\it Geometric-SDFs} of hollow cylinders. The screws also contains deformable rings modeled by {\it  Geometry-assisted-SDFs} { (g, blue in h)} constrained not to deviate much from hollow cylinders {(f)}.}}
  \label{fig:teaser}
\end{figure}

In this work, we propose a novel implicit representation combining geometric primitives and generic surfaces to enforce local regularities, consistency, and which enables parametric manipulation. To this end, we use three types of primitives defined as follow:
%
\begin{enumerate}

\item  Generic primitives that may have arbitrarily complex shapes.

\item  Simple geometric primitives, such as spheres and cylinders. 

\item  Primitives that bear a close resemblance to geometric ones but can deviate from them, such as car wheels that are almost but not quite cylinders. 

\end{enumerate}
These three kinds of primitives are represented implicitly through their Signed Distance Functions (SDF) that we refer to as {\it Generic-SDF}, {\it Geometric-SDF}, and {\it Geometry-assisted-SDF} respectively. Fig.~\ref{fig:teaser} depicts their respective roles in our data. They are parameterized in terms of implicit latent vectors and explicit geometric parameters, such as the radius and width of wheels, which are decoded together into 3D shapes.

This makes it possible to represent object parts that are true geometric primitives as such, parts that are similar but not identical by assisted primitives, and the remaining ones as generic surfaces that can assume any shape. Hence, our 3D models are interpretable in terms of parts that have a semantic meaning, unlike those of~\cite{Hao20, Paschalidou21}, at minimal loss of accuracy. In short, our contribution is a novel SDF-based model that effectively combines the different kinds of primitives that are used for Computer Aided Design (CAD) objects to produce interpretable shapes that retain their internal consistency when edited. This opens up new avenues for user-friendly 3D shape manipulation and design with stronger surface regularities, unlike with purely deep implicit shapes, and automatic parameter detection from images and sketches, as we explore below.



\section{Related work}
\label{sec:related}


Manufactured objects are often represented as CAD models made of geometric primitives. In recent years, the emergence of deep-learning has fostered a strong interest in developing more sophisticated primitives along new ways to represent complex surfaces and to combine these two kinds of approaches. We review them briefly below. 

\subsection{Primitive-based representations}

Existing primitive-based methods for 3D shape representation aim to derive interesting abstractions by decomposing the 3D shapes into semantically meaningful simpler parts. In addition to simple primitives, more sophisticated ones such as cuboids~\cite{Tulsiani17b, Zou17a, Niu18, Sun19b, Kluger21, Smirnov20}, superquadrics~\cite{Paschalidou19, Paschalidou20}, anisotropic Gaussians~\cite{Genova19}, or convexes~\cite{Deng20c} have been proposed. All these works rely on a single primitive type to represent the entire shape. Hence, for complex shapes, reconstruction accuracy directly depends on the number of primitives used. To improve on this, the approach of~\cite{Paschalidou21} defines a bijective function between traditional spherical primitives and a corresponding set of deformed shapes that can be arbitrarily complex.

As far as shape representation and shape manipulation are concerned, these deep learning-based methods have many limitations. They are typically trained in an unsupervised way to yield self-discovered arrangements of parts that do not necessarily correspond to semantically meaningful ones. Typically, the more primitives are used the more severe the problem becomes. For example, in~\cite{Paschalidou21}, the latent space is parameterized by sphere parameters that have no direct relationship to the geometric parameters of the corresponding parts. Thus, there is no correspondence between the latent dimensions and true geometric parameters, such as location, orientation, or size, which is something we provide and can modify explicitly. This is partially addressed by works such as~\cite{Cascaval21}  that use a program composed of geometric operations to represent complex objects. However, such models are extremely hard to fit to data. Other works such as~\cite{Le21a} aim to fit geometric surface patches to high resolution point clouds, but eventually answer a different problem as they do not learn a general shape representation that can be explored.


\subsection{Explicit and implicit surfaces}

Among existing 3D surface representations, meshes made of vertices and faces are one of the most popular and versatile types. There are excellent meshing algorithms~\cite{Peters08}  that can refine a mesh so that it accurately represents a known 3D shape, but there are far fewer that can produce a detailed 3D mesh of an initially unknown 3D shape of arbitrary topology. There have been many early attempts using only 3D meshes~\cite{McInerney99}, but they require {\it ad hoc} heuristics that do not generalize well. 

An alternative to surface meshes is to use an implicit descriptor where the surface is defined by the zero-crossing of a volumetric function $\Psi: \mathbb{R}^3 \rightarrow \mathbb{R}$ that may, for example, evolve over time~\cite{Sethian99}. The strength of this implicit representation is that the zero-crossing surface can change topology without explicit re-parameterization. Until recently, its main drawback was thought to be that working with volumes, instead of surfaces, massively increased the computational burden. This changed dramatically with the introduction of continuous deep implicit-fields. They represent 3D shapes as level sets of deep networks that map 3D coordinates to a signed distance function~\cite{Park19c} or an occupancy field~\cite{Mescheder19, Chen19c}. This mapping yields a continuous shape representation that is lightweight but not limited in resolution. This representation has been successfully used for single-view reconstruction~\cite{Mescheder19,Chen19c,Xu19b} and 3D shape-completion~\cite{Chibane20a}. It has also been shown~\cite{Remelli20b,Guillard22a} that 3D meshes could be extracted from these implicit representations without compromising differentiability. This makes it possible to back-propagate through the meshing of the SDF, such as when optimizing a 3D shape so that its projection match a specific contour~\cite{Guillard21} or to maximize its aerodynamic performance~\cite{Baque18}. Recent works~\cite{Peng20c,Ibing21,Liu21b} have further extended the scope of implicit representations by developing network models that can yield fine-grained shape reconstructions.


\subsection{Hybrid representations} 

Hybrid methods attempt to combine different representations to add functionalities. A common trend is to use a shared latent space across different representations such as voxel-based, image-based, or point-based representations to enhance performance in discriminative tasks as classification and segmentation~\cite{Hegde16, Chen19c, Muralikrishnan19}. However, the \DualS{} approach of~\cite{Hao20} is the only one we know of that relies on an hybrid representation for shape generation and manipulation. It builds a shared latent space for two distinct shape representations: a sphere-based approximation and an implicit function-based high-fidelity reconstruction. The shared latent space couples the two representations and, as a result, the fine-grained model can be manipulated by changing the parameters of the spheres. However as the method of~\cite{Paschalidou21}, \DualS{} suffers from the limited interpretability of the spherical primitives that do not correspond to semantically meaningful parts and their geometric attributes, which is a key difference with our approach.

\section{Approach}
\label{sec:approach}

\begin{figure*}[t]
  \centering
  \small  

\includegraphics[width=0.9\textwidth, trim={0cm 9.5cm 0cm 0cm}, clip]{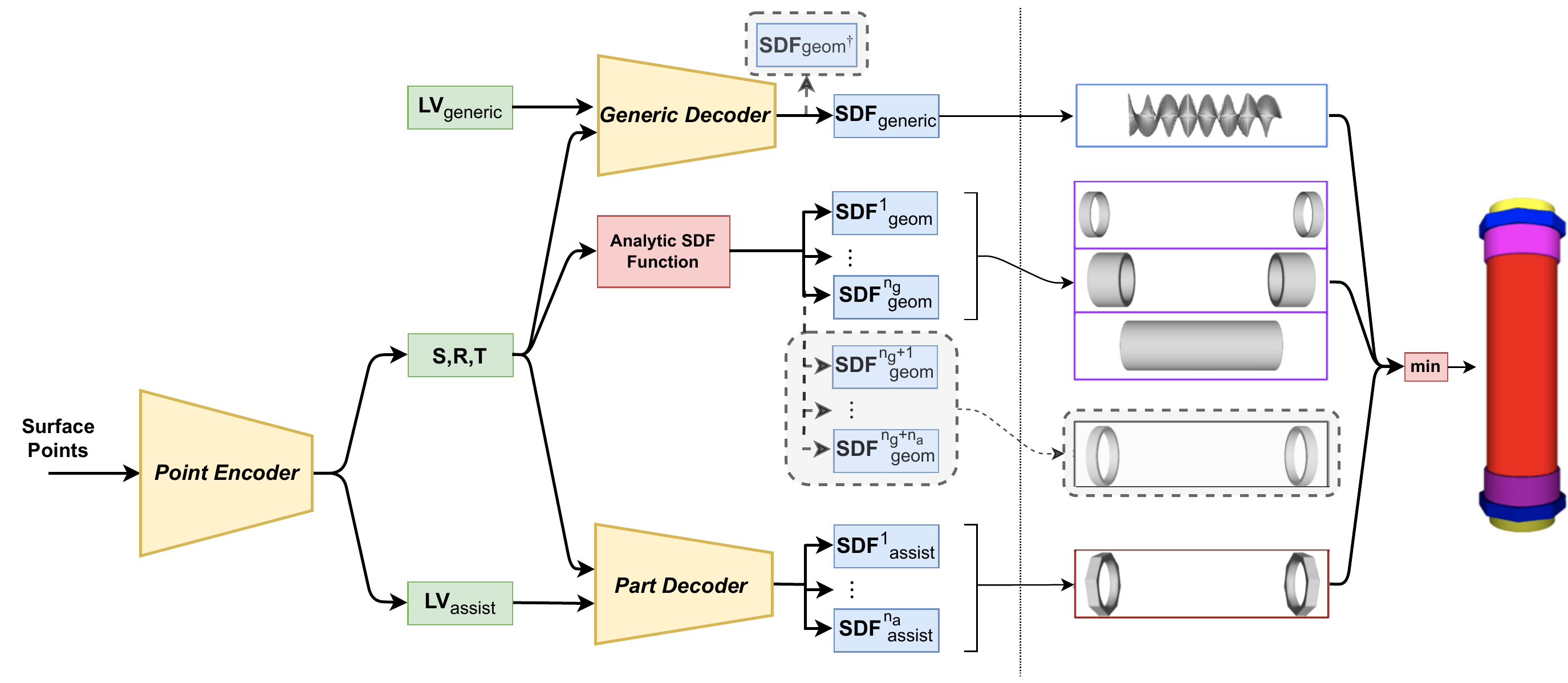}
  \caption{\textit{\HS{} architecture.} ({\it Left side}) The decoding part of the model takes as input a latent vector $\LV_{\text{generic}}$, geometric parameters $\bS$, $\bR$, and $\bT$, along with an auxiliary latent vector $\LV_{\text{assist}}$. They are respectively decoded into the three types of primitives we have defined. A \textit{Point Encoder} predicts $\bS$, $\bR$, $\bT$, and  $\LV_{\text{assist}}$ using 3D surface point clouds of the full shape. Grey boxes denote components used only for training purposes. The query point $\bp$ is concatenated to the inputs of the decoders and analytic SDFs, but is not shown for clarity. ({\it Right side}) Example reconstruction for the mixer of Fig.~\ref{fig:teaser}. The blue, purple, and brown boxes contain the resulting surfaces: the generic primitive in blue, the geometric primitives in purple, and the geometry-assisted primitives in brown. Those in the black box denote the geometric shapes used to constrain the geometry-assisted primitives during training and are not used to reconstruct the final shape.}
  \label{fig:arch}
\end{figure*}

We first introduce our hybrid implicit representation, then our proposed model that decodes latent vectors and geometric parameters into complex 3D shapes, as illustrated in Fig.~\ref{fig:arch}, and eventually the training losses.


\subsection{HybridSDF representation}
 
Signed Distance Functions (SDF) are mappings that associate to 3D points \bp~$\in \mathbb{R}^{3}$ a value in $\mathbb{R}$ that specifies their distance to a closed surface. By convention, the value is negative if \bp{} is inside the surface and positive otherwise. Hence, the surface is described by the zero-crossing of the SDF. This enables surfaces to change topology easily and differentiably, which can be exploited for shape optimization purposes. Furthermore, boolean operations such as union and intersection that are traditionally used in constructive solid geometry (CSG) can be implemented using simple $\min$ and $\max$ operators on SDFs, which makes it easy to combine them into a single final shape~\cite{Haugo17}. For these reasons, we represent our three kinds of primitives as their signed distance functions, which can be computed as
\begin{equation}
	\SDF_{\text{prim}}(\bp)=\Psi_{\text{prim}}(\bS_{\text{prim}}, \bp),
\end{equation}
where $\Psi_{\text{prim}}$ is a differentiable volumetric function, augmented to also take as input the parameters $\bS_{\text{prim}}$ describing the shape. They can be explicit---such as radius and height---, implicit, such as latent vectors, or a combination of both. Additionally, we have rotation and translation parameters $\bR$ and $\bT$ that define the primitive's 6D pose within the overall shape space. In practice, they are implemented by transforming the query points $\bp$ as
\begin{equation} \label{eq:point_rt}
	\bp'=\bR^{\top}(\bp - \bT),
\end{equation}
which send them to the primitive's canonical space where the SDF is computed with $\Psi_{\text{prim}}$.

The SDF of our three primitive types are the following:
\begin{itemize}
	\item  {\bf Generic-SDF} ($\SDF_{\text{generic}}$): arbitrarily complex primitive whose SDF is $\SDF_{\text{generic}}(\bp)=\Psi_\text{g}(\bS_\text{g}, \bp)$, where $\Psi_\text{g}$ is an arbitrary volumetric function and $\bS_\text{g}$ its shape parameters, usually a latent vector $\LV_{\text{generic}}$ or the combination of one with explicit parameters. Rotation and translation are not used for this primitive.
	
	\item {\bf Geometric-SDF} ($\SDF_{\text{geom}}$): simple geometric primitive parameterized by a small number of shape parameters $\bS$. It can be computed analytically through a known function $\Psi_{\text{geom}}$. For example, the SDF of a sphere is given by $\Psi_\text{sphere}(\bS, \bp)=\|\bp\| - r$, where $\bS = (r)$ contains the sphere radius. Similar functions can be easily written for cylinders and other CAD primitives~\cite{Haugo17}.
		
	\item {\bf Geometry-assisted-SDF} ($\SDF_{\text{assist}}$): primitive with a close resemblance to a simple geometric one. As with generic primitives, its SDF is given by $\SDF_{\text{assist}}(\bp)=\Psi_\text{a}(\bS_\text{a}, \bp)$, where $\Psi_\text{a}$ is an arbitrary volumetric function and $\bS_\text{a}=(\LV_{\text{assist}}, \bS_{\text{assist}})$ is the concatenation of a latent vector and geometric shape parameters $\bS_{\text{assist}}$. Additionally, we want to constrain its SDF to be close to its assisting geometry's. Therefore, we define another geometric-SDF $\SDF_{\text{geom}}$ with the same shape parameters $\bS_{\text{assist}}$. It will only be used during training where we enforce $\SDF_{\text{assist}}(\bp)\approx\SDF_{\text{geom}}(\bp)$ through the geometry-assistance loss $\mathcal{L}_{ga}$, as described in Section~\ref{sec:training}. Both share the same rotation and translation parameters. For instance, the car wheels in Fig.~\ref{fig:teaser} are given by $\SDF_{\text{assist}}$, which are constrained to be similar to $\SDF_{\text{geom}}$ of the assisting cylinders. Note that only $\SDF_{\text{assist}}$ is used to build the final shape.
\end{itemize}
These three different primitive types are depicted by  Fig.~\ref{fig:teaser} and on the right side of Fig.~\ref{fig:arch}. We use neural networks to implement the arbitrary volumetric functions $\Psi_\text{g}$ and $\Psi_\text{a}$, which we refer to as deep implicit functions. Eventually, the full shape is taken as the union of all primitives, thus its SDF can be directly computed by applying the $\min$ operator over all SDFs:
\begin{align}
  \SDF_\text{full}        & = \min  (\SDF_\text{generic}, \SDF_\text{geom},  \SDF_\text{assist}) \,, \label{eq:sdf_full}
\end{align}
with
 \begin{align}
	 \SDF_\text{geom}  & =  \min (\SDF_\text{geom}^1, \ldots, \SDF_\text{geom}^{n_g})  \, ,     \\
	 \SDF_\text{assist}   &=    \min ( \SDF_\text{assist}^1,  \ldots,  \SDF_\text{assist}^{n_a}) \, ,     
\end{align}
where $n_g$ and $n_a$ are the number of geometric and assisted primitives respectively.  The shape is then defined as the zero-crossing of $\SDF_\text{full}$.

The number of primitives depends on the part decomposition of the shapes and, in practice, is known to the engineers performing the design. Generally, parts with strong regularities should be represented by either geometric or geometry-assisted primitives, while more complex ones should be generic primitives.

\subsection{Network}
\label{sec:network}

We use a mix between an auto-decoder and an auto-encoder framework. Our shapes are decoded from a vector of parameters, as in~\cite{Park19c, Hao20}. However, unlike in these works, we handle a combination of implicit and explicit, or latent and geometric, parameters and we rely on a disentangled representation where the different parameters account for separate dimensions. This makes it possible to support disentangled manipulation within that parameter space. As described below, we use a point cloud encoder to predict the parameters of the geometric and geometry-assisted primitives, while the latent vector of the generic one is trained in the auto-decoder fashion, as in~\cite{Park19c}. In Appendix~\refappendix{appendix:hss}{D}, we also explore a purely auto-decoder variant of our approach where all parameters are instead predicted from a unique latent vector, as in~\cite{Park19c, Hao20}.

For the decoding, our network is made of multiple branches that decode the various primitive types for a given input, as illustrated in Fig.~\ref{fig:arch}.
The first branch decodes the concatenation of a latent vector $\LV_{\text{generic}}$ and the geometric parameters $\bS$, $\bR$, and $\bT$ into the $\SDF_{\text{generic}}$ through the \textit{Generic Decoder} network, whose architecture is taken from~\cite{Park19c}. We add an auxiliary output $\SDF_{\text{geom}^{\dagger}}$ during training that must predict all geometric primitives in order to ensure the consistency between them and the generic primitive, as explained in Section~\ref{sec:training}.
Then, we use analytical functions to compute the $\SDF_{\text{geom}}$ with the explicit parameters of the different geometric primitives. During training, this also computes the SDF of the assisting geometries.
Finally, the last branch decodes $\SDF_{\text{assist}}$ of the geometry-assisted primitives through the \textit{Part Decoder} network, similar in architecture to~\cite{Park19c} but with only three layers as the shapes it produces are simpler. Its input is the concatenation of a latent vector $\LV_\text{assist}$ with the shape parameters $\bS$ of the assisting geometry, while $\bR$ and $\bT$ are used to transform the position of the query point, as described above.

\paragraph{Point encoder.}

The explicit parameters $\bS$, $\bR$, and $\bT$, as well as $\LV_{\text{assist}}$, are predicted by a \textit{Point Encoder} that processes point clouds obtained from the training surfaces, as shown in Fig.~\ref{fig:arch}. Its architecture is similar to that of PointNet~\cite{Qi17a} with fully connected layers and ReLU non-linearities, and additional skip connections. 
We have empirically found this encoder to be necessary because directly training for these parameters in an auto-decoding framework would update them only once per epoch, which lead to early convergence in bad local minima, as shown in Appendix~\refappendix{appendix:pe}{C.4}.
During training, the \textit{Point Encoder} provides the values for $\bS$, $\bR$, $\bT$, and $\LV_{\text{assist}}$, while in inference reconstruction it generates initial estimates that are optimized. It is however not needed for manipulation as users can instead manually edit the parameters. For different reconstruction tasks, such as single-image reconstruction which we explore below, it may be replaced by an encoder that accepts a different modality.


\subsection{Training losses}
\label{sec:training}

\HS{} learns to represent a set of 3D meshes from which SDF samples are extracted. It additionally leverages part labels, also in the form of meshes, to ensure that the primitives model the appropriate object parts. The model is trained on a weighted sum of losses with respect to its networks weights and latent vector $\LV_{\text{generic}}$, jointly with the \textit{Point Encoder}. Mind that the formulas presented below are given for a single shape for clarity but are really summed over the training set, or rather a mini-batch per iteration. 
For each shape, we generate $K$ sample points in the 3D space on which all losses are computed to speedup training. In the following text, it is assumed that $\SDF_{\circ}$ refers to a vector of the $K$ points' signed distances, using the $\delta$-clamped distances: $\text{clamp}_\delta(x)=\min(\delta, \max(-\delta, x))$, as in~\cite{Park19c}.


\paragraph{Reconstruction losses.} 
In essence, the reconstruction is a regression problem aiming to make the predicted SDF as close as possible to the ground-truth (GT). 
We write
\begin{equation}
	\mathcal{L}^{f}_{r} = \frac{1}{K}  \left\lVert\SDF_{\text{full}} - \SDF^{\text{GT}}_{\text{full}} \right\rVert_1\, ,
	\label{eq:loss_recf}
\end{equation}
that is, the $L1$-loss computed over the $K$ samples for the full shape, where $\SDF^\text{GT}_{\text{full}}$ are the GT signed distance values.

For the parts, the labels can be noisy in practice, \eg if they are automatically generated~\cite{Kalogerakis17}. Therefore, we introduce the robust part reconstruction loss
\begin{equation}
\mathcal{L}^{p}_{r} = \sum_\text{part} \frac{1}{K'} \sum_{l=1}^{K'} Sort\left\{ \left|\SDF_{\text{part}} - \SDF^{\text{GT}}_{\text{part}} \right| \right\}_l \,,
\label{eq:loss_recp}
\end{equation}
where $\SDF_{\text{part}}$ is the geometric- or geometry-assisted-SDF corresponding to the part, $\SDF^{\text{GT}}_{\text{part}}$ its ground truth, $Sort$ rearranges the samples in increasing order of values, and $K'=\lfloor\gamma K\rfloor$, with $\gamma\in[0, 1]$. In other words, we only consider the $K'$ samples with the lowest error. The hyper-parameter $\gamma$ trades off robustness for reconstruction fidelity: the higher the noise, the lower $\gamma$ should be set.
For shape without part labels, we simply set $\mathcal{L}^{p}_{r}=0$.


\paragraph{Geometry assistance.}
The geometry-assisted primitives are constrained to resemble their corresponding geometries by
%
\begin{equation}
	\mathcal{L}_{ga} = \sum_{n=1}^{n_a}\frac{1}{K}  \left\lVert\SDF_{\text{assist}}^{n} - \SDF_{\text{geom}}^{n_g+n} \right\rVert_1 \,,
	\label{eq:loss_ps}
\end{equation}
where $\SDF_{\text{geom}}^{n_g+n}$ is the SDF of the geometric primitive used to constrain $\SDF_{\text{assist}}^{n}$. 


\paragraph{Intersection loss.}
To prevent primitives from overlapping, we minimize an intersection loss between all pairs:
 \begin{equation}
	 \mathcal{L}_{ic} = \sum_{x,y} \frac{1}{K} \Big\lVert \Theta\big(\SDF_{\text{x}}, \SDF_{\text{y}}\big) \Big\rVert_1 \, ,   \label{eq:loss_inter}
\end{equation}
where $\Theta$ computes the overlap between two SDFs as
\begin{equation}
	\Theta (\SDF_{\text{x}},\SDF_{\text{y}}) = \max \! \Big( \!\! - \! \max \big( \SDF_{\text{x}}, \SDF_{\text{y}} \big), \mathbf{0} \Big) \, .
\end{equation}
%


%
%


\paragraph{Consistency loss.}
In order to enforce the \textit{Generic Decoder} to account for the explicit geometric parameters $\bS$, $\bR$, and $\bT$, a secondary output is added that must predict the SDF of all geometric primitives. This output is trained with the following loss:
\begin{equation}
\mathcal{L}_{cs} = \frac{1}{K} \Big\lVert \SDF_{\text{geom}^{\dagger}} - \min\limits_{n=1}^{n_g + n_a} \SDF_{\text{geom}}^{n} \Big\rVert_{1} \,,
 \label{eq:loss_HSd}
\end{equation}
where $\SDF_{\text{geom}^{\dagger}}$ denotes the auxiliary output as depicted in Fig.~\ref{fig:arch}. As we show in Appendix~\refappendix{appendix:l_cs}{C.3}, we found this necessary in order to train the decoder to model the strong dependency of the generic primitive on the geometric parameters.

\paragraph{Regularization loss.}
Finally, we apply $L2$-regularization to all the latent vectors, $\LV_{\text{generic}}$ and $\LV_{\text{assist}}$, as in \DeepS{}, and refer to the sum as $\mathcal{L}_{reg}$.

\section{Experiments}
\label{sec:exp}

Our approach is designed to provide stronger regularity for the parts corresponding to the geometric and geometry-assisted primitives, alongside precise and easy 3D shape manipulation via geometric parameter edition, by using explicit parameters describing shape attributes ($\bS$) and 6D pose ($\bR$, $\bT$). Additionally, it enables to automatically extract these parameters from single images and sketches. These facets of our work are explored below, and we provide additional results and an ablation study in Appendices~\refappendix{appendix:exp}{E} and~\refappendix{appendix:ablation}{C} respectively.


\paragraph{Datasets.} 
We report experimental results on two categories of objects: {\it Cars} from ShapeNet~\cite{Chang15} to evaluate our method against existing work on frequently used shapes, and a new dataset of {\it Static Mixers} that are representative of complex manufactured objects.

\emph{Cars.}
We represent the cars in terms of a geometry-assisted-SDF for each wheel (using cylinders) and a generic-SDF for the car body, as shown in Fig.~\ref{fig:teaser}(a). We obtained the wheel labels from the noisy mesh labels of~\cite{Kalogerakis17}. This gives us 2283 training shapes, 247 of which have part labels, and 500 test shapes.

\emph{Static Mixers.}
Used to mix fluids or powders that have different chemical properties, they comprise a static internal helix that is encased in a central tube with screws at both ends, as shown in Fig.~\ref{fig:teaser}(h). It is natural to represent the central tube and the tube-like parts associated with the screw by geometric-SDFs (hollow cylinders), the screw rings by geometry-assisted-SDFs (hollow cylinders), and the helix whose shape can vary most by a generic-SDF. Examples are provided in Appendix~\refappendix{appendix:data}{A}. We use 1500 shapes for training and 450 for testing. During training, we use part labels for 100 of the exemplars.

\paragraph{Baselines.}
We compare against \DeepS~\cite{Park19c}, an auto-decoding network predicting purely generic SDFs, similar to our \textit{Generic Decoder}, \DualS~\cite{Hao20}, an hybrid approach for shape manipulation, \HSQ~\cite{Paschalidou20}, a primitive-based representation using a hierarchy of superquadrics, and \NP~\cite{Paschalidou21}, another hybrid method that uses network-predicted primitives to improve fidelity.
For all of the above baselines, we used the code provided by the authors and we trained their networks on our data.

\subsection{Shape reconstruction}
\label{sec:exp_recon}

We train our approach and the baselines on the training shapes, then we reconstruct the cars and mixers from the test sets and report quantitative results in Table~\ref{table:quant_eval}. We use $L_2$-Chamfer Distance (CD), Earth Mover's Distance (EMD), and shell-IoU (sIoU) as in~\cite{Remelli20b,Guillard20}. Shell-IoU is the intersection over union computed on voxelized surfaces of reconstructed and ground truth shapes and should be maximized, in contrast to CD and EMD. 


\newlength{\mytabcolsep}
\setlength\mytabcolsep{\tabcolsep}
\setlength\tabcolsep{2pt}
\begin{table}
\begin{center}
{
	\begin{tabular}{@{}llccccccc@{}}
		\toprule
		&& \multicolumn{3}{c} {\textbf{Mixers}} && \multicolumn{3}{c} {\textbf{Cars}}  \\
		&& CD$\downarrow$ & EMD$\downarrow$ & sIoU$\uparrow$&& CD$\downarrow$ & EMD$\downarrow$ & sIoU$\uparrow$ \\ \midrule
		\DeepS && {\it 2.27} & {\it 2.76} & {\it 86.3} && {\bf 11.9} & {\bf 1.44} & {\bf 54.9}\\
		\DualS  && 4.56 & 3.17 & 77.8 &&  17.4 & 1.88 & 49.1 \\
		\HSQ  && 19.6  & 7.03 & 28.3  && 27.8 & 2.63 & 36.8 \\
		\NP && 6.43 &  13.8 & 44.1 && 25.2 & 2.17 & 40.0\\ \midrule
		\HS  && {\bf 1.57} & {\bf  2.51} & {\bf 90.9} && {\it 12.5} & {\it 1.48} & {\it 52.9}\\   
		\bottomrule
	\end{tabular}
}
\caption{\textit{Reconstruction results on the test shapes.} We report average CD (30,000 points) multiplied by 10$^4$, EMD (10,000 points) multiplied by 10$^2$, and sIoU in [\%]. Bold and italic numbers are best and second best performances respectively.}
\label{table:quant_eval}
\end{center}
\end{table}
\setlength{\tabcolsep}{\mytabcolsep}

\newlength{\reconfigwidth}
\setlength{\reconfigwidth}{0.12\textwidth}
\newlength{\reconfigwidthmix}
\setlength{\reconfigwidthmix}{0.09\textwidth}

\setlength\mytabcolsep{\tabcolsep}
\setlength\tabcolsep{2pt}

\begin{figure*}
 \centering
 \small  
 \begin{tabular}{cccccc}
   \includegraphics[width=\reconfigwidth]{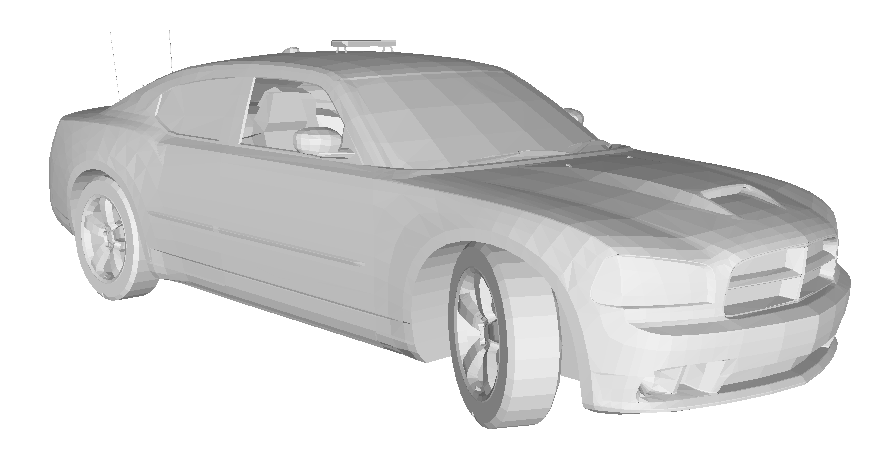} &
   \includegraphics[width=\reconfigwidth]{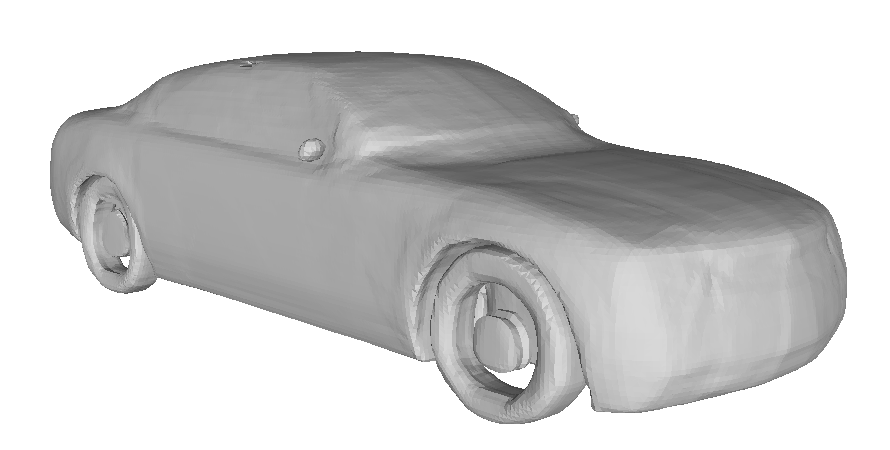}&
   \includegraphics[width=\reconfigwidth]{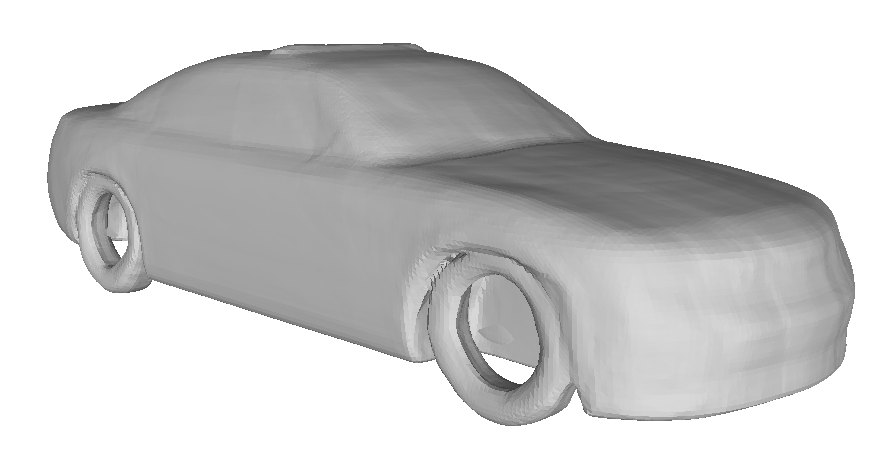} &
   \includegraphics[width=\reconfigwidth]{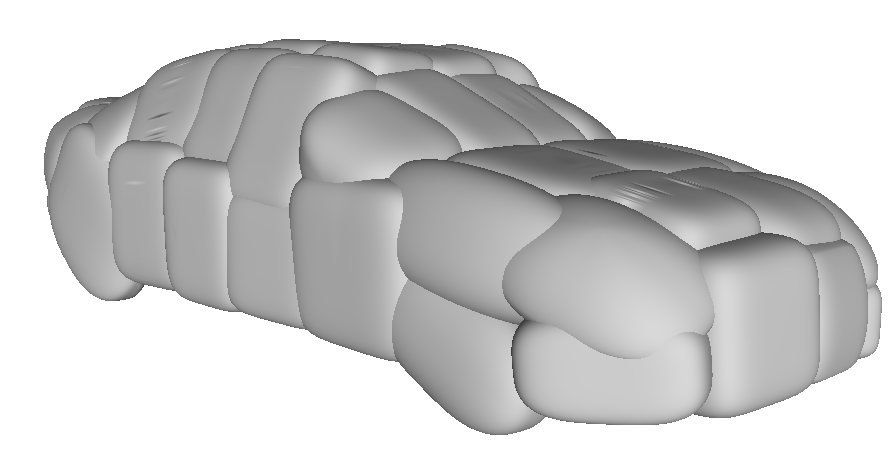} &
   \includegraphics[width=\reconfigwidth]{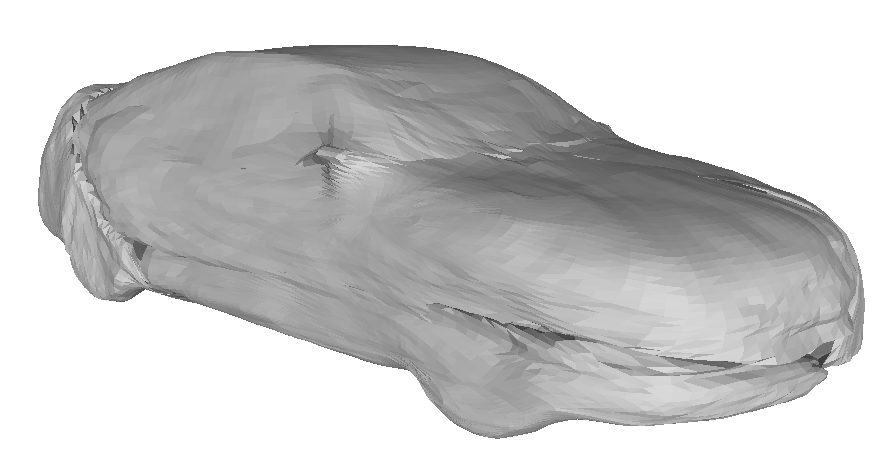}&
   \includegraphics[width=\reconfigwidth]{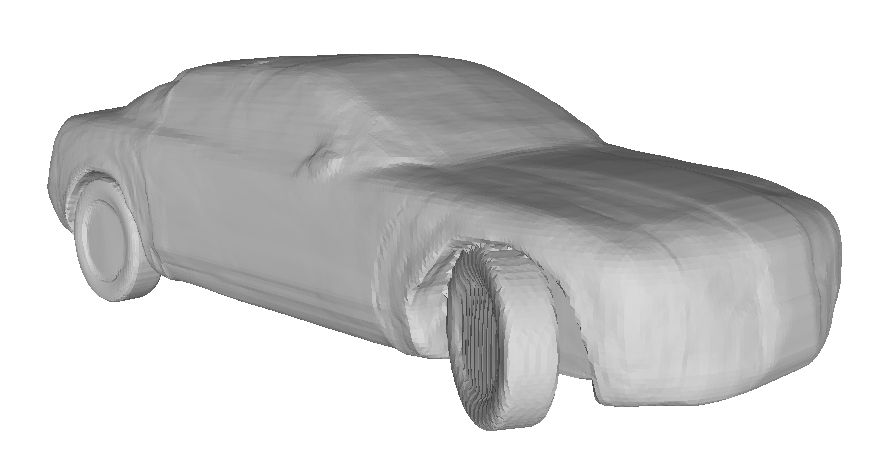}\\
   \includegraphics[width=\reconfigwidth]{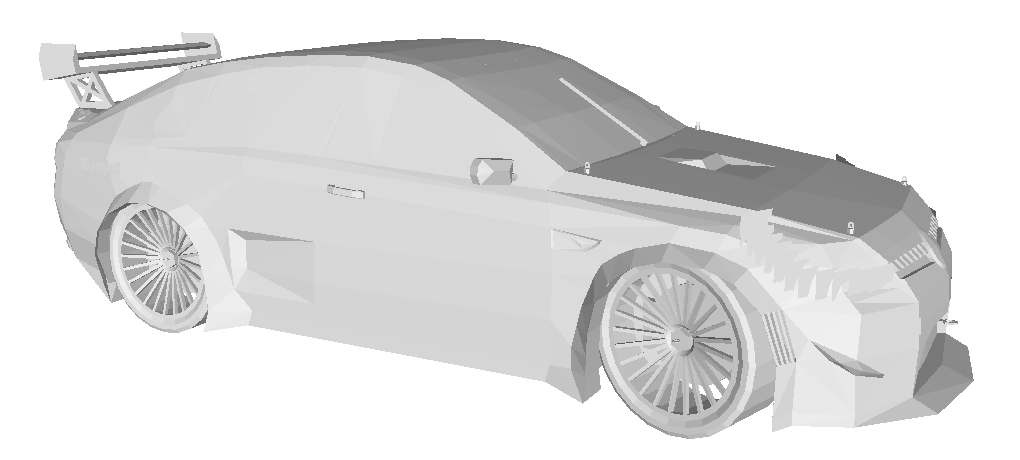} &
   \includegraphics[width=\reconfigwidth]{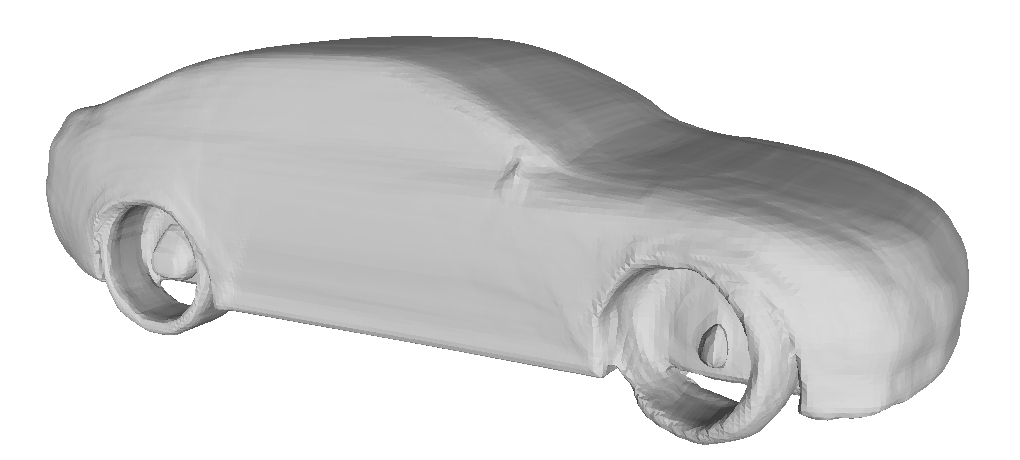}&
   \includegraphics[width=\reconfigwidth]{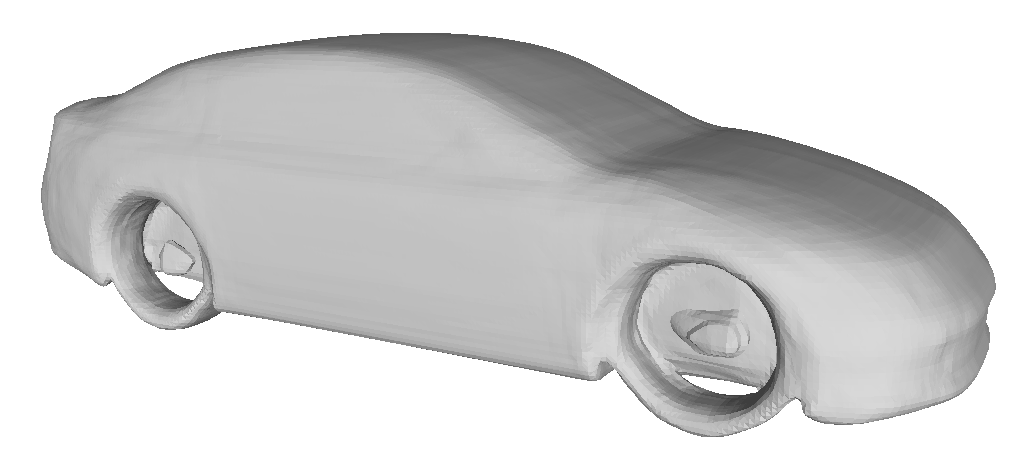} &
   \includegraphics[width=\reconfigwidth]{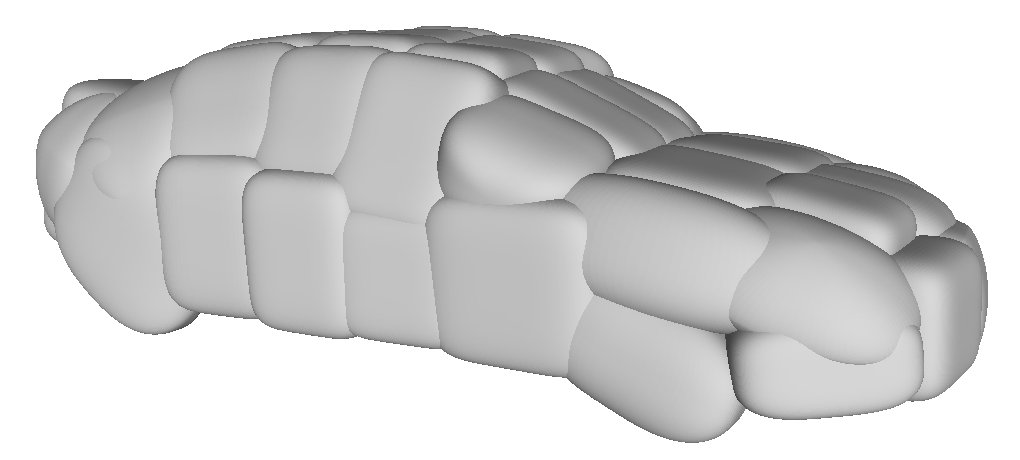} &
   \includegraphics[width=\reconfigwidth]{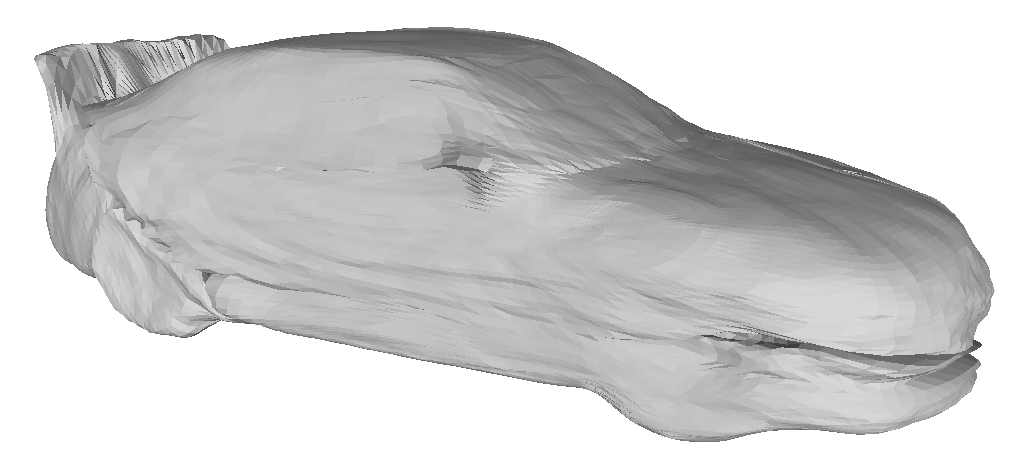}&
   \includegraphics[width=\reconfigwidth]{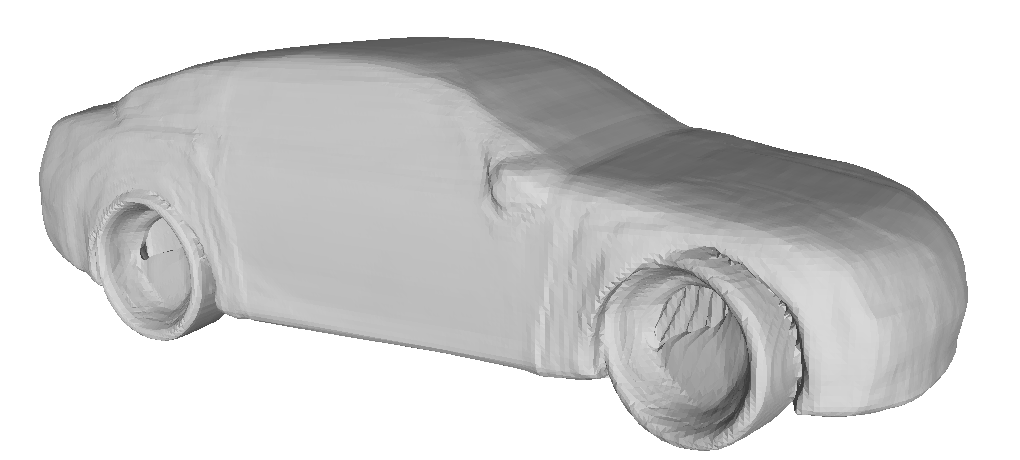}\\
   \includegraphics[width=\reconfigwidth]{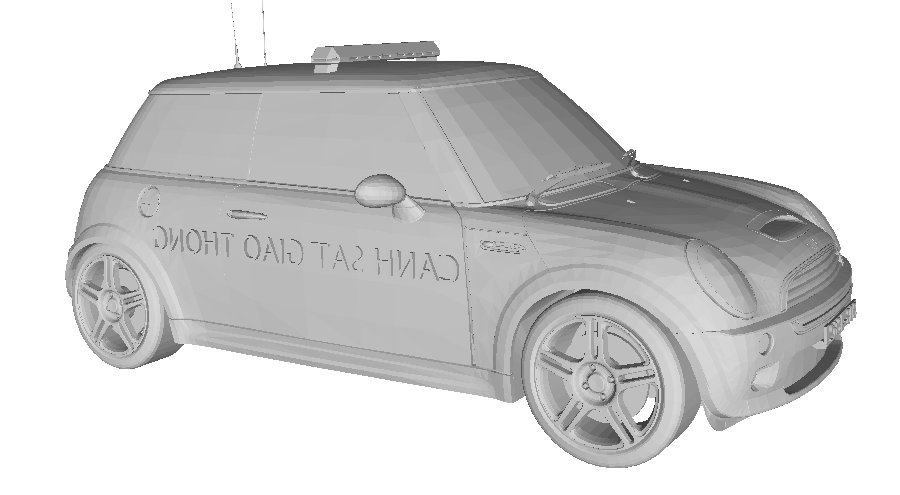} &
   \includegraphics[width=\reconfigwidth]{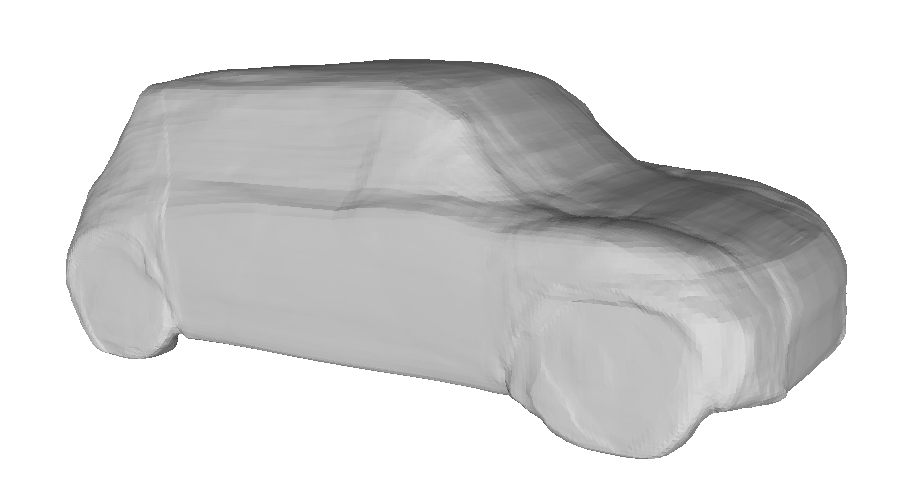}&
   \includegraphics[width=\reconfigwidth]{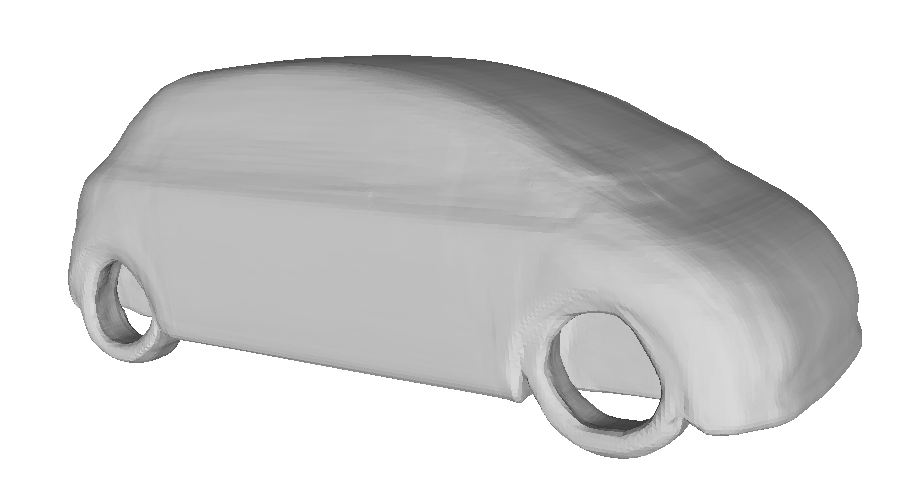} &
   \includegraphics[width=\reconfigwidth]{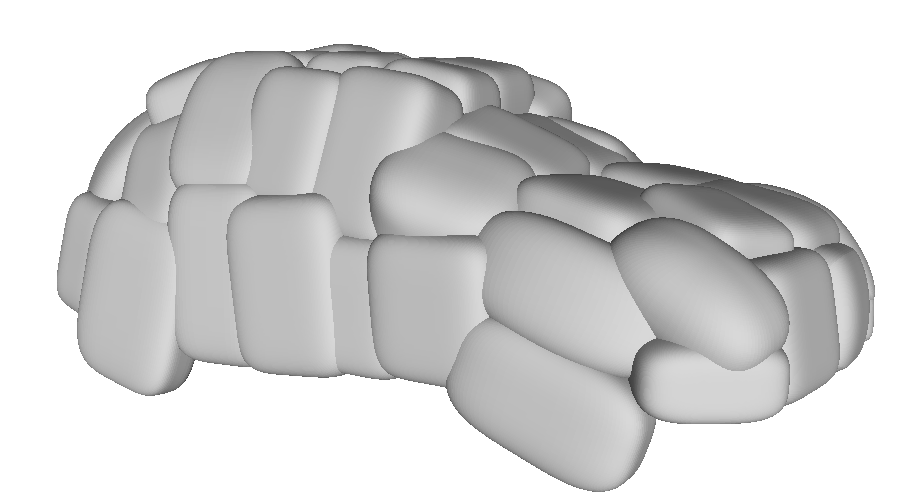} &
   \includegraphics[width=\reconfigwidth]{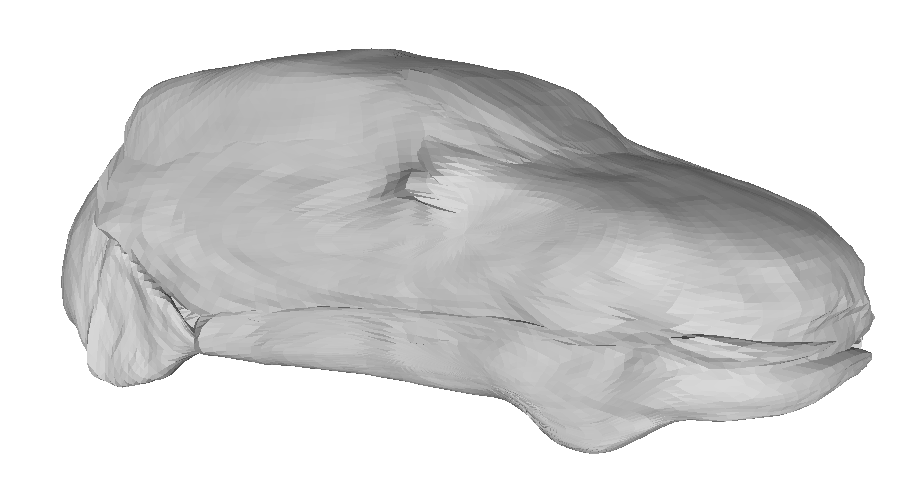}&
   \includegraphics[width=\reconfigwidth]{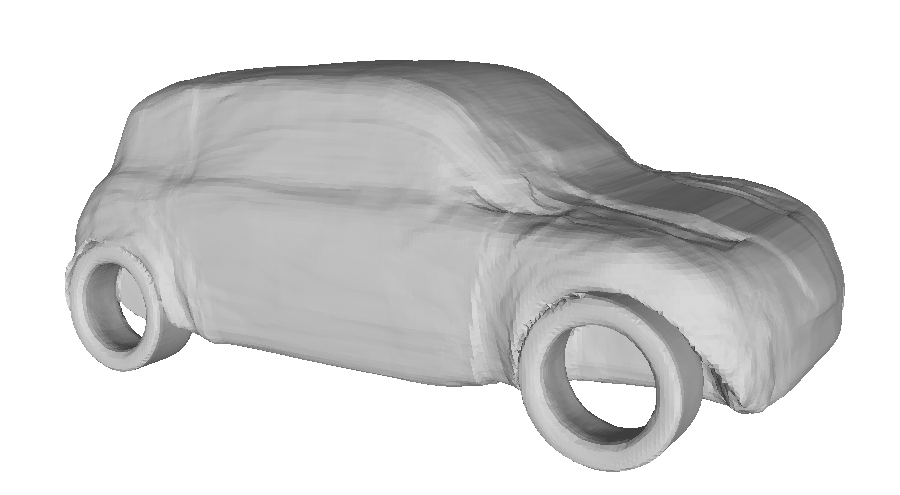}\\ 
\includegraphics[width=\reconfigwidth]{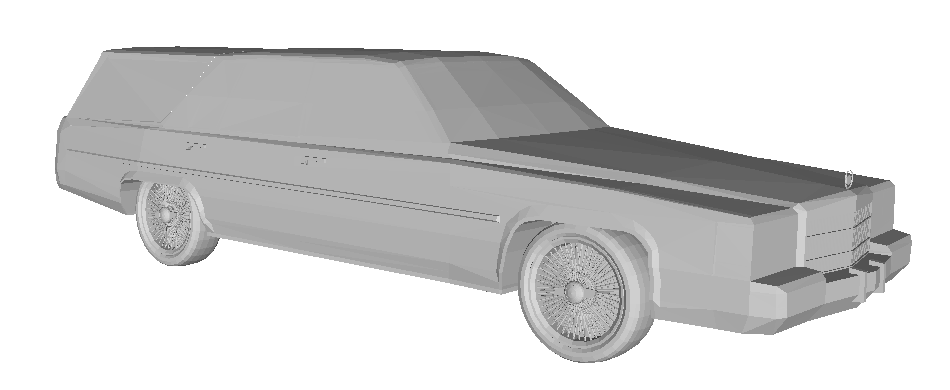} &
\includegraphics[width=\reconfigwidth]{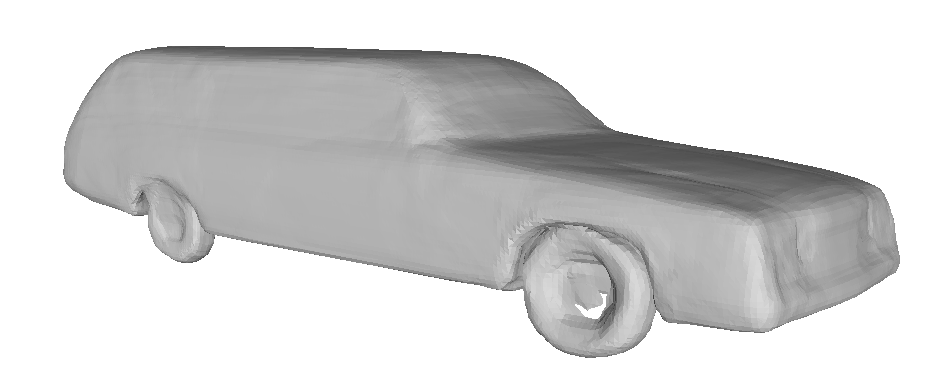}&
\includegraphics[width=\reconfigwidth]{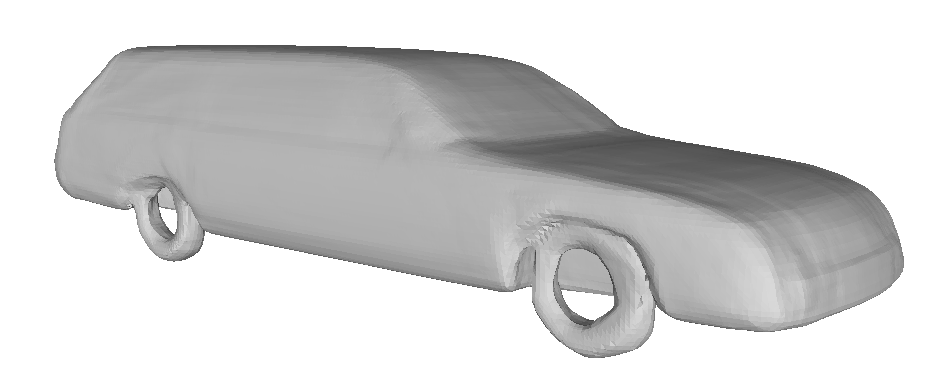} &
\includegraphics[width=\reconfigwidth]{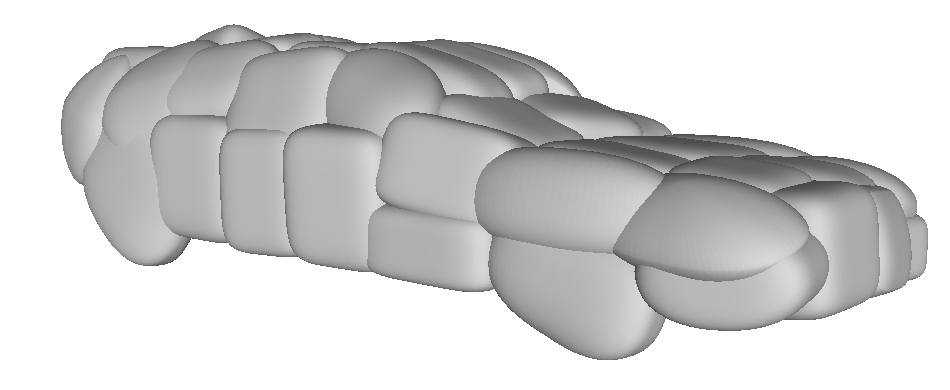} &
\includegraphics[width=\reconfigwidth]{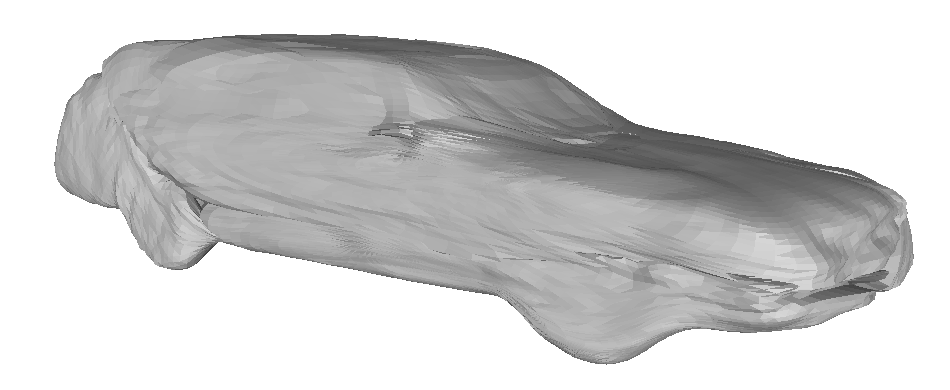}&
\includegraphics[width=\reconfigwidth]{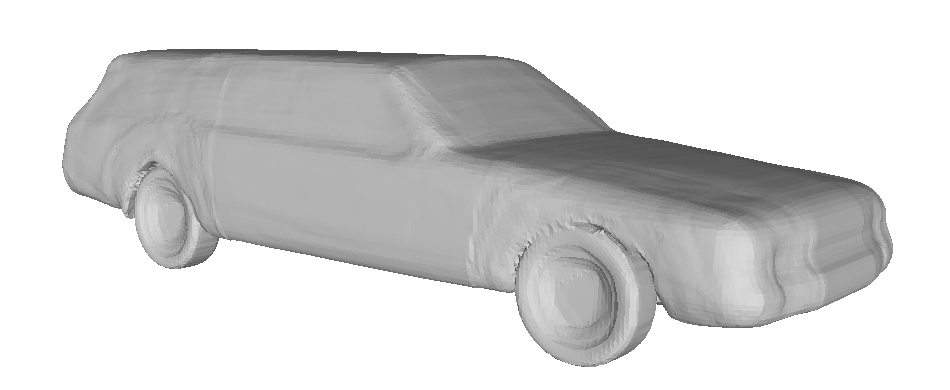}\\
   \includegraphics[width=\reconfigwidth]{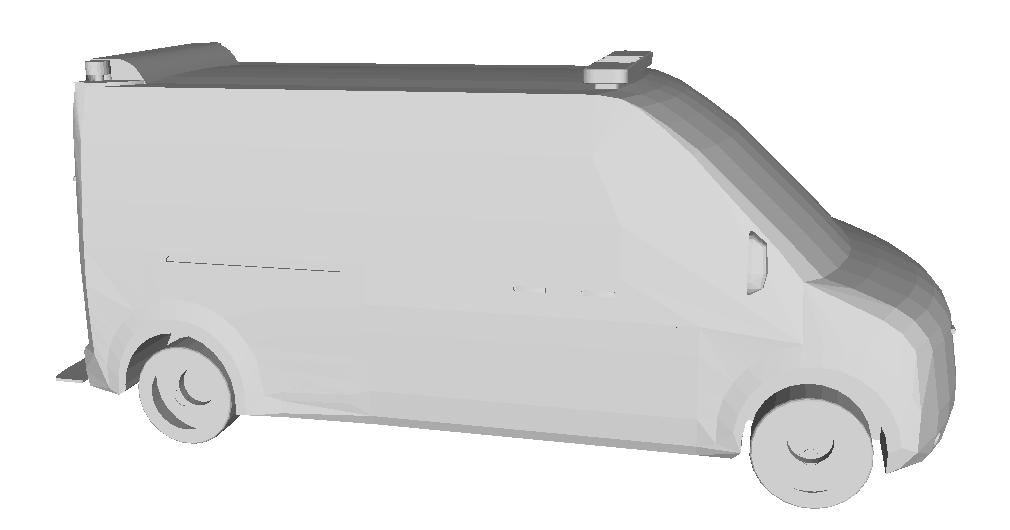} &
   \includegraphics[width=\reconfigwidth]{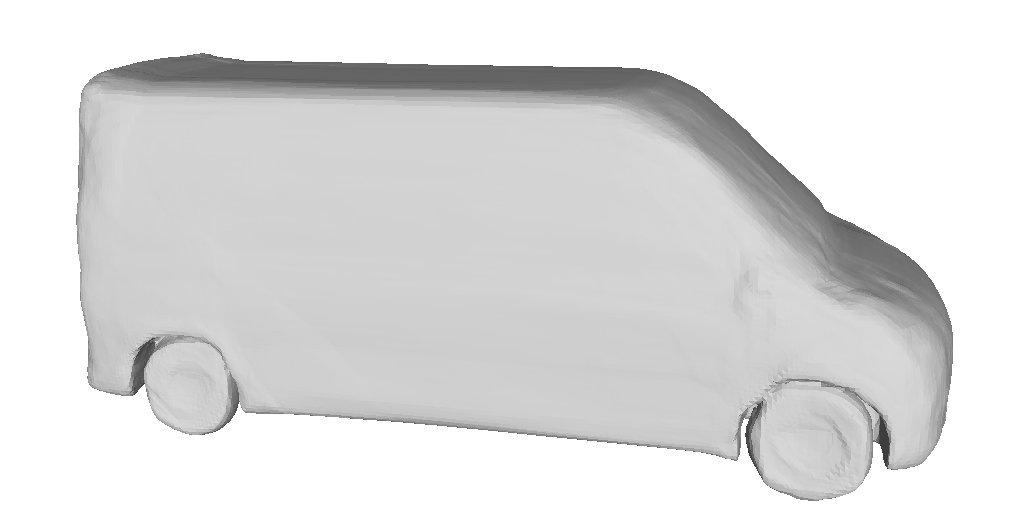}&
   \includegraphics[width=\reconfigwidth]{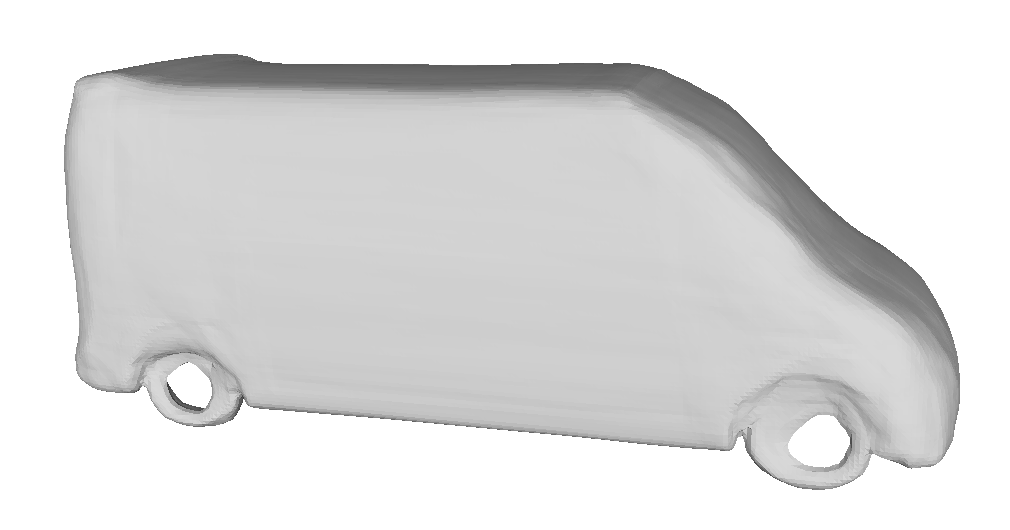} &
   \includegraphics[width=\reconfigwidth]{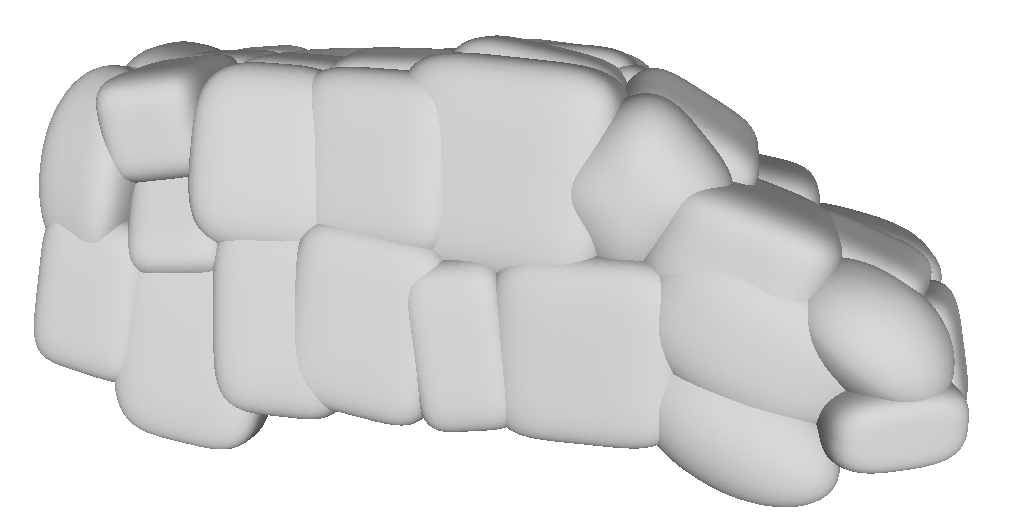} &
   \includegraphics[width=\reconfigwidth]{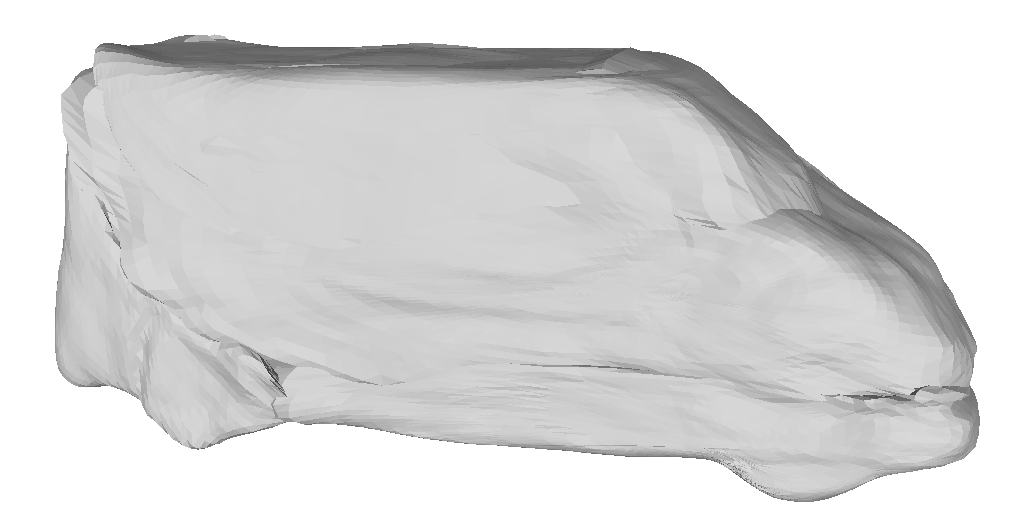}&
   \includegraphics[width=\reconfigwidth]{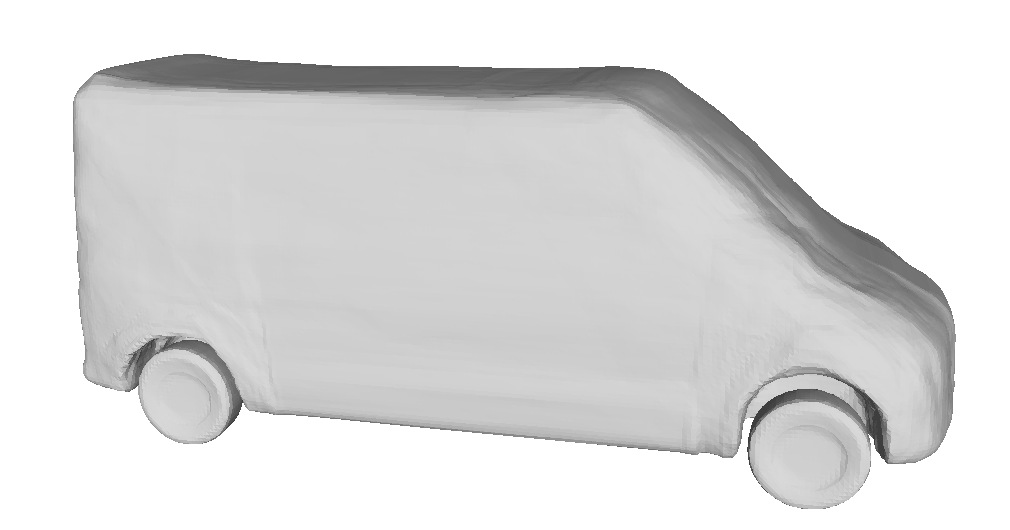}\\ 
   \includegraphics[width=\reconfigwidth]{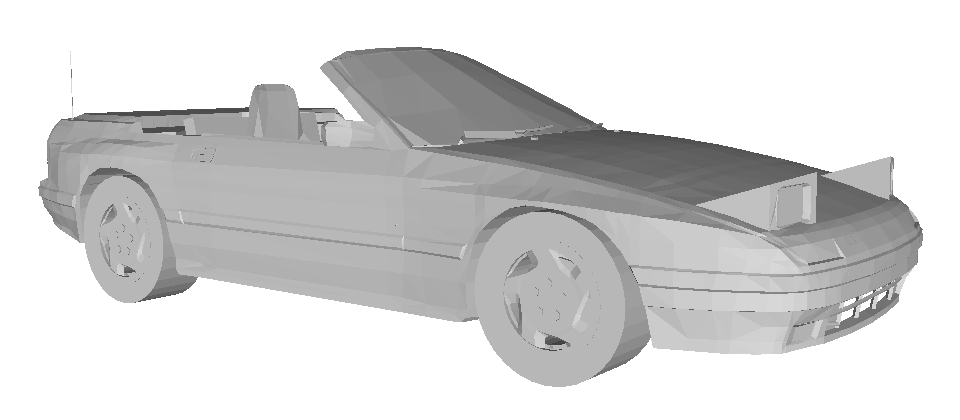} &
   \includegraphics[width=\reconfigwidth]{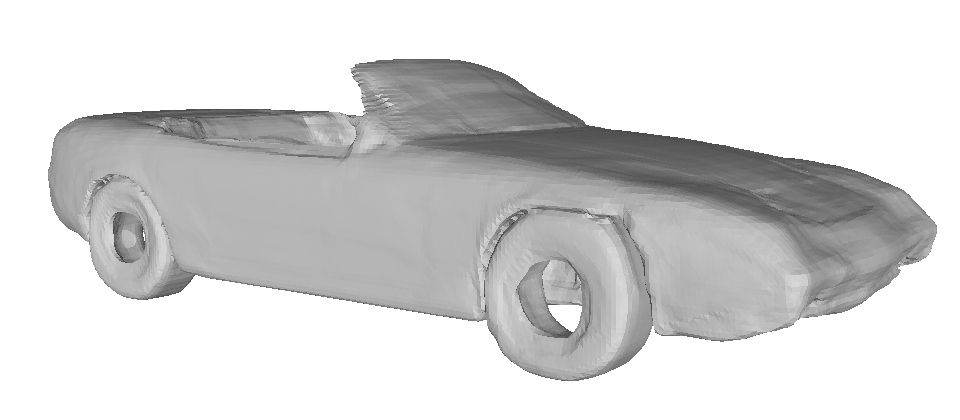}&
   \includegraphics[width=\reconfigwidth]{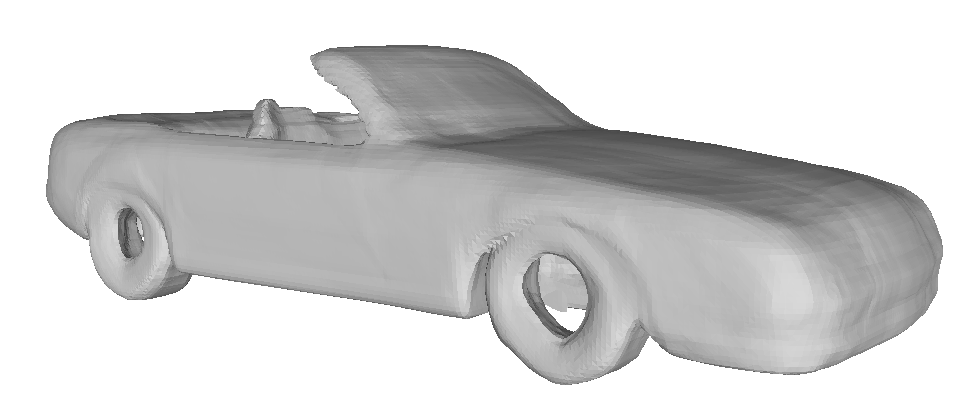} &
   \includegraphics[width=\reconfigwidth]{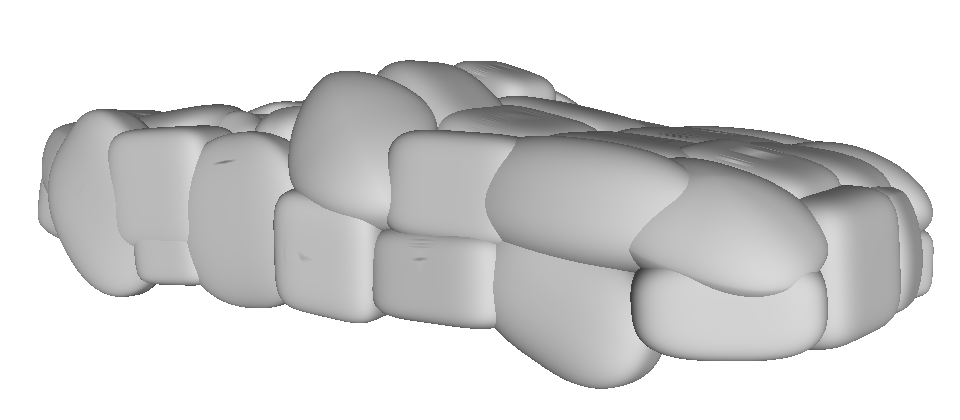} &
   \includegraphics[width=\reconfigwidth]{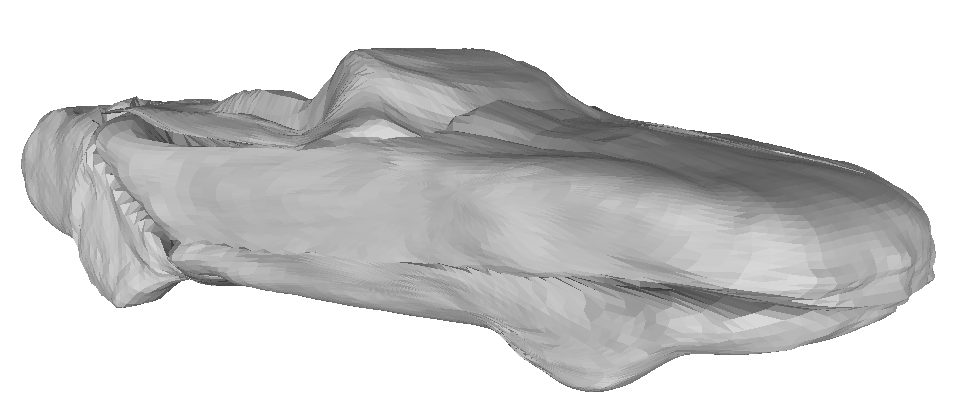}&
   \includegraphics[width=\reconfigwidth]{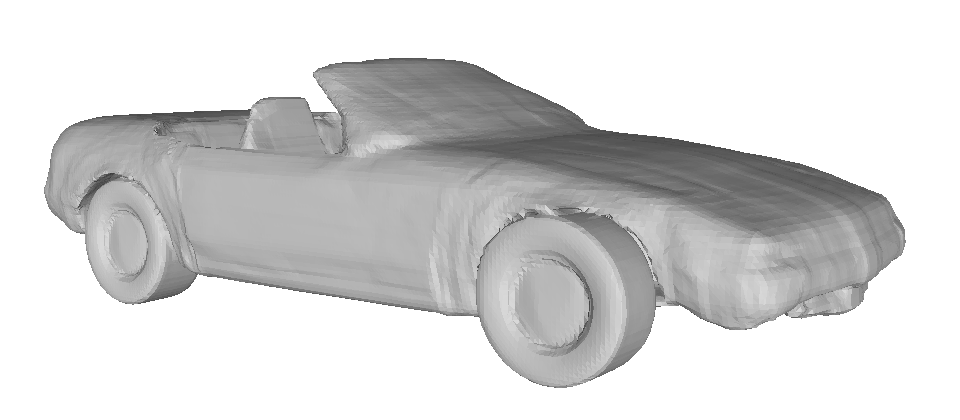}\\
   \hline 
   \includegraphics[width=\reconfigwidthmix,trim=0 0 0 -5]{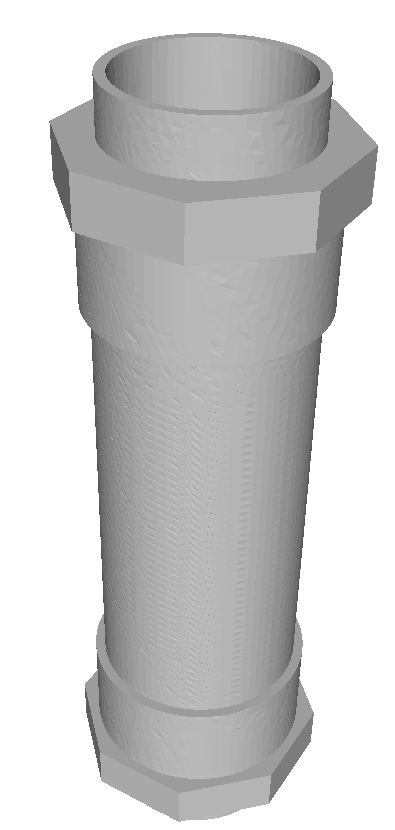} &
   \includegraphics[width=\reconfigwidthmix]{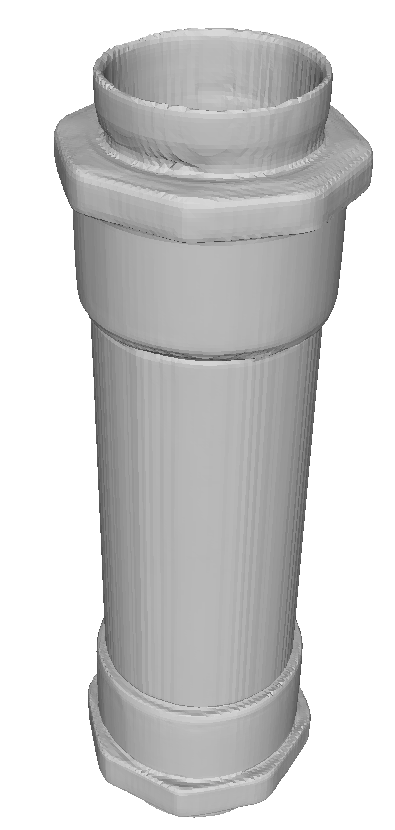}&
   \includegraphics[width=\reconfigwidthmix]{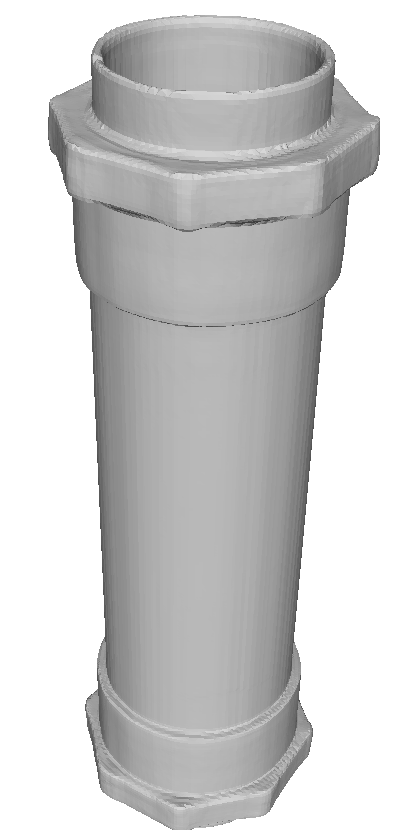} &
   \includegraphics[width=\reconfigwidthmix]{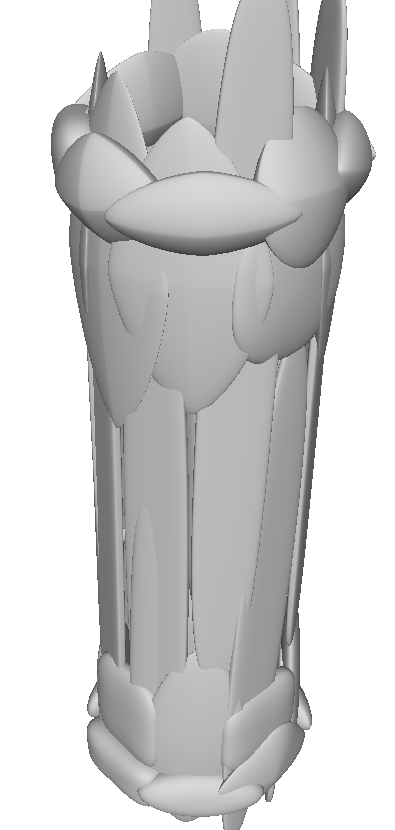} &
   \includegraphics[width=\reconfigwidthmix]{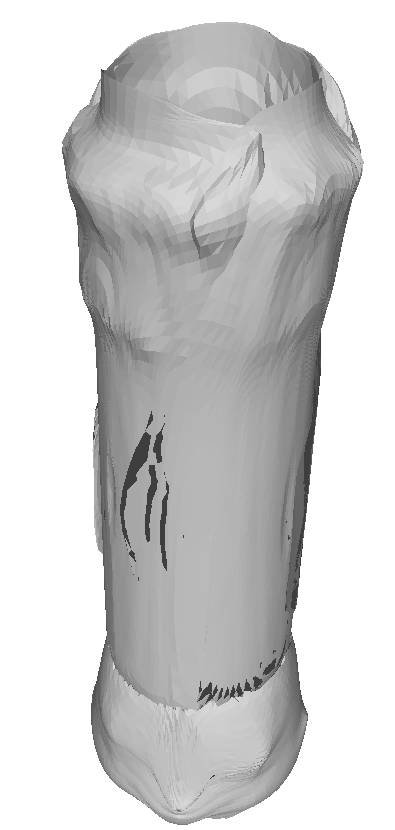}&
   \includegraphics[width=\reconfigwidthmix]{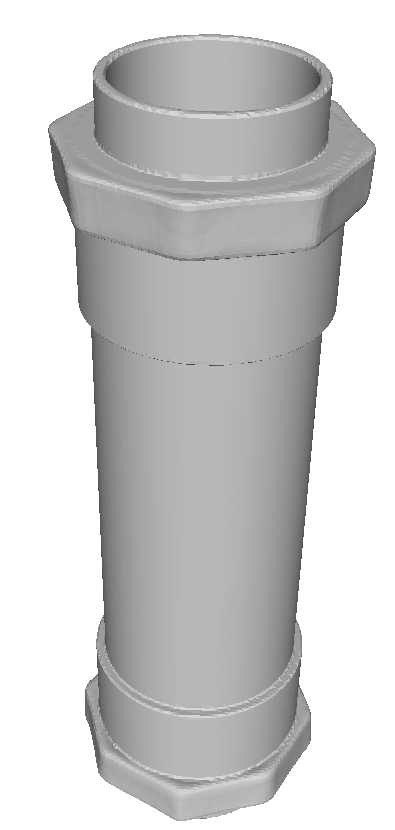}\\
   \includegraphics[width=\reconfigwidthmix]{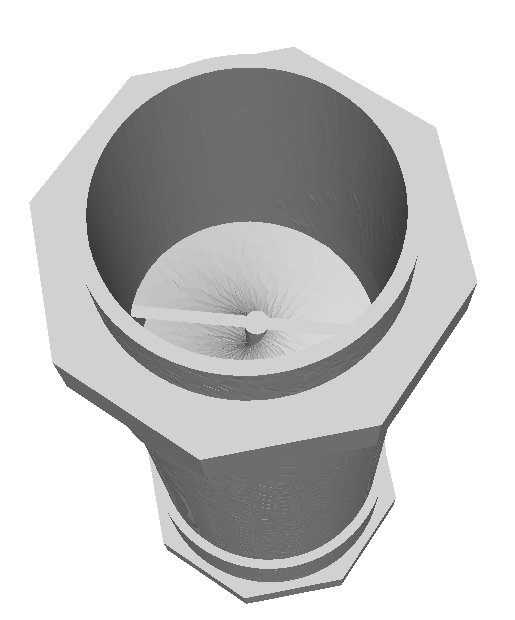} &
   \includegraphics[width=\reconfigwidthmix]{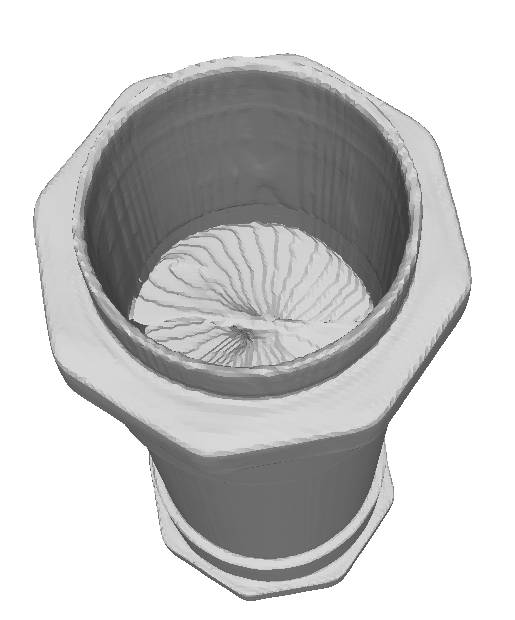}&
   \includegraphics[width=\reconfigwidthmix]{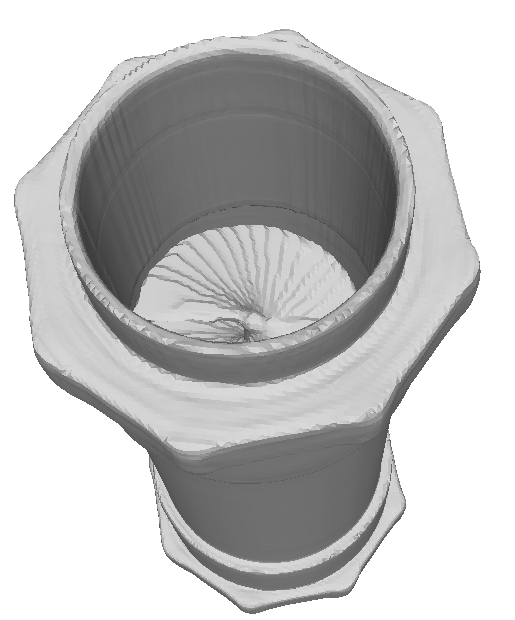} &
   \includegraphics[width=\reconfigwidthmix]{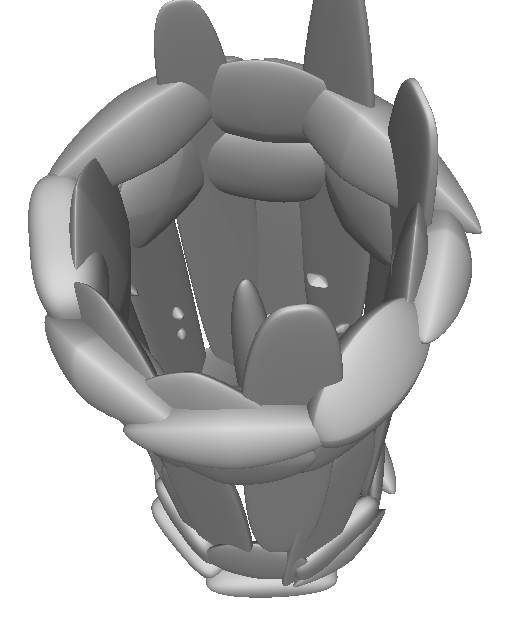} &
   \includegraphics[width=\reconfigwidthmix]{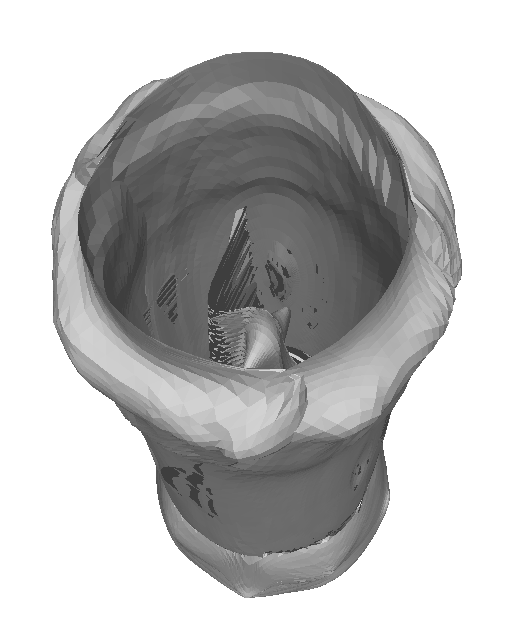}&
   \includegraphics[width=\reconfigwidthmix]{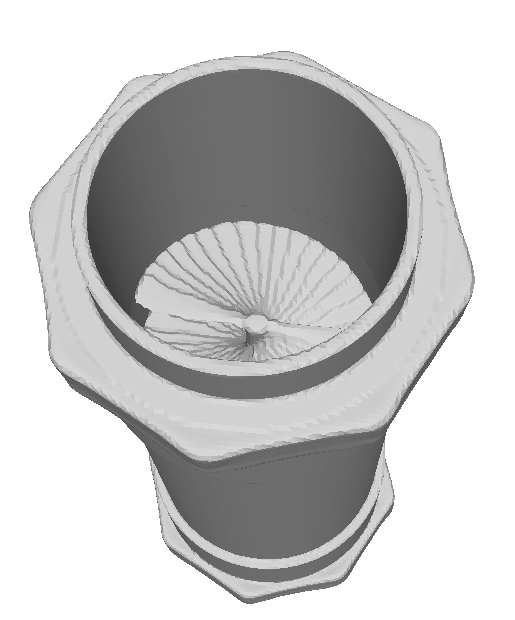}\\ 
   \includegraphics[width=\reconfigwidthmix]{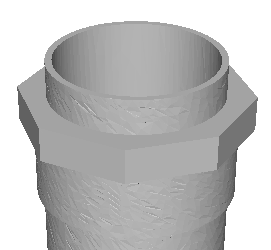} &
   \includegraphics[width=\reconfigwidthmix]{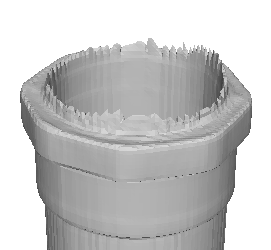}&
   \includegraphics[width=\reconfigwidthmix]{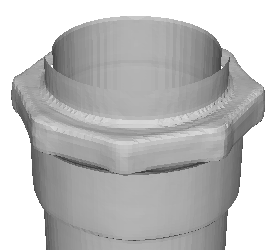} &
   \includegraphics[width=\reconfigwidthmix]{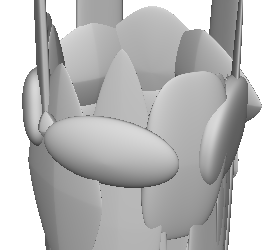} &
   \includegraphics[width=\reconfigwidthmix]{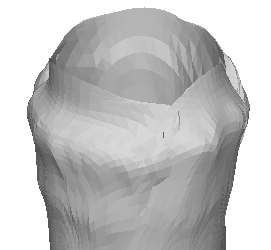}&
   \includegraphics[width=\reconfigwidthmix]{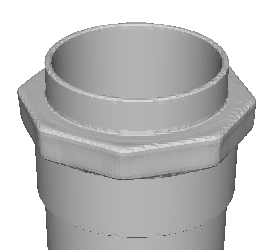}\\
   GT & \DeepS & \DualS & \HSQ & \NP & \HS \\
  \end{tabular}
  \caption{\textit{Qualitative comparison of reconstruction results.} {  ({\it Top rows}) Ground-truth for different cars and corresponding reconstructions. \DeepS{}, \DualS{}, and \HS{} deliver the most visually appealing results. However,  \HS{}, unlike \DeepS{} and \DualS{}, correctly captures the wheel orientation on the first example, and the wheel wells on the third. ({\it Bottom rows}) Ground-truth for different mixers and reconstructions. \DeepS{}, \DualS{}, and \HS{} again yield the best looking results but both \DeepS{} and \DualS{} produce surface irregularities near the screw, while \HS{} does not. Interestingly, \HSQ~and \NP~completely fails to represent the helix, underscoring that iso-primitives are not the optimal way to represent shapes.}}
  \label{fig:recon}
\end{figure*}

\setlength{\tabcolsep}{\mytabcolsep}

For the the mixers, our approach yields the best accuracy. This highlights that \HS{} is well suited to manufactured objects partially defined by geometric primitives. On the cars, \DeepS{} does marginally better, which may be counterintuitive given the qualitative mistakes \DeepS{} sometimes makes, as can be seen in Fig.~\ref{fig:recon}. We ascribe this to the fact that our learning task is harder. The \textit{Generic Decoder} of \HS{} must learn to represent the body of the car without wheels, even though only about $10\%$ of the training shapes have part labels, which are noisy. Moreover, it has to adapt the car body to various wheel parameter values for shape edition, which \DeepS{} cannot do.

\begin{figure}
	\centering
	\small  
	\begin{tabular}{c|c}  
		\includegraphics[width=0.45\columnwidth]{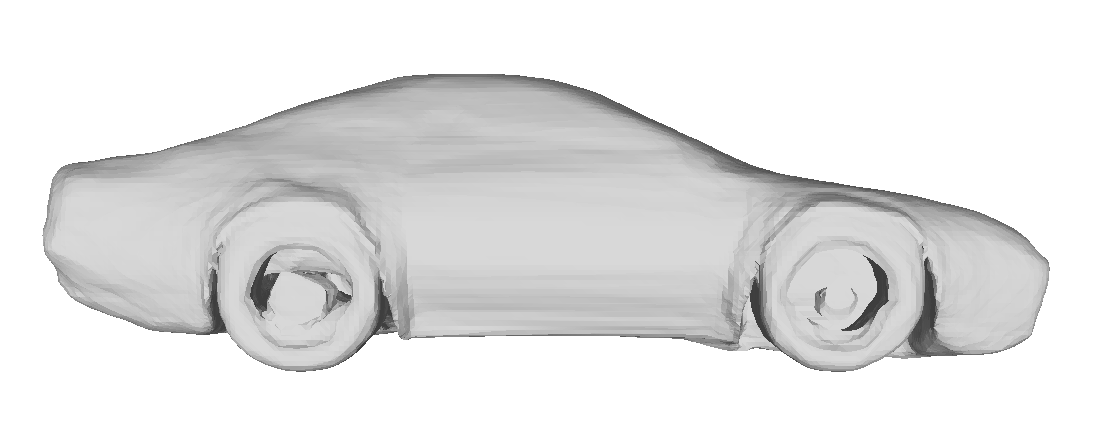} \hspace{0mm}&\hspace{0mm}
		\includegraphics[width=0.45\columnwidth]{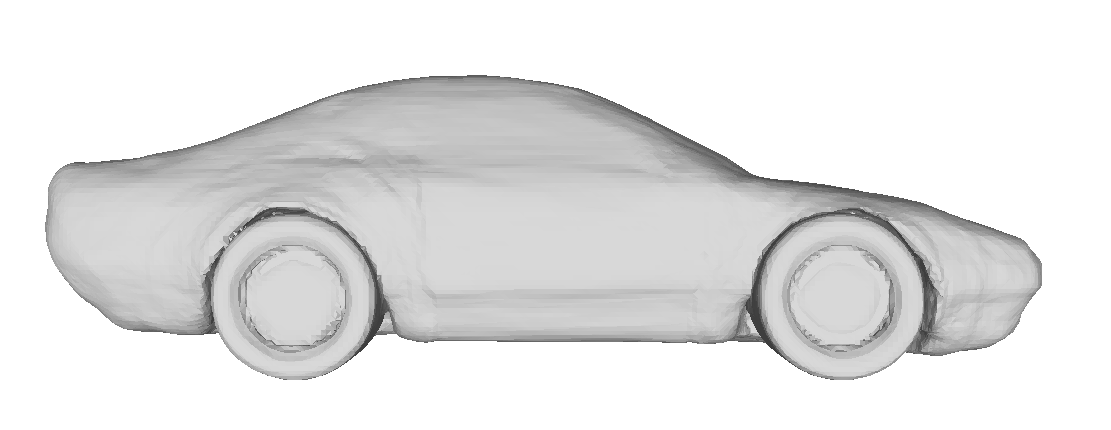} \\ 
		\includegraphics[width=0.45\columnwidth]{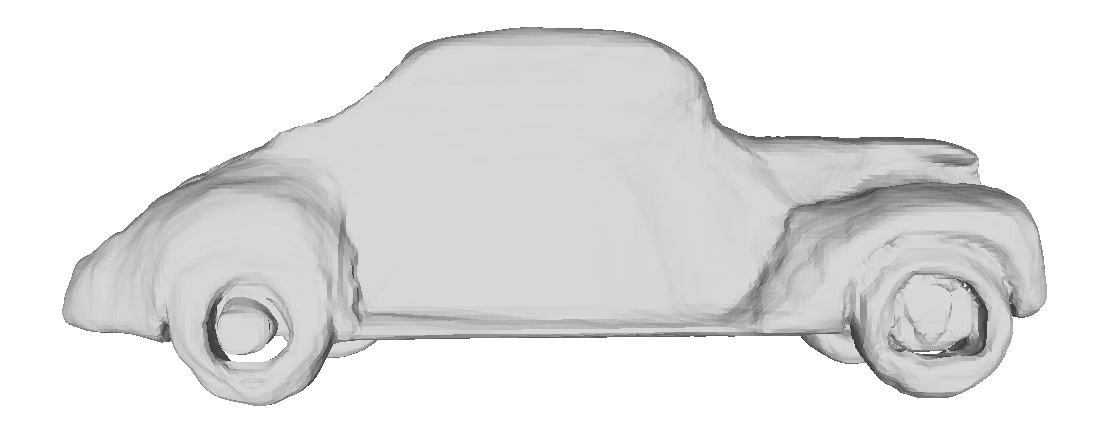} \hspace{0mm}&\hspace{0mm}
		\includegraphics[width=0.45\columnwidth]{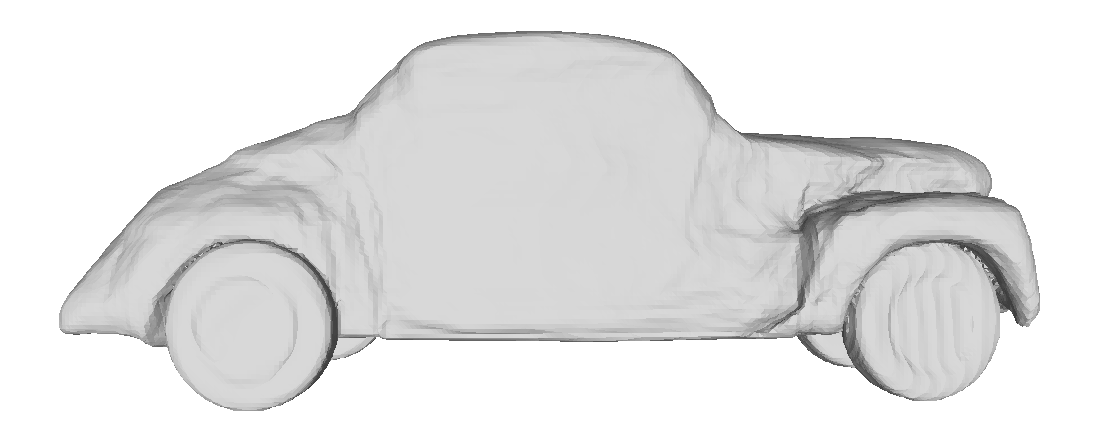} \\
		\DeepS{} & \HS{} \\
	\end{tabular}
	\caption{{ {\it Wheels in reconstructed test shapes.} Unlike some \DeepS{} reconstructions, the wheels produced by \HS{} are well defined, cylindrical, and separated from the car body.}}
	\label{fig:car_wheels}
\end{figure}

Therefore, this slight accuracy decrease we observe for cars represent the trade-off between reconstruction accuracy and the ability to manipulate shapes explicitly, as in many related works~\cite{Hao20, Smirnov20, Gadelha20}. Nonetheless, our representation yields an increase in quality and realism of the geometric and geometry-assisted parts. For cars, this translates to a better separation between body and wheels and to more cylindrical wheels thanks to geometry-assistance, as shown in Fig.~\ref{fig:car_wheels}, which also highlights that our method can exploit noisy labels generated by external segmentation pipelines. For the mixers, this manifests itself as the excellent surface regularity that the the geometrically assisted primitives provide. Furthermore, our approach allows parametric shape manipulation---\eg, changing wheel sizes or mixer radii---while incurring only minimal loss in accuracy in comparison to \DeepS{}, unlike related work such as \DualS{} that we discuss next.




\subsection{Parametric shape manipulation}
\label{sec:exp_manip}

A proxy representation was introduced in \DualS{}~\cite{Hao20} to offer editing abilities on top of \DeepS{}. However, it cannot easily be used to target specific parameter values, while our approach allows more accurate shape manipulation given inputs in the form of intepretable geometric parameters such as translation, rotation, and size. To demonstrate this, we manipulate the central tube of static mixers and compare against \DualS{}. For each shape in the training set, we randomly sample a target outer radius and thickness by perturbing the ground truth parameters. Then, the shape is automatically optimized to fit the new parameters.
For \DualS{}, we optimize a sphere from its coarse representation according to Equation (15) in~\cite{Hao20}. The sphere is automatically selected as the farthest one to the vertical axis, amongst those that are close to the middle horizontal plane.
Editing using \HS{} is performed by directly changing the values of the geometric parameters $\bS$, $\bR$, and $\bT$ and then decoding a new shape. It requires no optimization and is therefore both faster and exact.

The radius and thickness of the optimized tube are detected through a circle Hough transform on a horizontal cross-section of the SDF. We report the relative error to target parameters for the whole dataset in Table~\ref{table:mixer_manip}.
\DualS{} fails to manipulate the shapes to target specific parameter values, unlike our approach. This demonstrates that \HS{} is better suited for applications requiring more precise and explicit control, for example in engineering contexts.
Nonetheless, we note that \DualS{} and our approach can also be complementary. Indeed, they can be integrated to leverage their respective shape edition abilities, as we show in the Appendix~\refappendix{appendix:dualhs}{F}.


\setlength\mytabcolsep{\tabcolsep}
\setlength\tabcolsep{2pt}

\begin{table}
	\begin{center}
		{
			\begin{tabular}{@{}llcc@{}}
				\toprule
				 && \multicolumn{2}{c} {\textbf{Error [\%]}$\downarrow$}  \\
				 &&  Radius & Thickness \\ \midrule
				\DualS  && 12.7 & 37.8  \\
				\HS  && \textbf{0.4} &\textbf{ 4.5}  \\  
				\bottomrule
			\end{tabular}
		}
	\caption{\textit{Manipulation performance on the Mixers.} Numbers are the relative error to the target parameter values, averaged through the dataset. Note that \HS{} errors should be $0\%$ as the target parameters are directly used for reconstruction. Instead, they reflect the imprecision in our radius and thickness detection pipeline.}
\label{table:mixer_manip}
	\end{center}
\end{table}
\setlength{\tabcolsep}{\mytabcolsep}

In practice, the disentanglement of our approach enables a direct control of the geometric parameters, while independently maintaining or editing features from the implicit latent dimensions. This is illustrated in Fig.~\ref{fig:manip_all} where the type of helices and car body is maintained when manipulating the explicit parameters using \HS{}. 



\newlength{\manipheight}
\setlength{\manipheight}{2.6cm}
\newcommand{\intercellmanip}{\hspace{0mm}}

\setlength\mytabcolsep{\tabcolsep}
\setlength\tabcolsep{2pt}

\begin{figure*}[t]
\centering
\small  
\begin{tabular}{c|c|c}\intercellmanip
	\includegraphics[height=\manipheight]{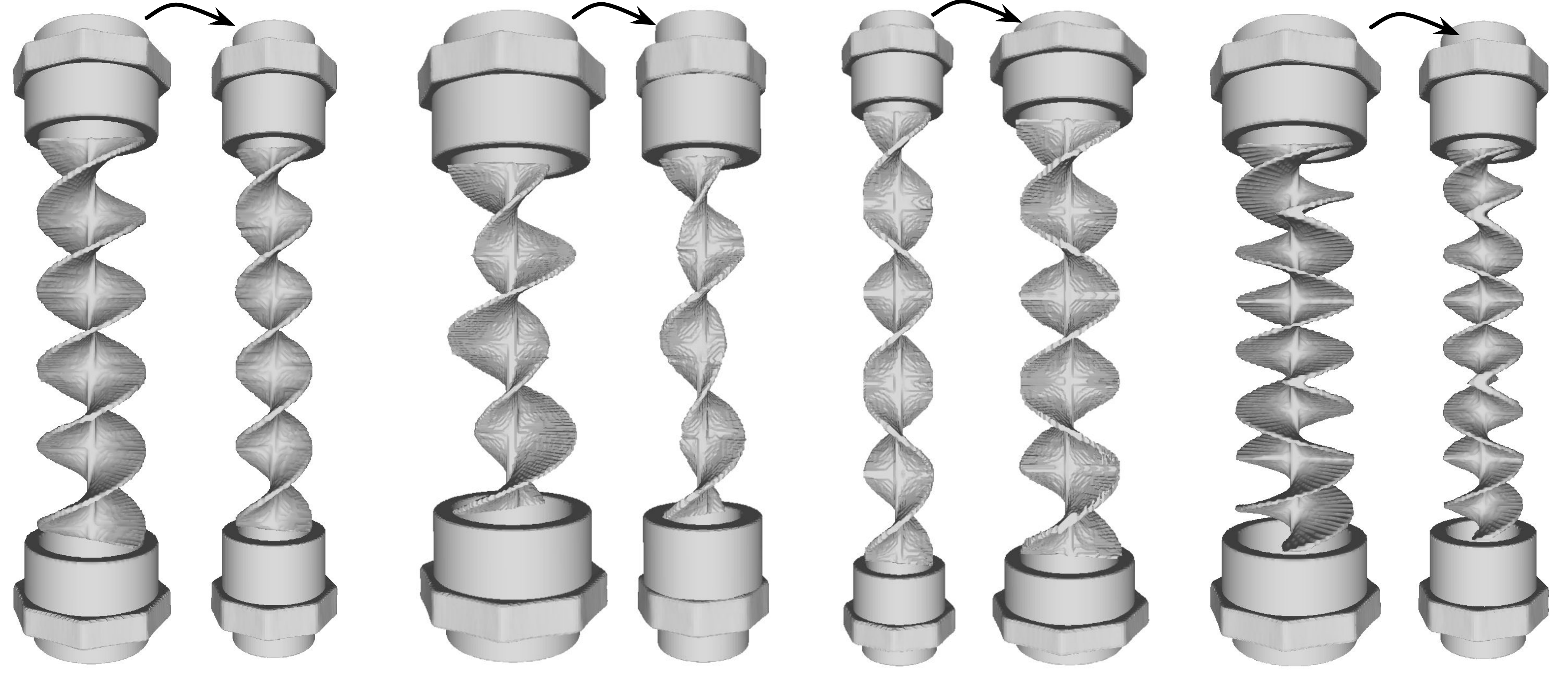} \intercellmanip&\intercellmanip
	\includegraphics[height=\manipheight]{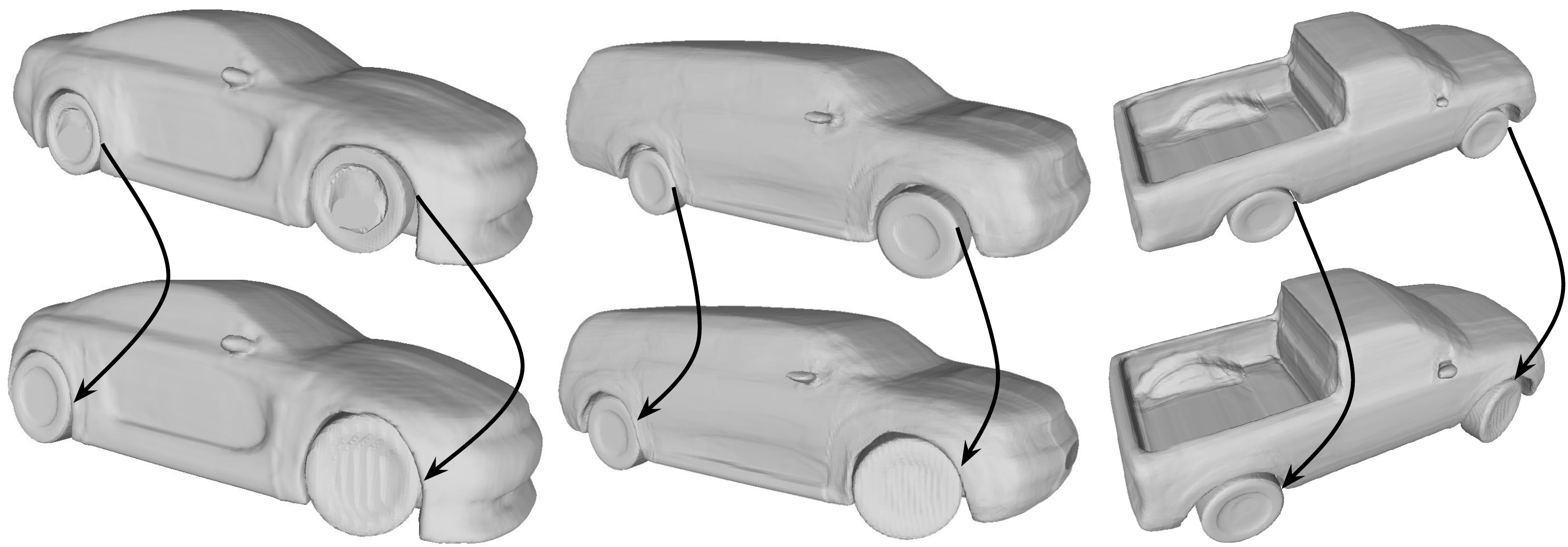} \intercellmanip&\intercellmanip
	\includegraphics[height=\manipheight]{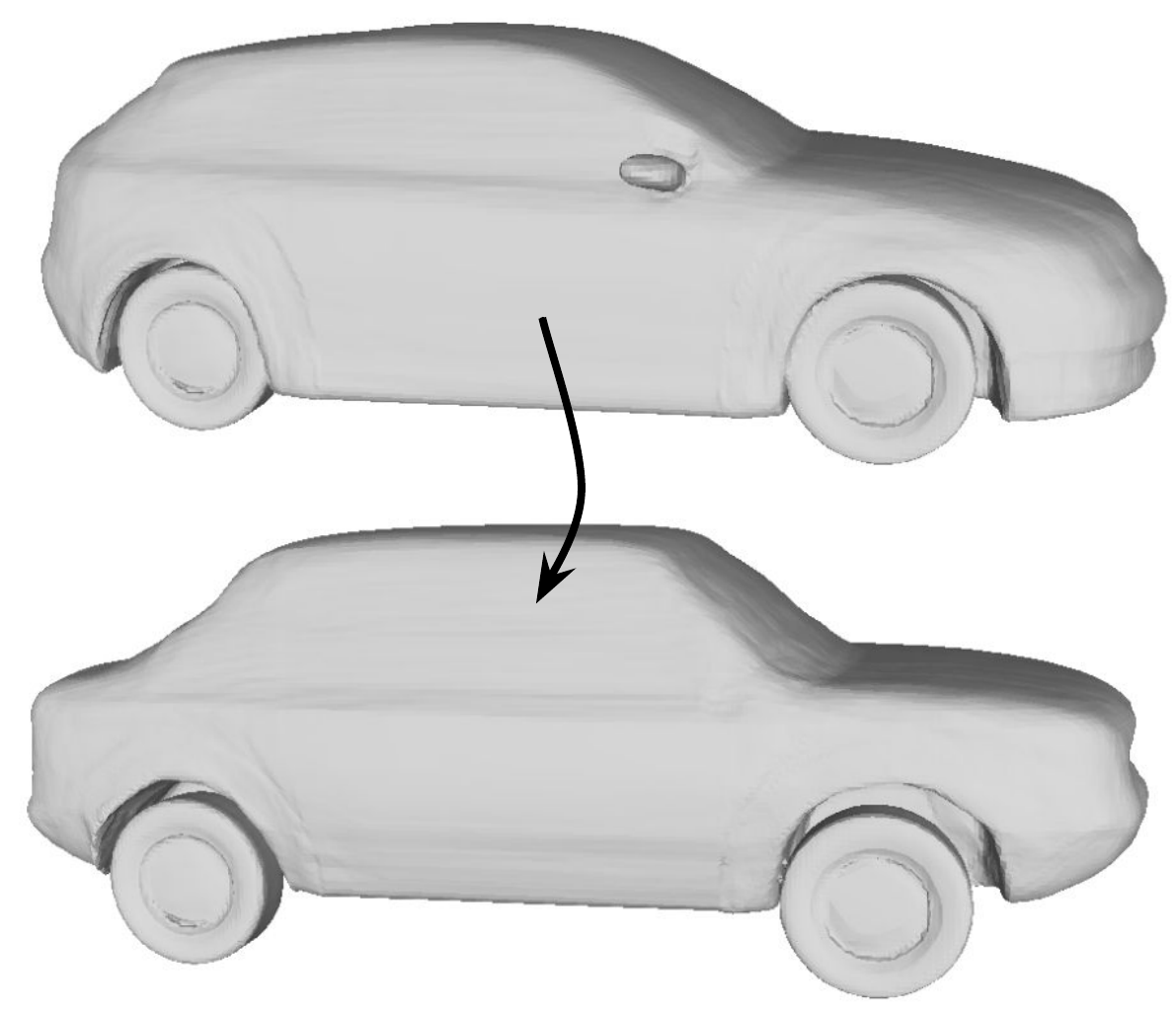} \intercellmanip\\
	\intercellmanip(a) \intercellmanip&\intercellmanip (b) \intercellmanip&\intercellmanip (c) \intercellmanip\\
\end{tabular}
\caption{ \textit{Parametric shape manipulation.} In each example, we show arrows going from the source object to the one obtained after manipulation using \HS{}. 
	(a) The geometric parameters $\bS$, $\bR$, and $\bT$ of the static mixers are modified. The ring and helix adapt while retaining their original key features such as their type and number of turns. 
	(b) The parameters of the wheels are edited and the car body changes accordingly: the wheel wells and the general size adapt (\eg, the third car widens) while the type of the car persists.
	(c) The latent of the car body $\LV_{\text{generic}}$ is changed, but the parameters of the wheels are kept constant. 
	A new car that fit the same wheels is obtained. For an equivalent experiment with the helices of the mixers, see Fig.~\ref{fig:sketch2mixer}.
}
	
\label{fig:manip_all}
\end{figure*}
\setlength{\tabcolsep}{\mytabcolsep}

\subsection{Single-image reconstruction}

We use the learned parameterized representation of the cars and mixers for single-view reconstruction. Similarly to~\cite{Chen19c}, we train \textit{encoders} to map an image space to \HS{}'s parameter space. This eventually permits our method to reconstruct the 3D shapes while automatically recovering some part geometric parameters.
For cars, we train an \textit{encoder} to predict all latent vectors and geometric parameters from renderings obtained from~\cite{Choy16}. 
We show test reconstruction results in Fig.~\ref{fig:image2car}. Not only does this yield realistic shapes but it also estimates the geometric parameters that characterize them, such as the radius of the wheels.
For mixers, only the outer parts are visible in sketches such as those of Fig.~\ref{fig:sketch2mixer}. We therefore train another \textit{encoder} to predict the parameters $\bS$, $\bR$, and $\bT$, and latent $\LV_\text{assit}$ from the sketches. One can then manually choose the central helix by supplying a latent vector $\LV_{\text{generic}}$, and it will directly fit to the tube length and width. This could lead to automatic retrieval of CAD objects from engineering drawings.


\newlength{\imagetocarfigwidth}
\setlength{\imagetocarfigwidth}{0.3\columnwidth}

\setlength\mytabcolsep{\tabcolsep}
\setlength\tabcolsep{2pt}

\begin{figure}
	\centering
	\small  
	\begin{tabular}{ccc}
		\includegraphics[width=\imagetocarfigwidth]{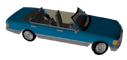} &
		\includegraphics[width=\imagetocarfigwidth]{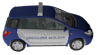} &
		\includegraphics[width=\imagetocarfigwidth]{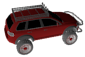} \\		
		\includegraphics[width=\imagetocarfigwidth]{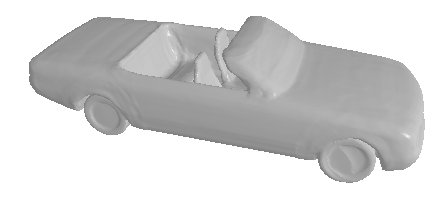} &
		\includegraphics[width=\imagetocarfigwidth]{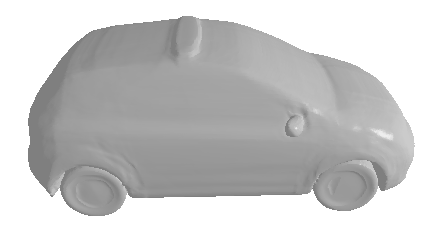} &
		\includegraphics[width=\imagetocarfigwidth]{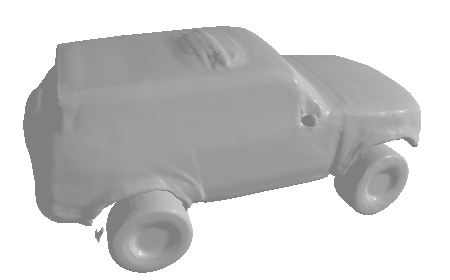} \\		
		$r_\text{ours}=0.114$ & $r_\text{ours}=0.143$ & $r_\text{ours}=0.172$ \\
		$r_\text{true}= 0.113$ & $r_\text{true}=0.146$ & $r_\text{true}=0.176$ \\
	\end{tabular}
	\caption{{\it Single-view reconstruction.} (\textit{Top row}) Input images. (\textit{Middle row}) Reconstructions using \HS{}. (\textit{Bottom row}) The estimated wheel radius $r_\text{ours}$ and the ground truth $r_\text{true}$, as measured on the ground truth meshes.}
	\label{fig:image2car}
\end{figure}
\setlength{\tabcolsep}{\mytabcolsep}

\newlength{\sketchtomixerfigwidth}
\setlength{\sketchtomixerfigwidth}{0.13\columnwidth}
\newcommand{\intracell}{\hspace{0mm}}
\newcommand{\intracellbis}{\hspace{-2mm}}

\setlength\mytabcolsep{\tabcolsep}
\setlength\tabcolsep{2pt}

\begin{figure}
	\centering
	\small  
	\begin{tabular}{cccccc}
		\includegraphics[width=\sketchtomixerfigwidth]{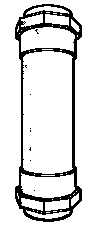} & \intracell
		\includegraphics[width=\sketchtomixerfigwidth]{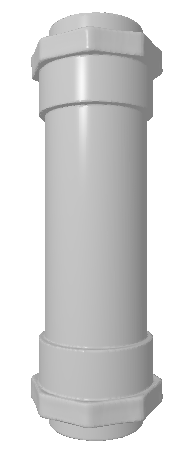} & \intracell
		\includegraphics[width=\sketchtomixerfigwidth]{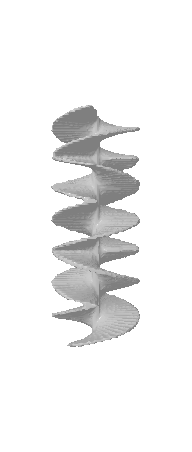} & \intracellbis
		\includegraphics[width=\sketchtomixerfigwidth]{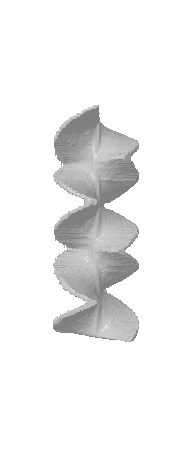} & \intracellbis
		\includegraphics[width=\sketchtomixerfigwidth]{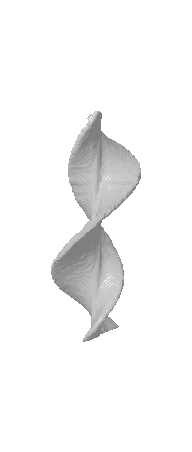} & \intracellbis
		\includegraphics[width=\sketchtomixerfigwidth]{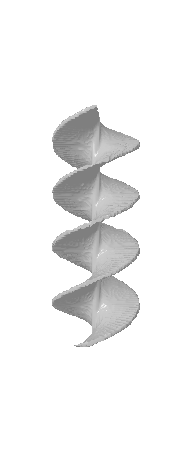} \\
		
		\includegraphics[width=\sketchtomixerfigwidth]{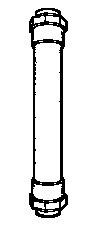} & \intracell
		\includegraphics[width=\sketchtomixerfigwidth]{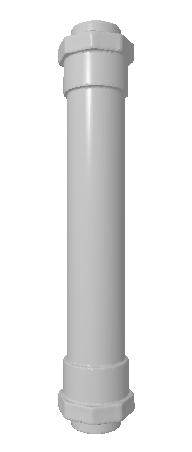} & \intracell
		\includegraphics[width=\sketchtomixerfigwidth]{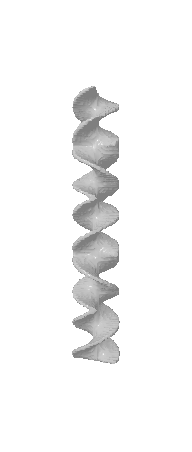} & \intracellbis
		\includegraphics[width=\sketchtomixerfigwidth]{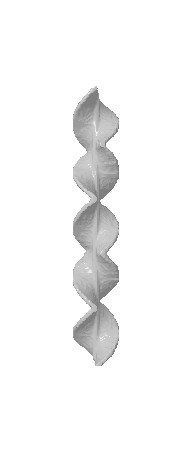} & \intracellbis
		\includegraphics[width=\sketchtomixerfigwidth]{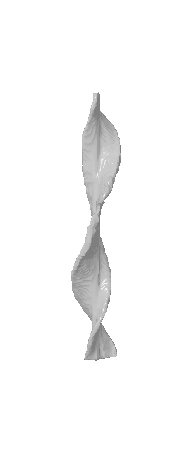} & \intracellbis
		\includegraphics[width=\sketchtomixerfigwidth]{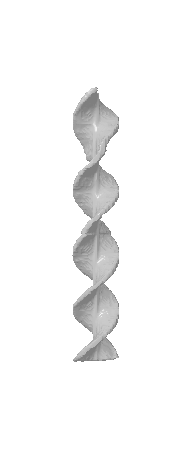} \\
		
	\end{tabular}
	\caption{{\it Mixer reconstruction from sketches. }(\textit{Left column}) Input sketches. (\textit{Second column}) Reconstructions using \HS{}. (\textit{Remaining columns}) Helices reconstructed by combining the predicted tube parameters with four latent vectors $\LV_{\text{generic}}$. The helices adapt to the tube size while preserving their features.}
	\label{fig:sketch2mixer}
\end{figure}
\setlength{\tabcolsep}{\mytabcolsep}

\section{Conclusion}

We have proposed a novel approach to combine deep implicit surfaces and geometric primitives for parametric manipulation that preserves representation accuracy, interpretability, and enforces consistency for complex objects made of multiple parts. Our disentangled representation makes it easy to define the geometric parameters of several parts and have the rest of the object adapt to these specifications, or to extract these parameters and reconstruct shapes from images and sketches.


In the current version of our architecture, using parameters significantly outside the training range can lead to failure, while consistency between parts is enforced by minimizing loss functions. In future work, we will look into improved generalization and imposing such consistency constraints as hard constraints~\cite{Marquez17} so that our loss have fewer terms. A further extension will include additional operations between parts, such as difference, similar to what is found in constructive solid geometry.

\paragraph{Acknowledgments}
This work was supported in part by the Swiss Innovation Agency. 

{\small
\bibliographystyle{ieee_fullname}
\bibliography{
	string, 
	misc, 
	vision,
	optim,
	biomed,
	cfd,
	graphics
}
}

\section*{Appendix}
\FloatBarrier
\appendix

\renewcommand{\thefigure}{S\arabic{figure}}
\renewcommand{\thetable}{S\arabic{table}}
\renewcommand{\theequation}{S.\arabic{equation}}
\setcounter{figure}{0}
\setcounter{table}{0}
\setcounter{equation}{0}

We first present the details of our data preparation in Appendix~\ref{appendix:data} and training and inference details in Appendix~\ref{appendix:train}. Then, we perform an ablation study on key training features employed in our approach in Appendix~\ref{appendix:ablation}. In Appendix~\ref{appendix:hss}, we explore a purely auto-decoder variant of our model where all parameters are predicted from a unique latent space. Following this, we report additional experimental results in Appendix~\ref{appendix:exp}. Eventually, we integrate our method to \DualS{}~\cite{Hao20} in Appendix~\ref{appendix:dualhs} to show that both of our and their manipulation tools can be combined.

\section{Data preparation}
\label{appendix:data}

We normalize our shapes to the centered sphere of radius $1/1.03$ and generate train and test SDF samples using the same sampling strategy as in~\cite{Park19c}. For the shapes that have part labels, we additionally compute the SDF at the same coordinates but with respect to the individual parts. For the \textit{Point Encoder}, we use 10000 points sampled from the surface of the full meshes as the input point clouds.

Regarding part labels for ShapeNet~\cite{Chang15} cars, we use the noisy labeled wheels as provided by~\cite{Kalogerakis17}. For the mixers, we have separately the labels of the main tube and the whole screws.
Moreover, the mixers come in different sizes but all have a central helix that comes in one of the three types depicted by Fig.~\ref{fig:mix_helix}.


\newcommand{\neghspace}{\hspace{-5mm}}


\begin{figure}[t]
  \centering
  \small  
  \begin{tabular}{ccc|ccc|ccc}
   					   \includegraphics[width=0.1\columnwidth]{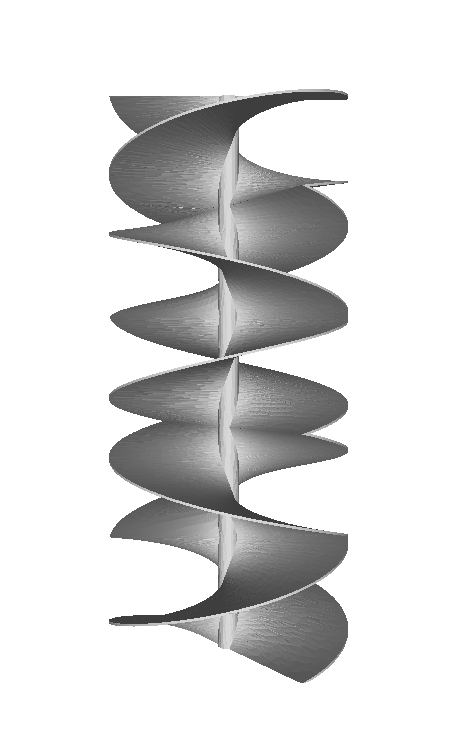}&
   \neghspace\includegraphics[width=0.1\columnwidth]{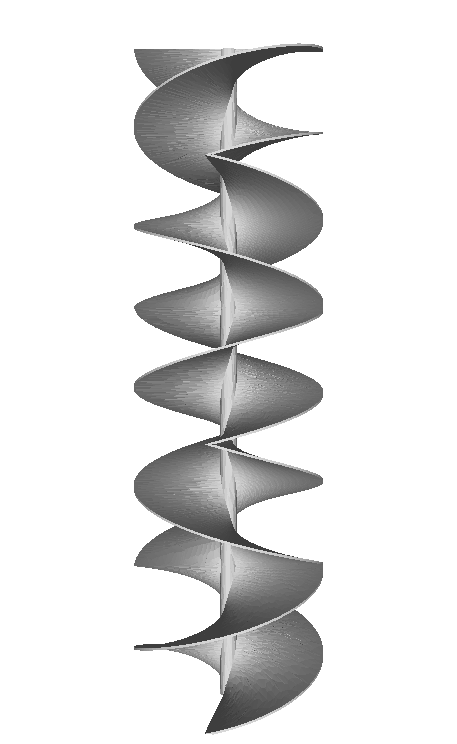}&
   \neghspace\includegraphics[width=0.1\columnwidth]{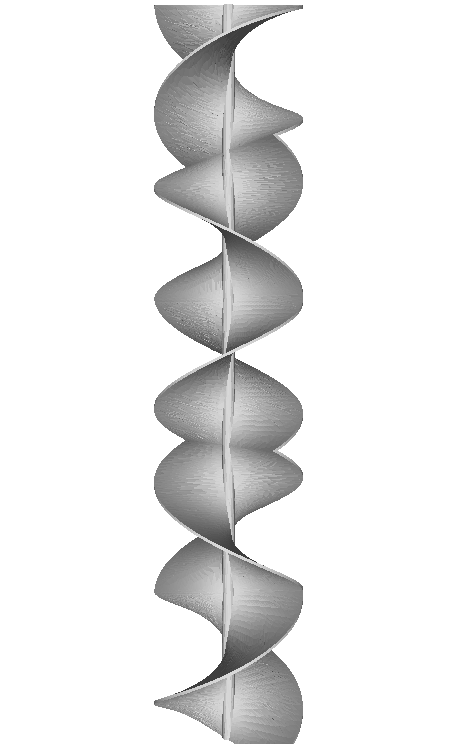}\hspace{3mm}&\hspace{3mm}
   \neghspace\includegraphics[width=0.1\columnwidth]{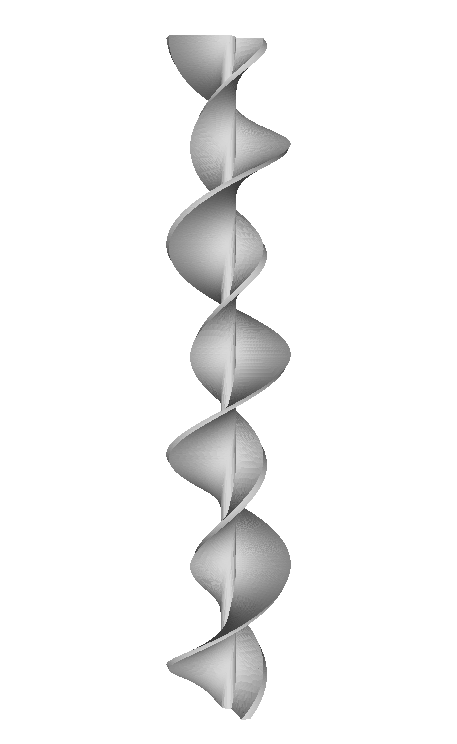}&
   \neghspace\includegraphics[width=0.1\columnwidth]{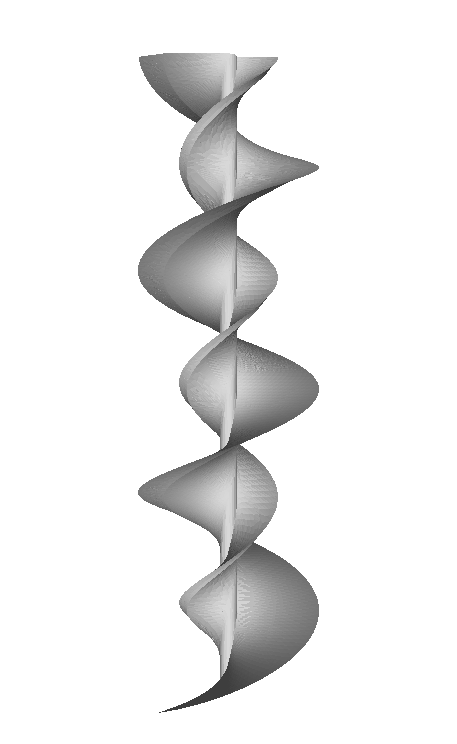}&
   \neghspace\includegraphics[width=0.1\columnwidth]{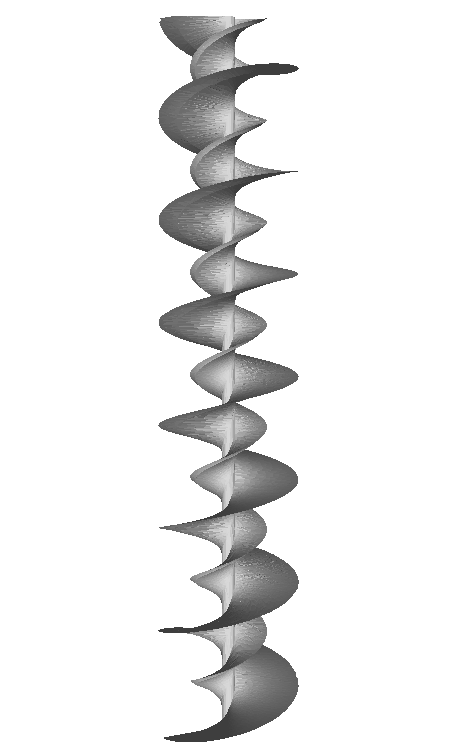}\hspace{3mm}&\hspace{3mm}
   \neghspace\includegraphics[width=0.1\columnwidth]{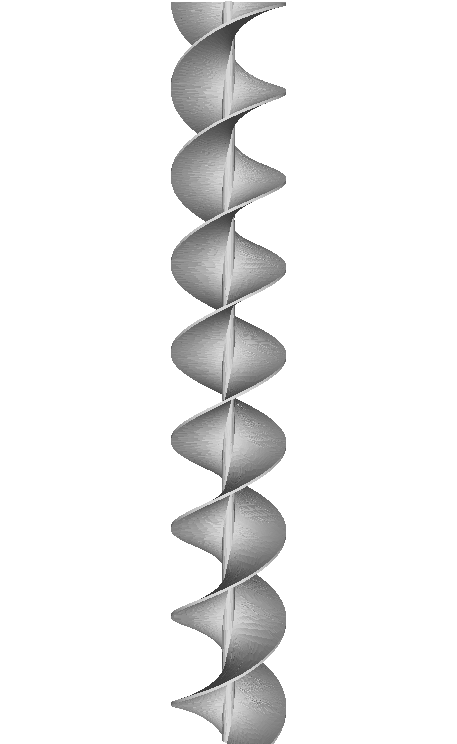}&
   \neghspace\includegraphics[width=0.1\columnwidth]{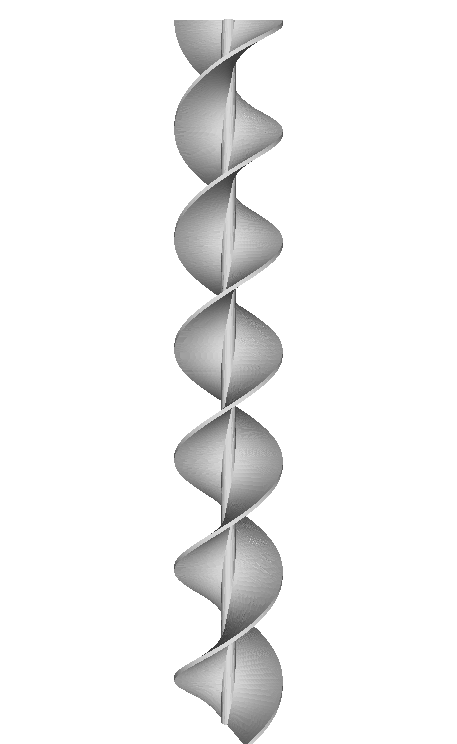}&
   \neghspace\includegraphics[width=0.1\columnwidth]{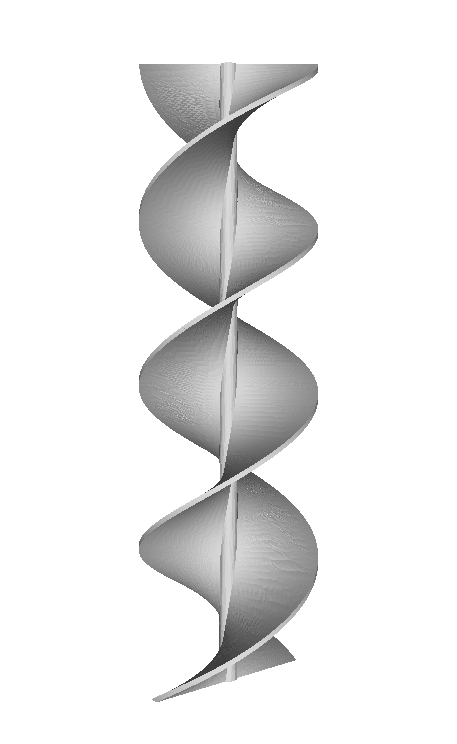}\\
   &\neghspace type 1 &&&\neghspace type 2 &&&\neghspace type 3 &
  \end{tabular}
  \caption{{\it Helix types.} {The static mixers have three types of helices, which can have different number of turns and  sizes.}}
  \label{fig:mix_helix}
\end{figure}

\section{Training and inference}
\label{appendix:train}

We trained our networks using the Adam optimizer~\cite{Kingma14a} with a learning rate of $5e^{-4}$ for the \textit{Generic Decoder}, $1e^{-3}$ for the latent vectors $\LV$ of the generic and geometry-assisted primitives, and $2e^{-4}$ for the {\it Point Encoder} and {\it Part Decoder}. The rest of Adam's parameter were set to their default values. We set the batch size to 32, use $K = 2000$ samples per shape at each iteration, and train all networks jointly for 2000 epochs. For the training loss, we use the empirically found weights $\lambda _{ga} = 0.1$,  $\lambda _{ic} = 5$, and  $\lambda _{reg} = 1e^{-4}$ with the unweighted reconstruction and consistency losses. The value of $\lambda _{ga}$ controls the level of similarity between the geometric primitives and the assisted ones. A high weight will force the geometry-assisted parts to become identical to the geometric primitives, while a very low value can allow the part-decoder to predict parts that are too different from the geometric ones. A very high value of $\lambda _{ic}$ can introduce undesired gaps between adjacent primitives whereas too low a value will fail to prevent overlaps. For $\lambda _{reg}$, we use the same values as in \DeepS{}.

During inference, the networks weights are fixed and only the latent vectors and explicit parameters are optimized with respect to the full reconstruction and regularization losses $\mathcal{L}_r^f$ and $\mathcal{L}_{reg}$, hence no part labels are needed during inference. The Adam optimizer is again used with learning rate $5e^{-3}$ for $\LV_{\text{generic}}$ and $5e^{-4}$ for $\bS$, $\bR$, $\bT$, and $\LV_{\text{assist}}$. We use batches of $K=8000$ samples and train for $800$ iterations. The \textit{Point Encoder} is used to generate the initial estimates for the parameters of the geometric and geometry-assisted primitives.


The input size of the \textit{Generic Decoder} is set to 256, as in~\cite{Park19c}. Therefore, we take the dimension of $\mathbf{LV}_{\text{generic}}$ to be $256$ minus the number of geometric parameters so that the full input is correctly of size $256$. We additionally set the individual latent vectors $\LV_{\text{assist}}^j$ to be of size $8$, again because the shapes produced by the \textit{Part Decoder} are assumed to be simpler. Similarly shaped parts, such as the wheels of the car and the top and bottom screw parts of the mixer of Fig.~1, can share these latent vectors.

\section{Ablation study}
\label{appendix:ablation}

In Section~3.3 of the main paper, we introduce several loss functions that we use to train our proposed approach. The role of $\mathcal{L}^{f}_{r}$ and $\mathcal{L}^{p}_{r}$ from Eqs.~6 and 7 is clear as they are reconstruction losses that force the predicted shapes to approximate correctly the ground-truth. The need for the regularization loss $\mathcal{L}_{reg}$ has been established in \DeepS{}. Here we show why the remaining loss terms, ($\mathcal{L}_{ga}$, $\mathcal{L}_{ic}$, and $\mathcal{L}_{cs}$), contribute to convergence towards useful solutions. 
We additionally show the necessity of the \textit{Point Encoder} to predict the geometric parameters. We have tried a fully auto-encoding approach where it was also predicting $\LV_{\text{generic}}$, but the resulting reconstruction accuracy was poor. It could be improved with a separate or more powerful encoder, but the work of~\cite{Park19c} already shows that an auto-decoding framework gives accurate results while simplifying the design and training of the model. Therefore, we chose to directly optimize for $\LV_{\text{generic}}$.

We sum up the ablation results in Table~\ref{table:abla_eval_supp}. We report the average $L_2$-Chamfer Distance (CD) over the test sets, as well as if the various primitives converge to their respective parts, and if the resulting model maintain its full parametric manipulation capabilities. As can be seen, only with all its components does \HS{} allow for effective shape manipulation at its best reconstruction accuracy, with the primitives correctly converging.


\newcommand{\cmark}{\color{green}\ding{51}}%
\newcommand{\xmark}{\color{red}\ding{55}}%
\newcommand{\nomark}{}%

\newlength{\mytabcolsepsupp}
\setlength\mytabcolsepsupp{\tabcolsep}
\setlength\tabcolsep{3pt}
\begin{table}
\begin{center}
{
\begin{tabular}{@{}lccccc@{}}
	\toprule
	  & \multicolumn{2}{c} {CD$\downarrow$} && Correct & Manip. \\
	 & \textbf{Mixers} & \textbf{Cars} && part & ability  \\ \midrule
	 no $L_r^p$  & 13.0 & 2.07 &&  \xmark & \xmark \\ 
	 no $L_{ga}$  & 12.6 & \textbf{1.57} &&  \cmark & \xmark \\ 
	 no $L_{ic}$  & 12.6 & 1.60 &&  \xmark & \cmark \\ 
	 no $L_{cs}$ & 13.1 & 1.68 &&  \cmark & \xmark \\ 
	 no \textit{PE} & 13.8 & 2.53 &&  \xmark & \xmark \\  
	 \midrule
	 \HS{} & {\bf 12.5} & {\bf  1.57} && \cmark & \cmark \\   
	 \bottomrule
\end{tabular}
}
\end{center}
\caption{\textit{Ablation of \HS{}.} We ablate our model and evaluate the reconstruction on the test shapes using average CD (30,000 points) multiplied by $10^4$. Additionally, we report whether the primitives correctly converge to their respective parts (\textit{Correct part}) and if the model retain its full manipulation capabilities (\textit{Manip.\ ability}). \textit{PE} corresponds to the \textit{Point Encoder}.}
\label{table:abla_eval_supp}
\end{table}
\setlength{\tabcolsep}{\mytabcolsepsupp}

\subsection{Geometry assistance ($\mathcal{L}_{ga}$)}

\begin{figure}
	\centering
	\small  
	\begin{tabular}{cc}
		\includegraphics[width=0.45\columnwidth]{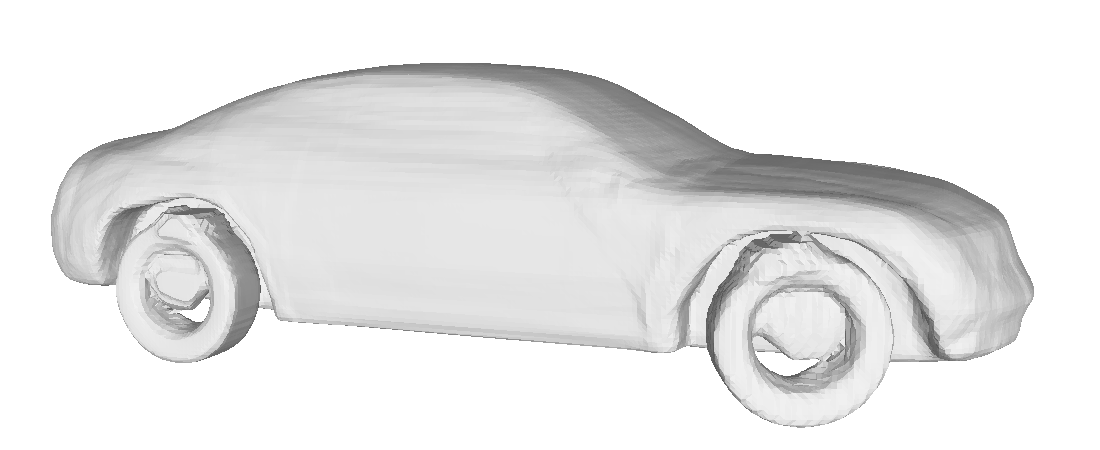}&
		\includegraphics[width=0.45\columnwidth]{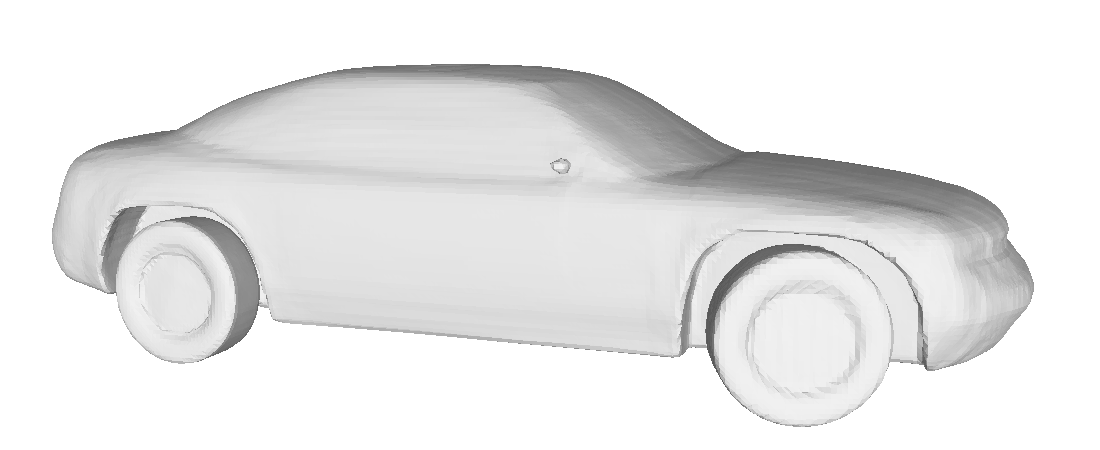} \\
	    Without $\mathcal{L}_{ga}$ & With $\mathcal{L}_{ga}$ \\
	\end{tabular}
	\caption{
			{\it Importance of  $\mathcal{L}_{ga}$ when training \HS{}.}
			Without the geometric-assistance during training, the wheels are not guaranteed to be nicely cylindrical. Here, they have learned spurious appendages. 
		}
	\label{fig:ga_ablation}
\end{figure}

The geometry assistance loss $\mathcal{L}_{ga}$ enforces the similarity between $\SDF_{\text{assist}}$ and its associated $\SDF_{\text{geom}}$. On the one hand, $\mathcal{L}_{ga}$ is essential for the convergence of $\SDF_{\text{assist}}$ to yield the desired parts, especially when there are only few shapes with part labels, otherwise the training can get derailed as it encourages $\SDF_{\text{assist}}$ to acquire shapes diverging from the geometry, as shown in Fig.~\ref{fig:ga_ablation}. 
On the other hand, it couples $\SDF_{\text{assist}}$ with the geometric parameters $\bS$ of the assisting geometry, which eventually allows for precise parametric manipulation. Indeed, without $\mathcal{L}_{ga}$ the shape parameters $\bS$ are meaningless for the geometry-assisted part and therefore cannot converge properly, nor be used for shape edition.


\subsection{Intersection constraints ($\mathcal{L}_{ic}$)}

The overlaps between primitives is prevented by minimizing $\mathcal{L}_{ic}$ of Eq.~9. To demonstrate the importance of this, we show in Fig.~\ref{fig:ab_inter} car reconstruction results from the test set when \HS{} is trained with and without $\mathcal{L}_{ic}$. This loss is key to establishing the natural gap, in the form of wheel wells, between the body of the car and its wheels. Removing it yields a significant overlap between the body and the wheels that may also lead to erroneous predictions. By contrast, enforcing the non-overlap constraint acts as a prior that helps the predicted wheels fit precisely the ground truth ones, thus establishing distinguishable wheel wells.


\newlength{\ablaicwidth}
\setlength{\ablaicwidth}{0.31\columnwidth}

\setlength\mytabcolsepsupp{\tabcolsep}
\setlength\tabcolsep{2pt}

\begin{figure}
  \centering
  \small  
\begin{tabular}{ccc}
	\includegraphics[width=\ablaicwidth]{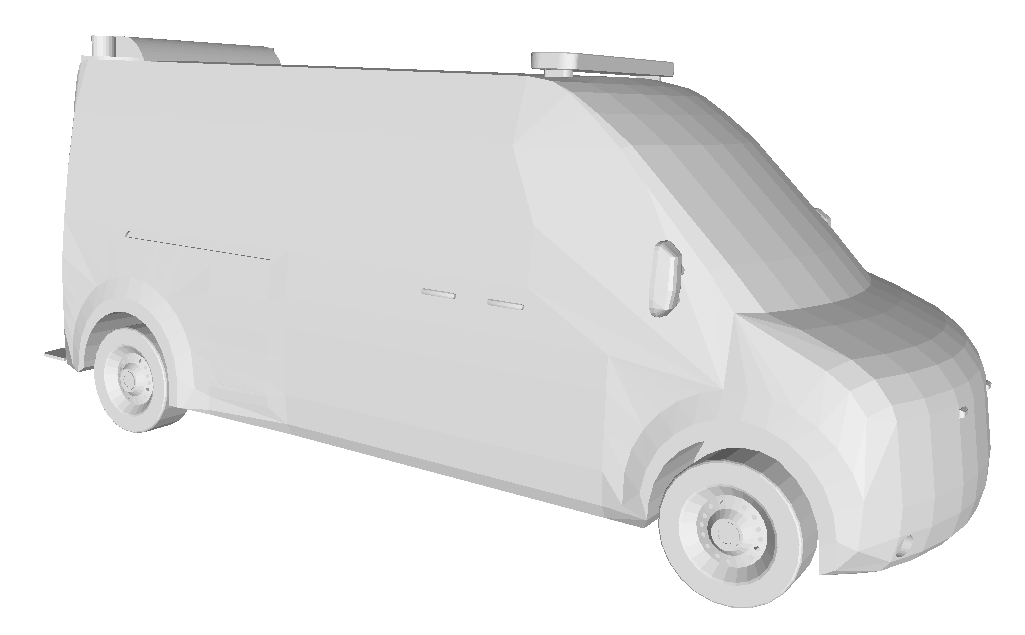}&
	\includegraphics[width=\ablaicwidth]{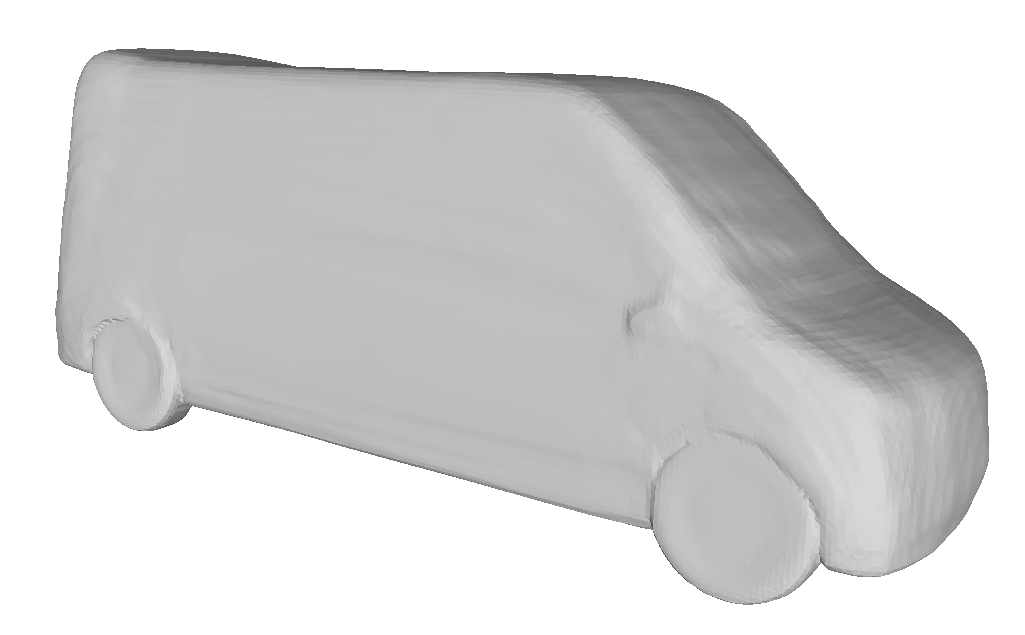}&
	\includegraphics[width=\ablaicwidth]{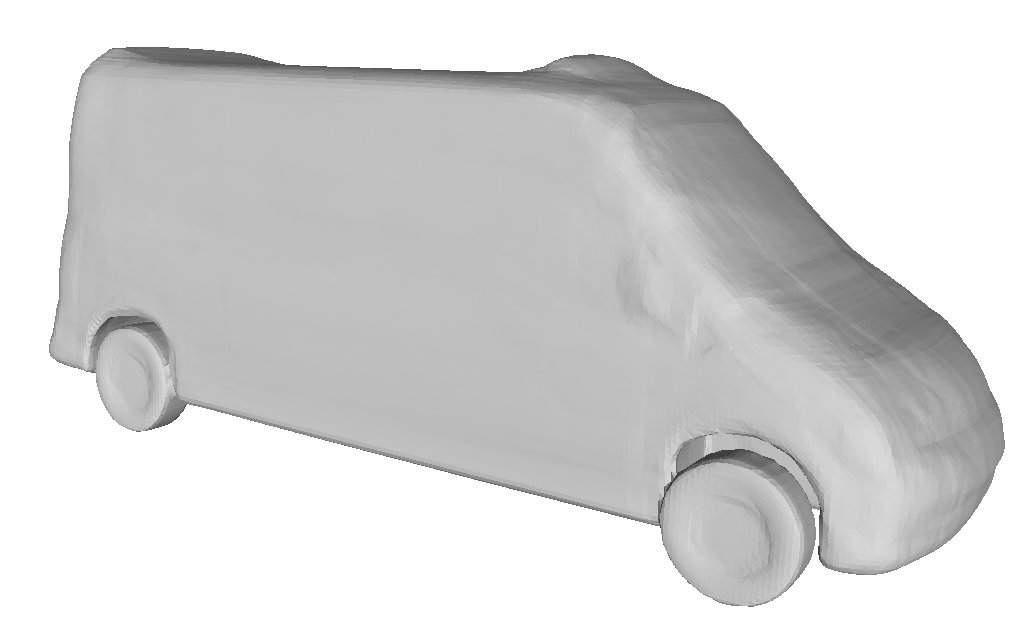}\\
	\includegraphics[width=\ablaicwidth]{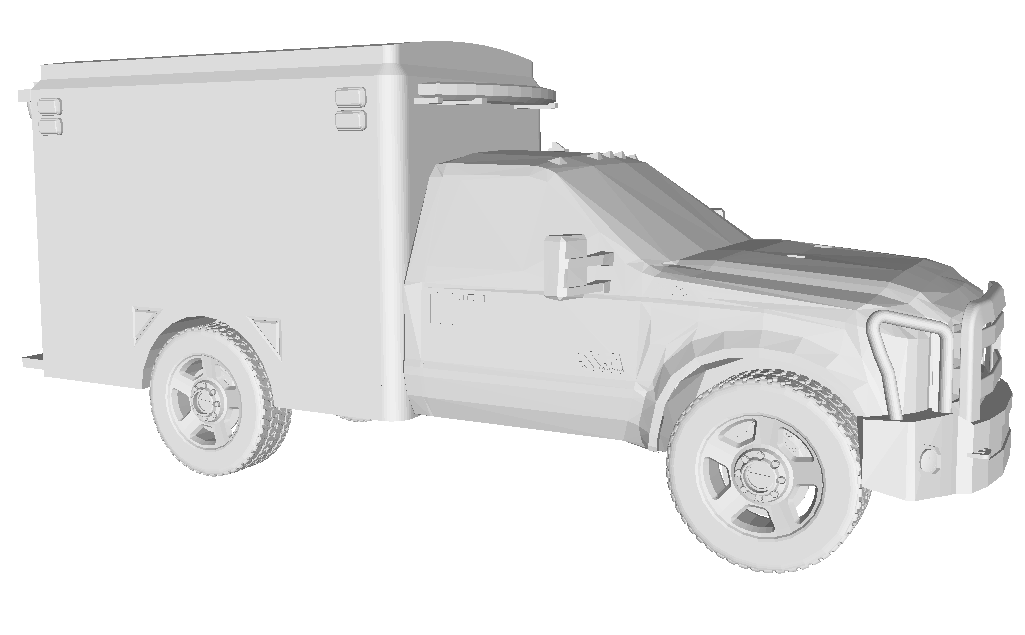}&
	\includegraphics[width=\ablaicwidth]{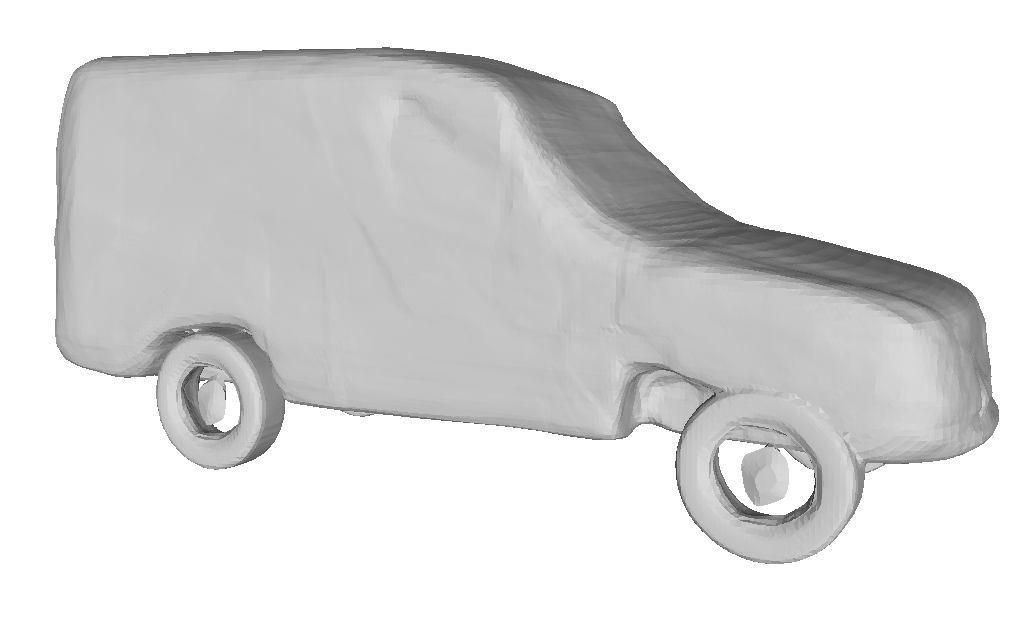}&
	\includegraphics[width=\ablaicwidth]{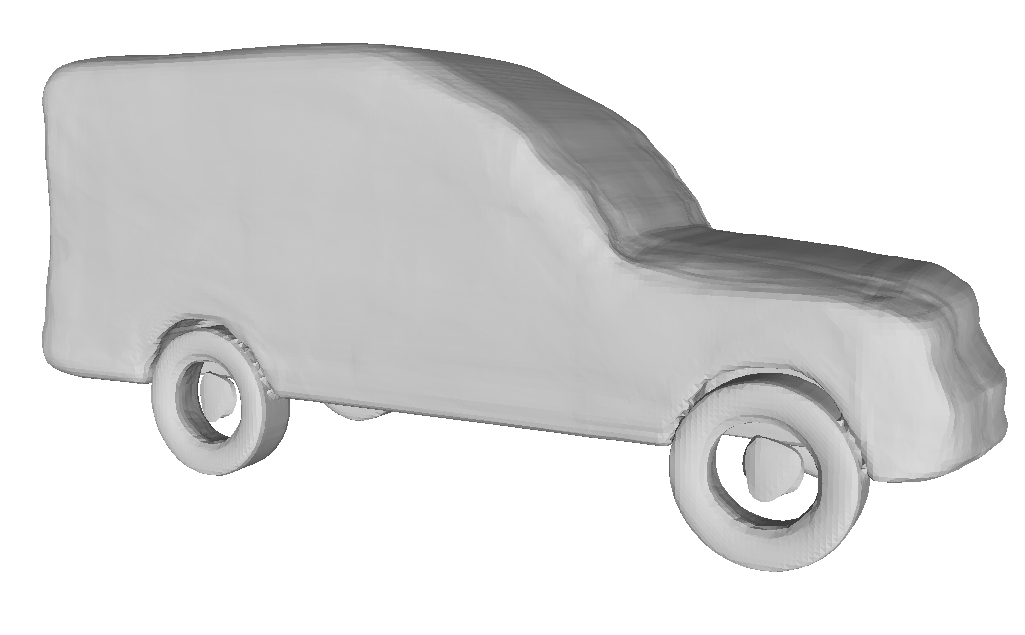}\\
	\includegraphics[width=\ablaicwidth]{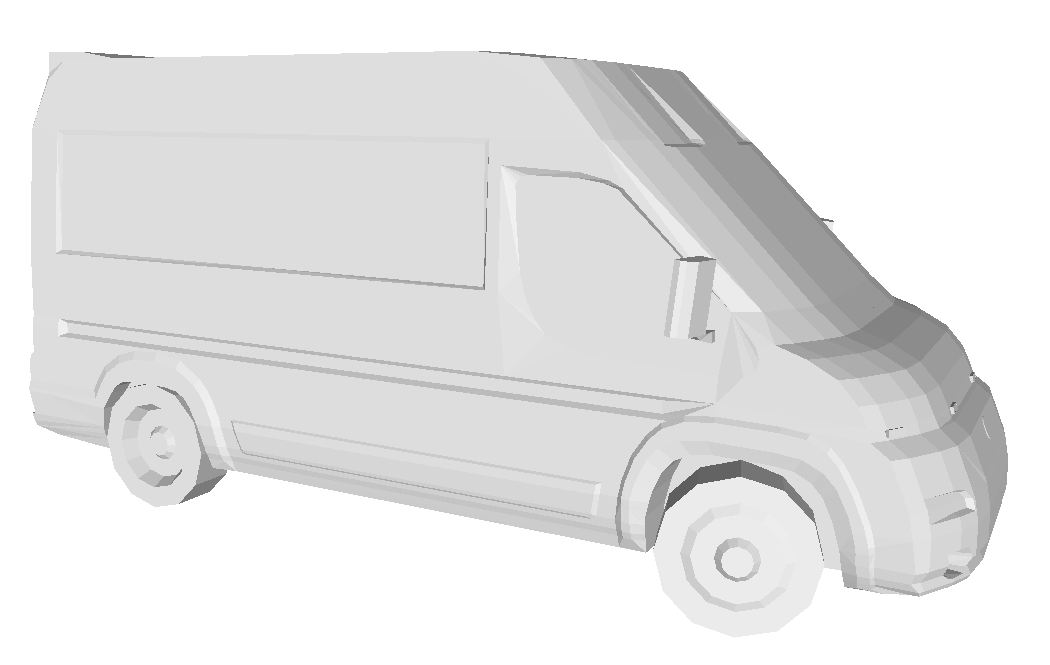}&
	\includegraphics[width=\ablaicwidth]{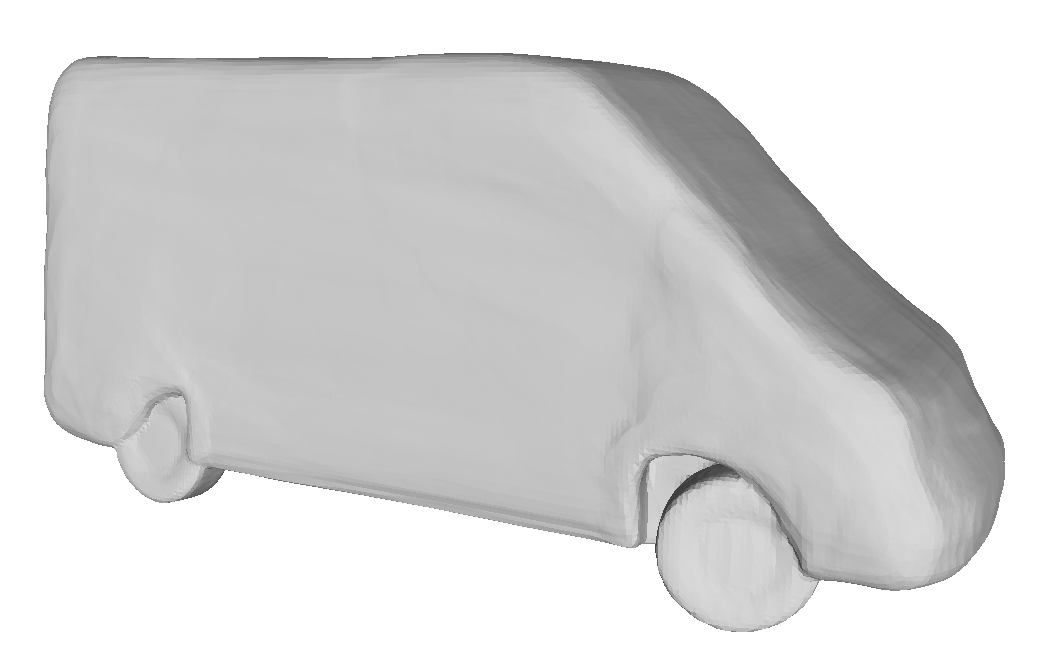}&
	\includegraphics[width=\ablaicwidth]{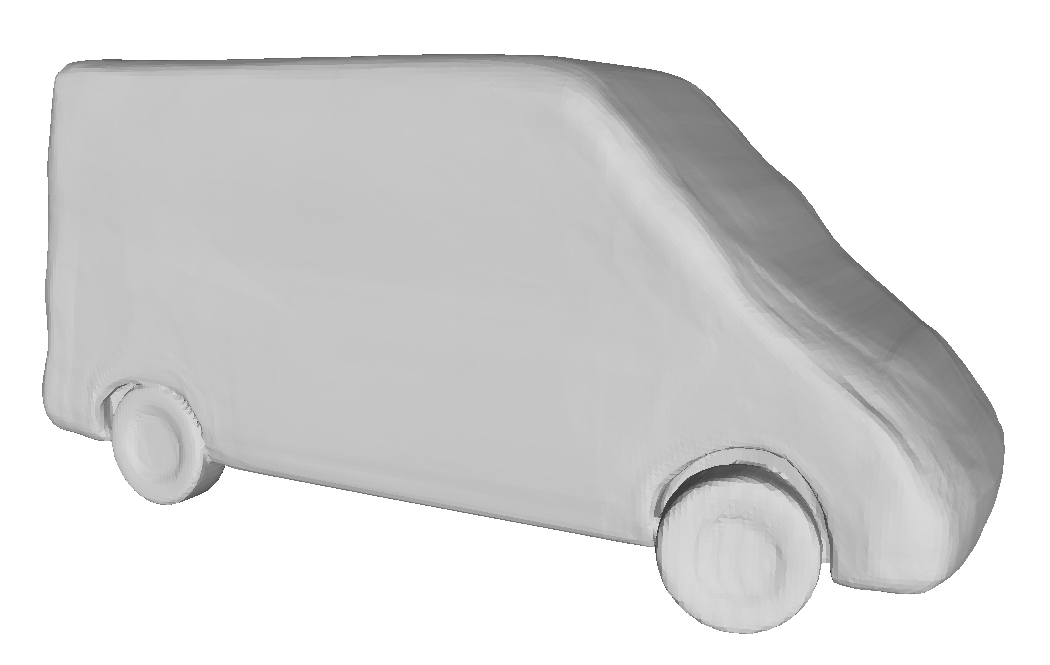}\\
    GT & Without $\mathcal{L}_{ic}$ & With $\mathcal{L}_{ic}$ \\
  \end{tabular}
  \caption{ {\it Importance of  $\mathcal{L}_{ic}$ when training \HS{}.} {Reconstruction examples using \HS{} trained without and with $\mathcal{L}_{ic}$. Enforcing non-overlap constraint by minimizing  $\mathcal{L}_{ic}$ results in much better defined wheel wells.}}
  \label{fig:ab_inter}
\end{figure}
\setlength{\tabcolsep}{\mytabcolsepsupp}

\subsection{Consistency of auxiliary output ($\mathcal{L}_{cs}$)}
\label{appendix:l_cs}

\begin{figure*}[t]
  \centering
  \small  
\begin{tabular}{c|c}
\includegraphics[width=0.45\textwidth]{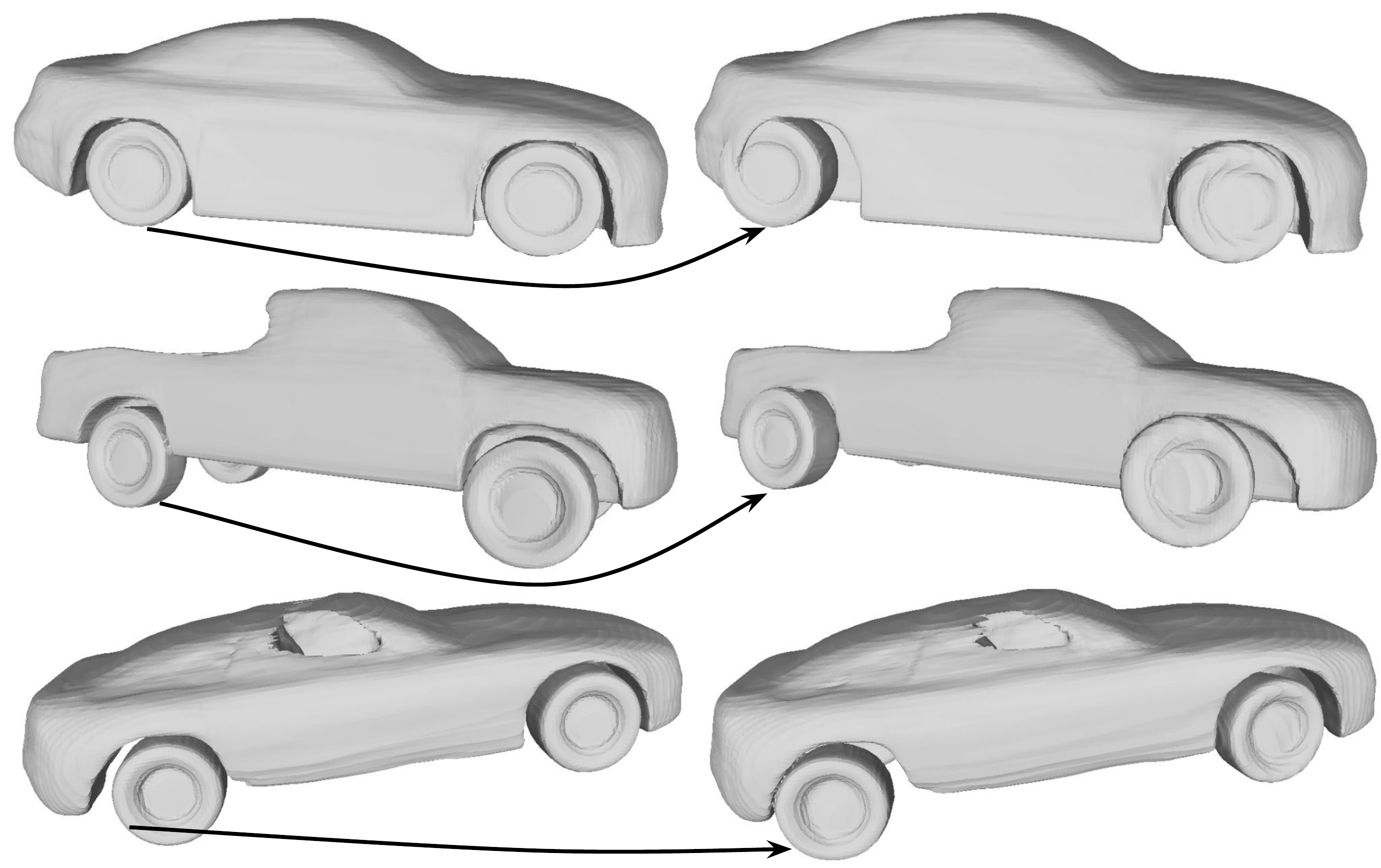}\hspace{1mm}& \hspace{1mm}
       \includegraphics[width=0.45\textwidth]{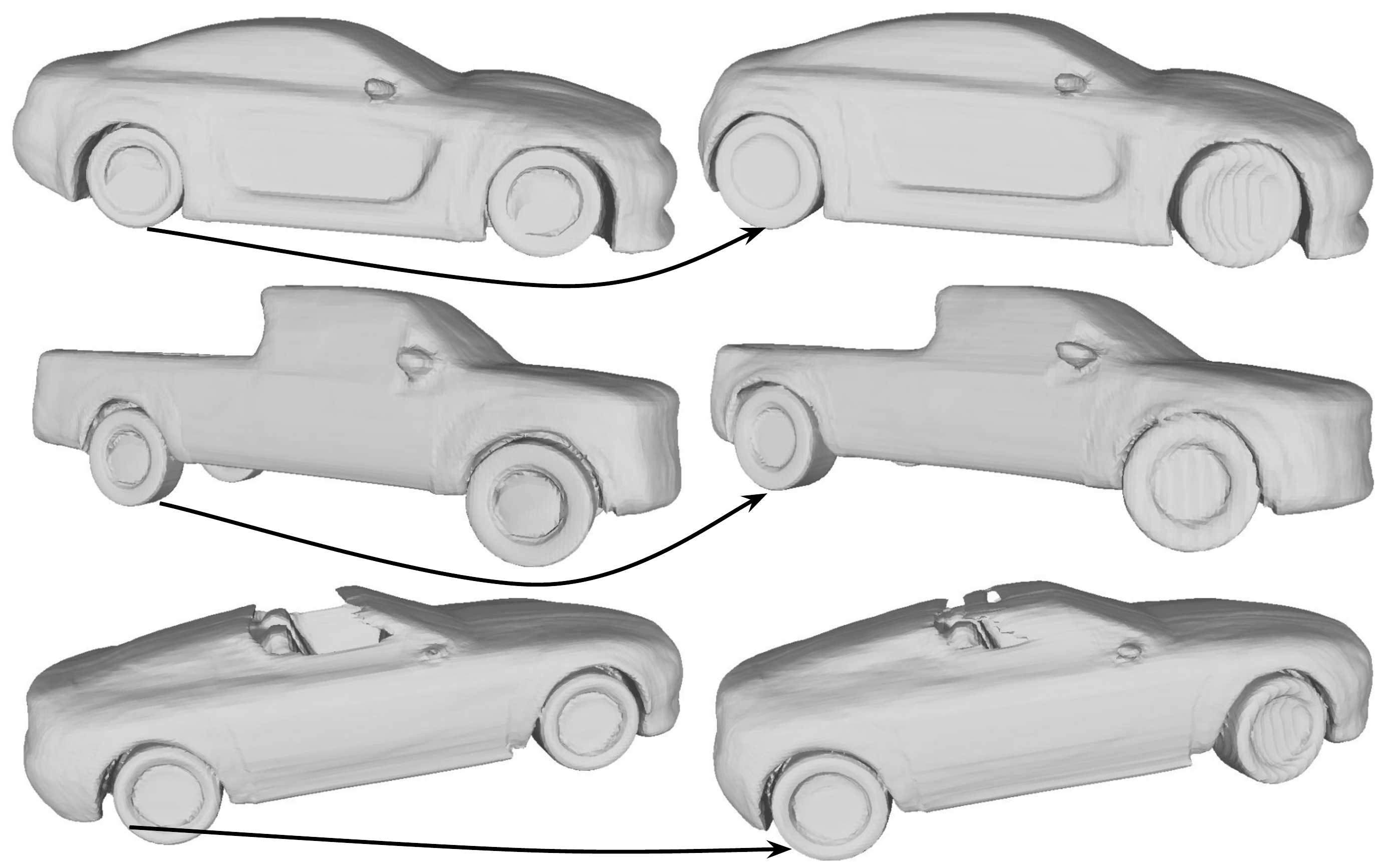}\\
       Without $\mathcal{L}_{cs}$  \hspace{1mm}&\hspace{1mm} With $\mathcal{L}_{cs}$       
  \end{tabular}
  \caption{ {\it Importance of $\mathcal{L}_{cs}$ when training \HS{}.} {Manipulation using models trained without and with $\mathcal{L}_{cs}$. For all pairs of cars connected by arrows, the wheels are moved or resized to edit the car on the left to the one on the right. Training without $\mathcal{L}_{cs}$ weakens the dependency of $\SDF_{\text{generic}}$ on the explicit geometric parameters $\bS$, $\bR$, and $\bT$, thereby creating a mismatch between the edited wheels and the car body. By contrast, minimizing $\mathcal{L}_{cs}$ during training ensures that the car body adapts correctly to changes in explicit wheel parameters at test time.}}
  \label{fig:car_aux_ablation}
\end{figure*}

Recall from Section~3.2 and Fig.~2 of the main text that the \textit{Generic Decoder} of \HS{} has an auxiliary output used to predict $\SDF_{\text{geom}}$. This forces the decoder to be consistent with the explicit geometric parameters by minimizing $\mathcal{L}_{cs}$ of Eq.~11 during training. Without this, the {\it Generic Decoder} tends to predict $\SDF_{\text{generic}}$ solely from $\LV_{\text{generic}}$, which weakens the dependency on the explicit parameters. In turn, this hampers shape manipulation, as shown in Fig.~\ref{fig:car_aux_ablation}. 


\subsection{Point encoder}
\label{appendix:pe}

\begin{figure*}[t]
  \centering
  \small  
\begin{tabular}{c|c}
	\begin{tabular}{cc}		
		\includegraphics[width=0.2\textwidth]{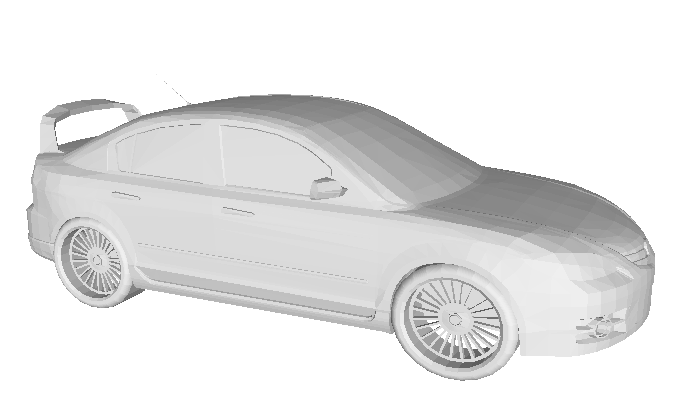} &
		\includegraphics[width=0.2\textwidth]{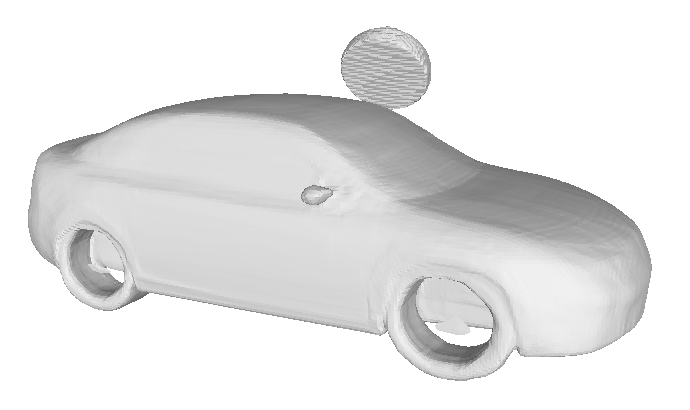} \\
		(a) & (b) \\
		\includegraphics[width=0.2\textwidth]{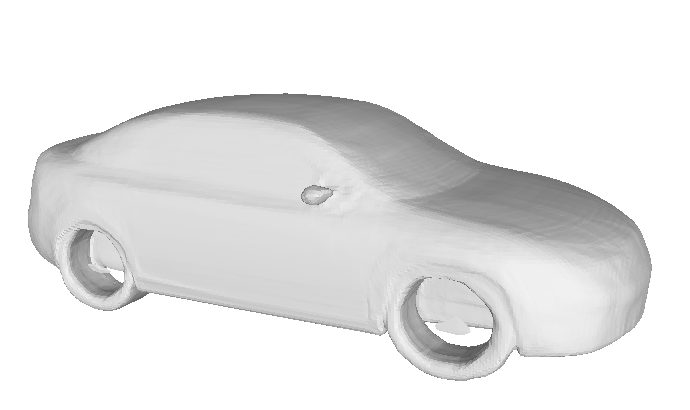} &
		\includegraphics[width=0.2\textwidth]{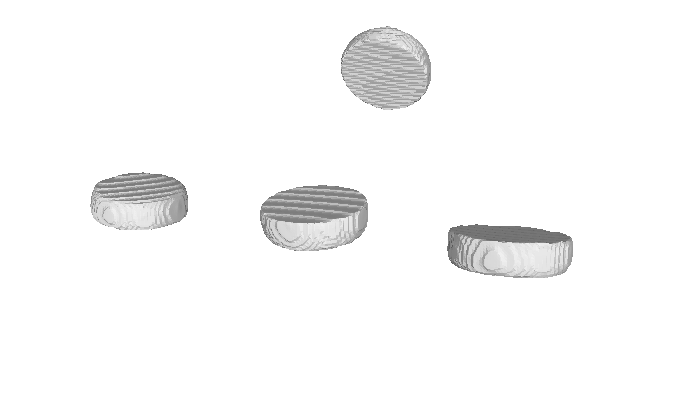} \\
		(c) & (d) \\
	\end{tabular}
	&
	\begin{tabular}{ccc}
		\includegraphics[width=0.1\textwidth]{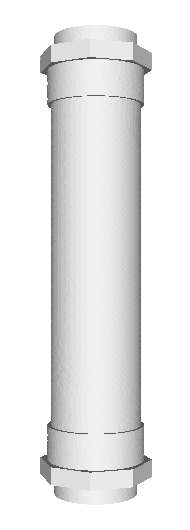} &
		\includegraphics[width=0.1\textwidth]{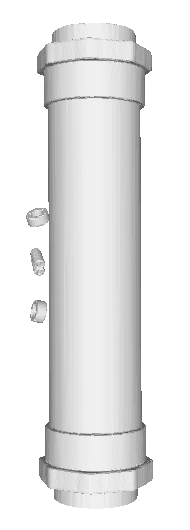} &
		\includegraphics[width=0.1\textwidth]{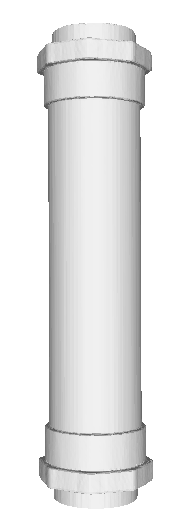} \\
		(e) & (f) & (g) \\
	\end{tabular}
	\\
  \end{tabular}
  \caption{ {\it Importance of the \emph{Point Encoder} when training \HS{}.} Reconstruction examples using \HS{} trained without a \textit{Point Encoder} on a training car (a-d) and mixer (e-g) for which part labels were not available. (a) Ground-truth car, (b) Full reconstruction, (c) Generic-primitive, and (d) Wheels. (e) Ground-truth mixer, (f) Full Reconstruction, and (g) Generic-primitive. The wheels and mixer parts do not converge correctly, leading the \textit{Generic Decoder} to compensate by reconstructing the full shape.}
  \label{fig:pe_ab}
\end{figure*}

As introduced in Section~3.2 and Fig.~2, we use a \textit{Point Encoder} to learn the explicit geometric parameters and latent vectors $\LV_{\text{assist}}$.
Indeed, if we were to optimize the $\bS$, $\bR$, and $\bT$ parameters of the geometric primitives in an auto-decoding fashion, they would receive only one update per epoch and quickly become inconsistent with the generic primitives, especially for shape without part labels. This is depicted by Fig.~\ref{fig:pe_ab} where the car wheels and mixer parts do not converge to their correct locations without the \textit{Point Encoder}, causing the generic-primitive to compensate by also predicting them.


\section{Shared latent space \HSs{}}
\label{appendix:hss}

In the main text, we have proposed \HS{} whose latent space can be considered disentangled as the parameters to the various primitives correspond to separate dimensions. Here, we present a purely auto-decoding variant where all parameters are instead decoded from a unique---or shared---latent vector, as in~\cite{Park19c, Hao20}. Hence, we dubbed this version \HSs{}, for a \textit{shared} latent space.
In practice, this model corresponds to different manipulation use cases. A shared latent space is useful for optimizing and adapting the whole shape to a subset of new parameters whereas the disentanglement of \HS{} enables a direct and precise control of all of them, while independently maintaining or editing features from the implicit latent dimensions. We expand on this below and offer reconstruction and manipulation results in Appendix~\ref{appendix:exp}.

\subsection{Network}

\begin{figure*}[t]
  \centering
  \small  
  \includegraphics[width=0.6\textwidth, trim={0cm 11cm 11.5cm 0cm}, clip]{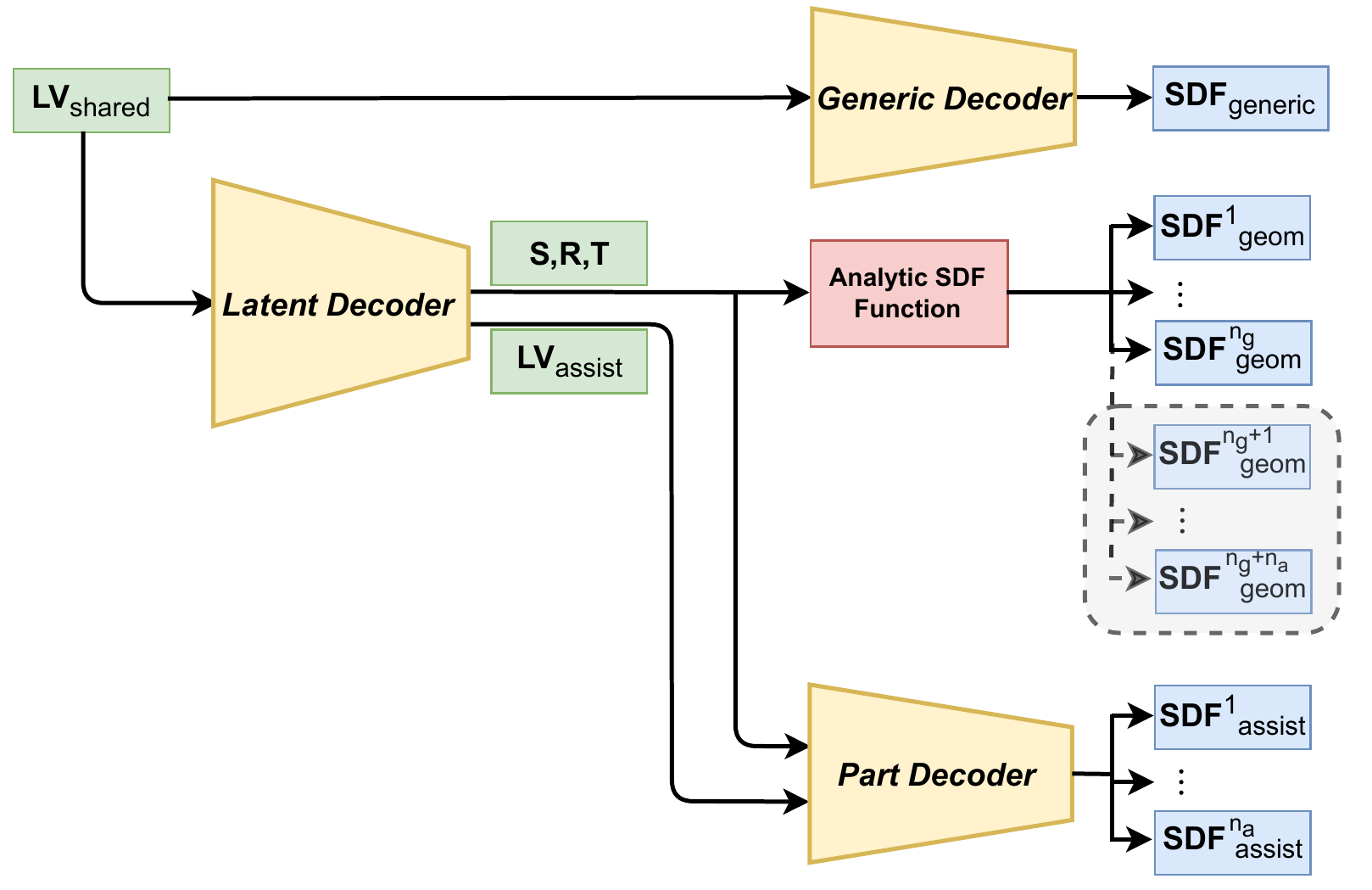}
  \caption{\textit{\HSs{} architecture.} The model takes a single latent vector $\LV_{\text{shared}}$ as input. It is decoded into an  $\SDF_{\text{generic}}$ and geometric parameters $\bS$, $\bR$, and $\bT$, along with auxiliary latent vectors $\LV_{\text{assist}}$, which are themselves decoded into the $\SDF_{\text{geom}}$  and $\SDF_{\text{assist}}$. The grey box denote components used only for training purposes.}
  \label{fig:arch_supp}
\end{figure*}

In \HSs{}, all parameters are encoded in a unique latent space, as shown in Fig.~\ref{fig:arch_supp}. While the \textit{Generic Decoder} takes the full $\LV_{\text{shared}}$ as input, we use a new \textit{Latent Decoder} to predict from $\LV_{\text{shared}}$ the explicit parameters $\bS$, $\bR$, and $\bT$, along with the auxiliary latent vectors $\LV_{\text{assist}}$. This network comprises a single fully connected layer that extracts primitive-specific features, followed by multiple independent branches of fully connected layers and ReLU non-linearities.


While the rest of the shape prediction is similar, there are two other main differences. Firstly, we don't need a \textit{Point Encoder} as the parameters are directly decoded from $\LV_{\text{shared}}$. Second, as the geometric parameters $\bS$, $\bR$, and $\bT$ are embedded within the shared latent space, there is no need to give them explicitly as input to the \textit{Generic Decoder}, nor to have the auxiliary output $\SDF_{\text{geom}^\dagger}$. In consequence, \HSs{} does not need to be trained with the consistency loss $\mathcal{L}_{cs}$ from Eq.~11, yet is able retain consistency between its primitives as shown on the manipulation examples of Fig.~\ref{fig:manip_all_supp}.

\subsection{Training and inference details}

The training and inference are performed similarly to \HS{}, with a learning rate of $5e^{-3}$ for $\LV_{\text{shared}}$ and $2e^{-4}$ for the \textit{Latent Decoder}.
Moreover, we take the latent vector $\LV_{\text{shared}}$ to be of size 256, and the individual latent vectors $\LV_{\text{assist}}^j$ to be of size $8$, as before.

\subsection{Manipulation}

For \HSs{}, parametric manipulation of the shape is achieved by optimizing the latent vector $\LV_{\text{shared}}$ to decode the target parameters. This corresponds to the following minimization problem:
\begin{equation}
	\min \limits_{\LV_{\text{shared}}} \Big\lVert \mathcal{D}_l(\LV_{\text{shared}})^{ij} - \mathbf{V}^{*} \Big\rVert_{1} + \lambda_{reg}\mathcal{L}_{reg}\, ,
	\label{eq:shared_man}
\end{equation}
where $\mathcal{D}_l$ refers to the {\it Latent Decoder}, $ij$ indexes the parameters to manipulate, $\mathbf{V}^{*}$ are the target values for those parameters, and $\mathcal{L}_{reg}$ is the latent vector regularization loss as described in Section~3.3 of the main text.

As explained before, a shared latent space is useful to edit a subset of the parameters to manipulate the entire shape in a combined way, or simply to use our hybrid 3D shape representation without having to manually handle the geometric parameters.
This is illustrated in Fig.~\ref{fig:manip_all_supp} where few parameter edits may lead to new shapes with \HSs{} while the type of helices and car body is maintained when using \HS{}.

\section{Additional experimental results}
\label{appendix:exp}

In this section, we provide additional examples for reconstruction and manipulation tasks. We also report results with the \HSs{} variant described in Appendix~\ref{appendix:hss}, manipulation results on a \textit{Chair} dataset using \HS{}, and show an application of drag optimization for car shapes.

\subsection{Shape reconstruction}
\label{appendix:recon}

In Figs.~\ref{fig:car_recon_supple} and~\ref{fig:mixer_recon_supple}, we provide more car and mixer reconstruction examples, using \HS{} and \HSs{}. 


\newlength{\reconfigwidthsupp}
\setlength{\reconfigwidthsupp}{0.115\textwidth}

\setlength\mytabcolsepsupp{\tabcolsep}
\setlength\tabcolsep{2pt}

\begin{figure*}[t]
  \centering
  \small  
\begin{tabular}{cccccccc}
   \includegraphics[width=\reconfigwidthsupp]{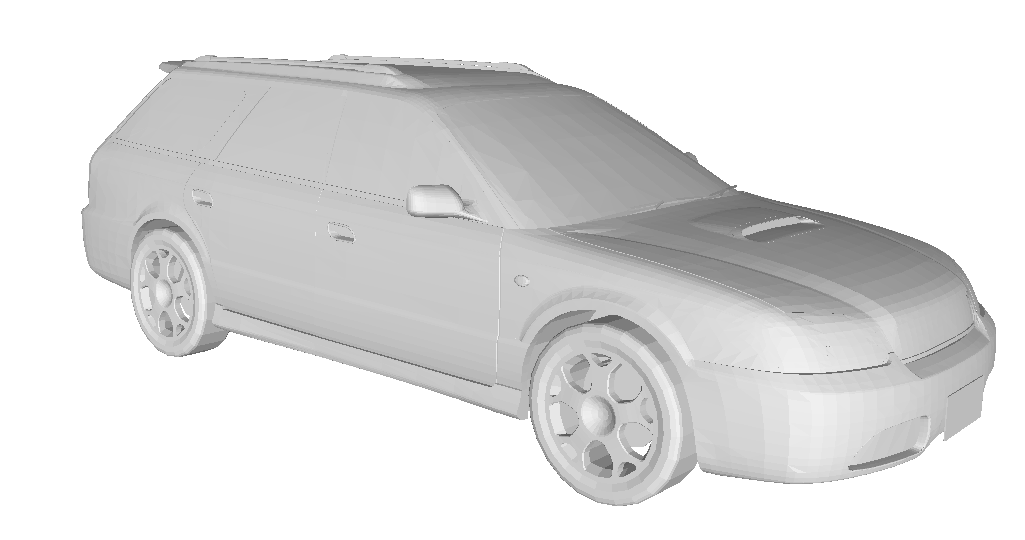} &
   \includegraphics[width=\reconfigwidthsupp]{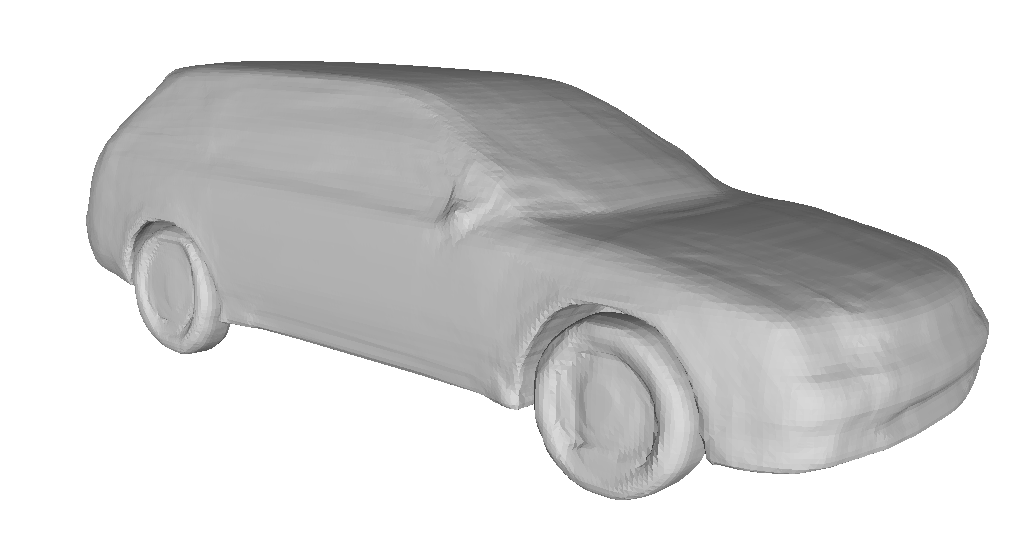}&
   \includegraphics[width=\reconfigwidthsupp]{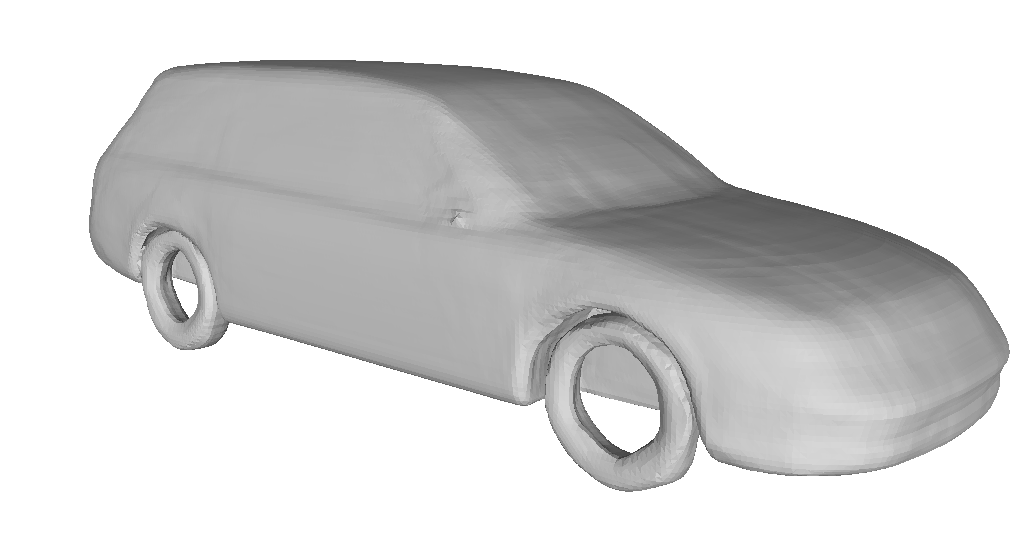} &
   \includegraphics[width=\reconfigwidthsupp]{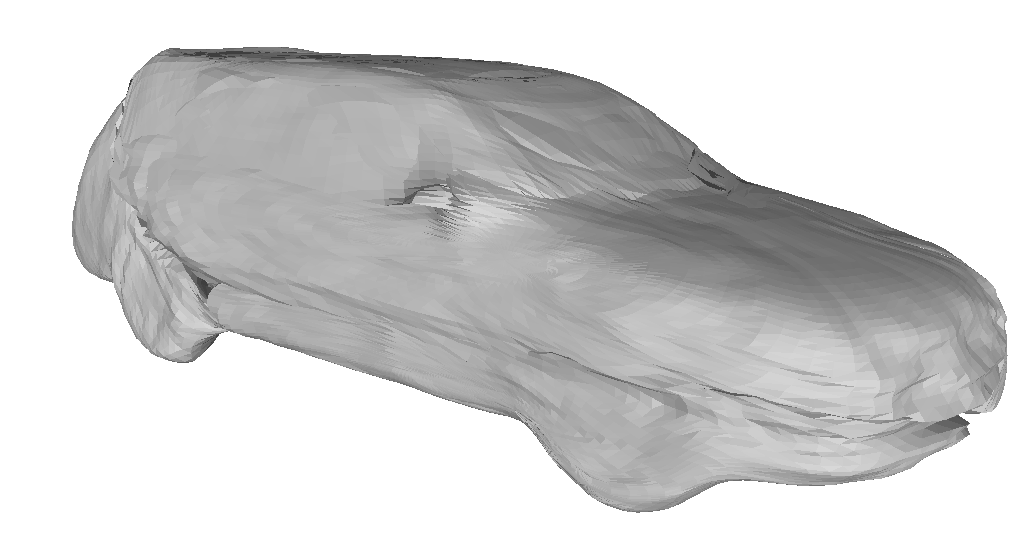}&
   \includegraphics[width=\reconfigwidthsupp]{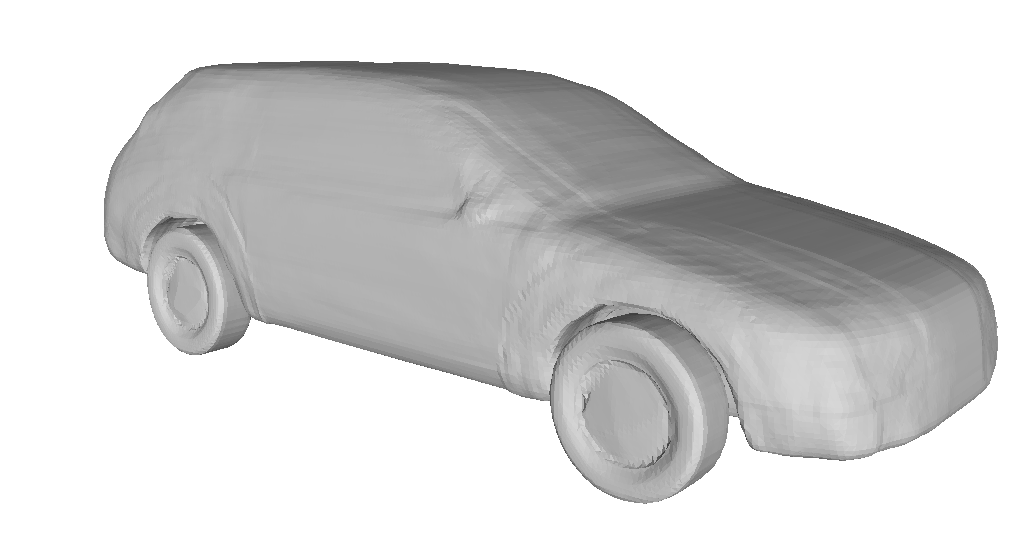}&
   \includegraphics[width=\reconfigwidthsupp]{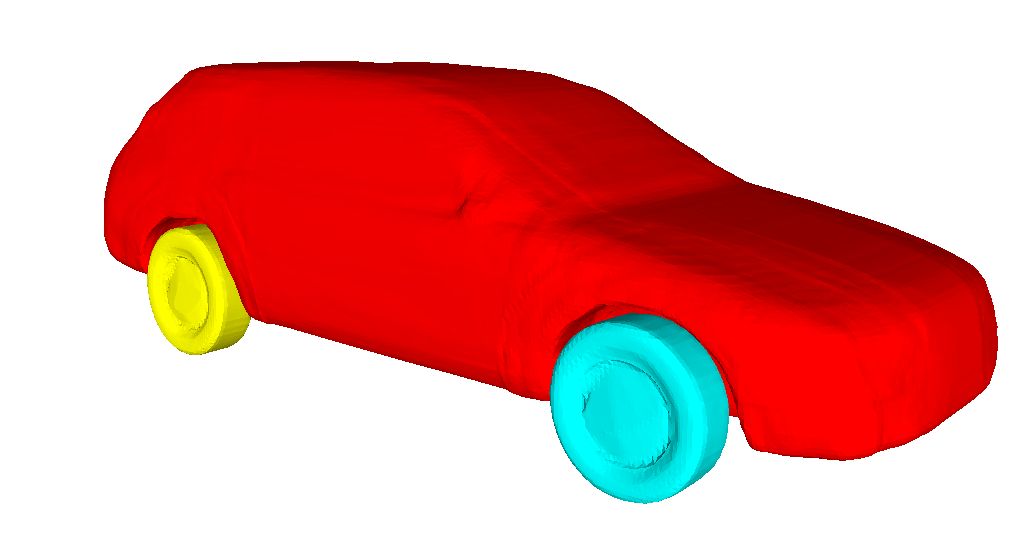}&
   \includegraphics[width=\reconfigwidthsupp]{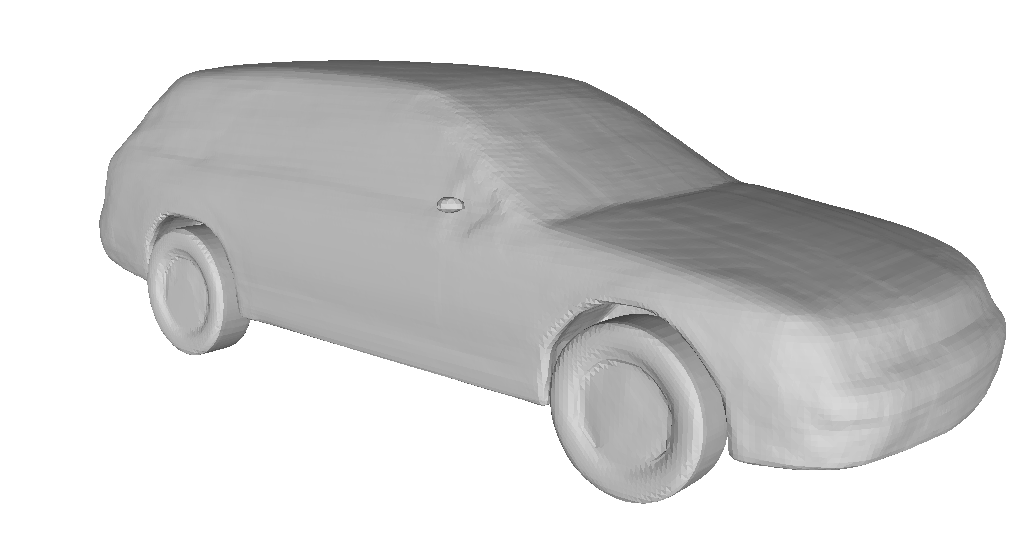}&
   \includegraphics[width=\reconfigwidthsupp]{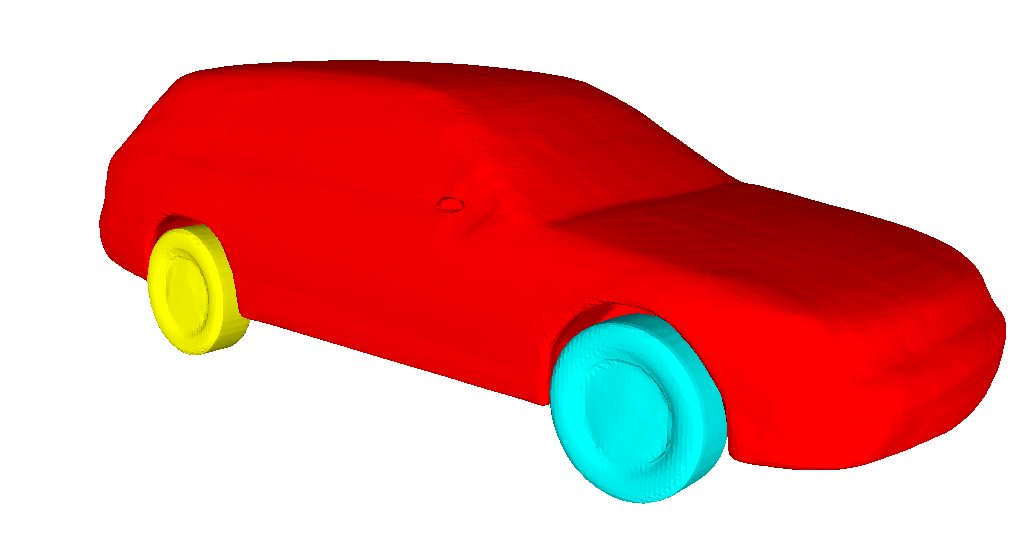}\\
   \includegraphics[width=\reconfigwidthsupp]{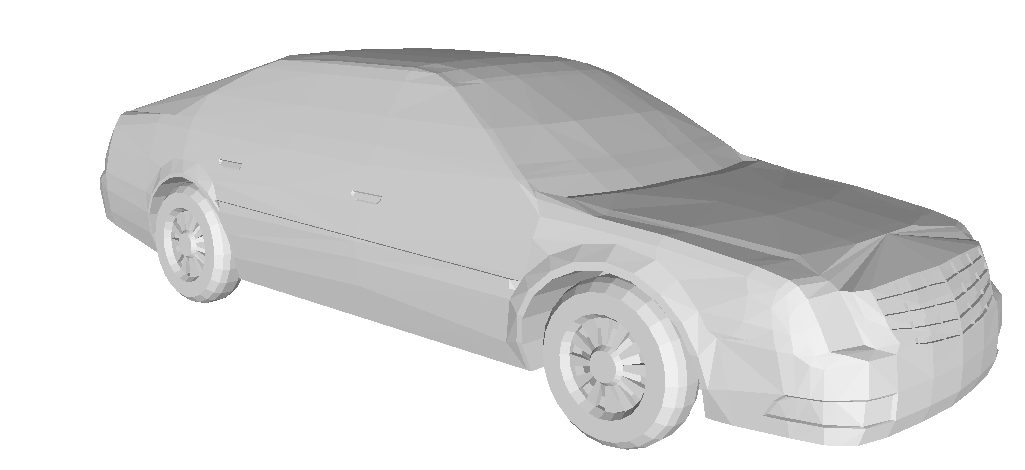} &
   \includegraphics[width=\reconfigwidthsupp]{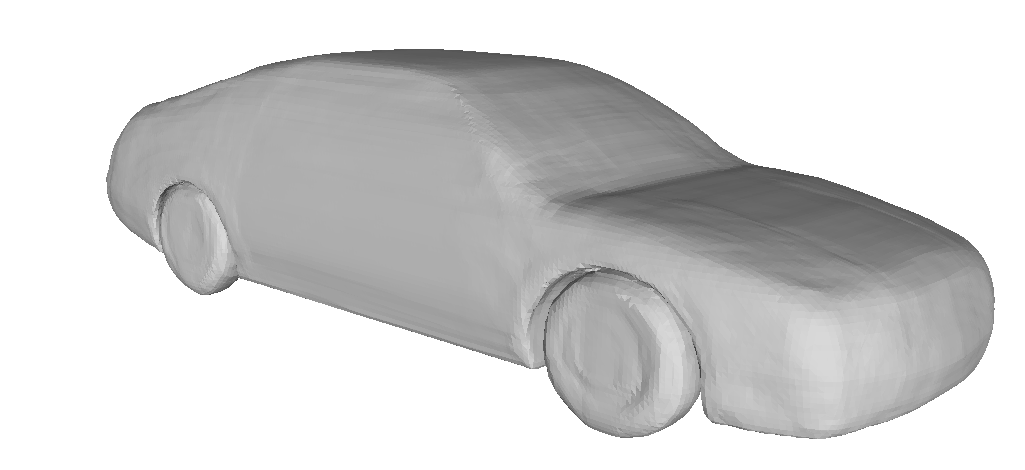}&
   \includegraphics[width=\reconfigwidthsupp]{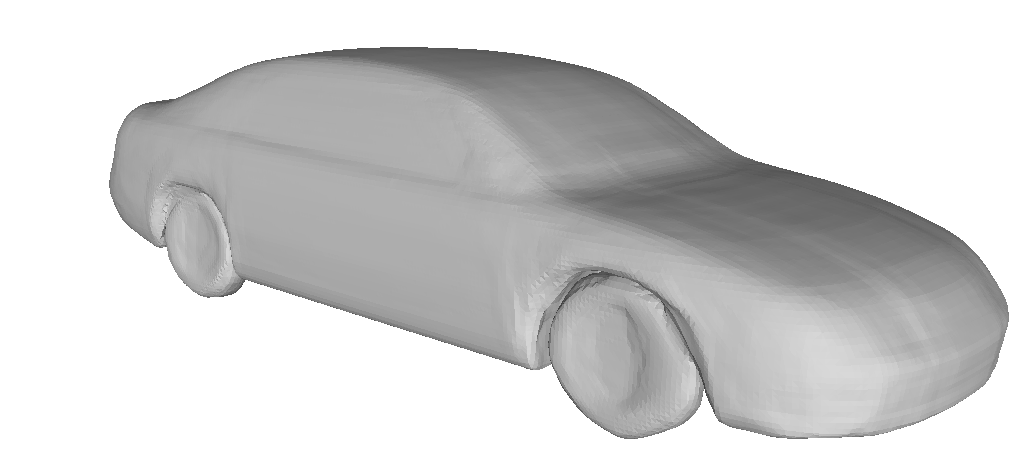} &
   \includegraphics[width=\reconfigwidthsupp]{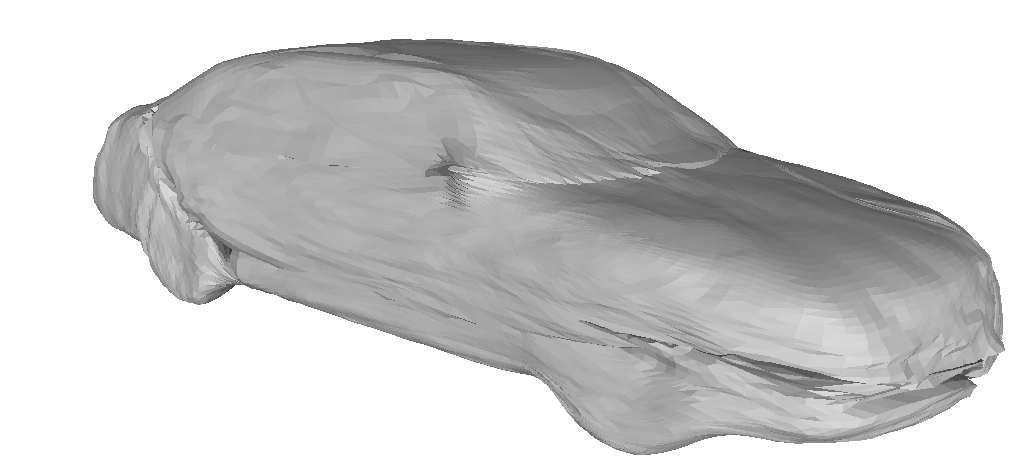}&
   \includegraphics[width=\reconfigwidthsupp]{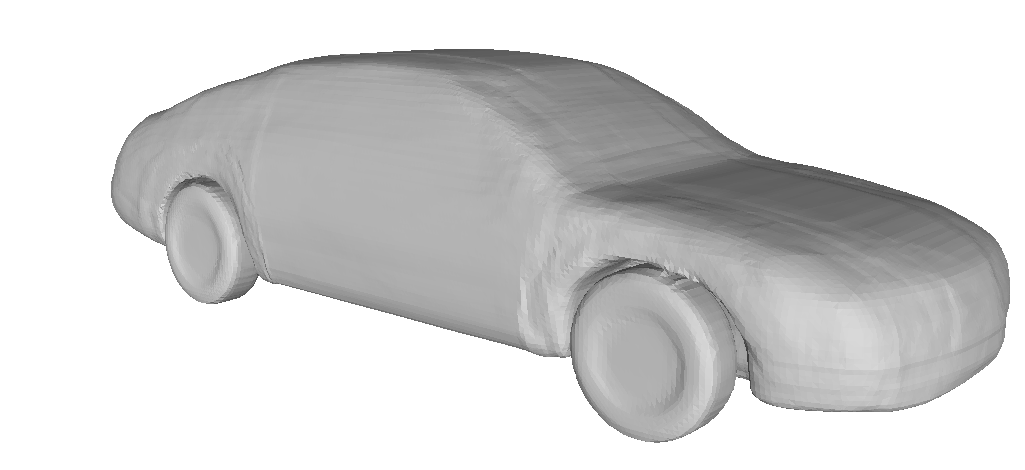}&
   \includegraphics[width=\reconfigwidthsupp]{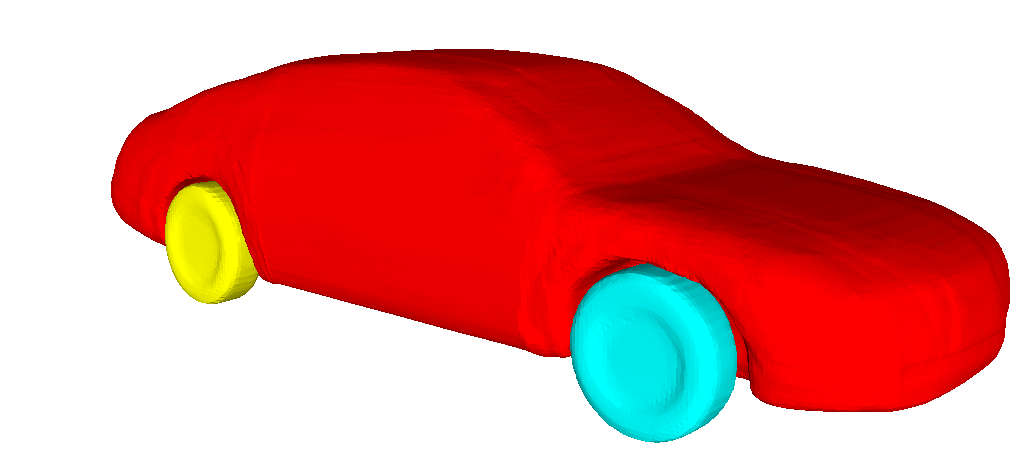}&
\includegraphics[width=\reconfigwidthsupp]{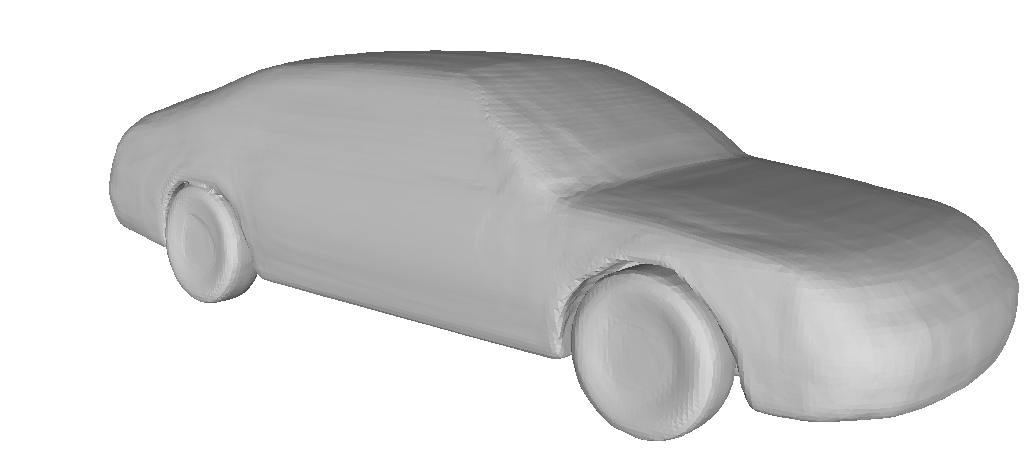}&
\includegraphics[width=\reconfigwidthsupp]{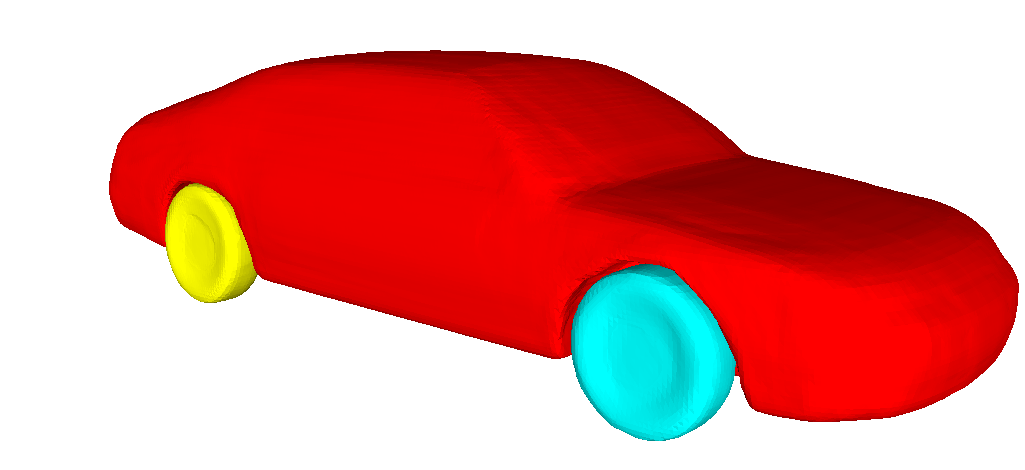}\\
   \includegraphics[width=\reconfigwidthsupp]{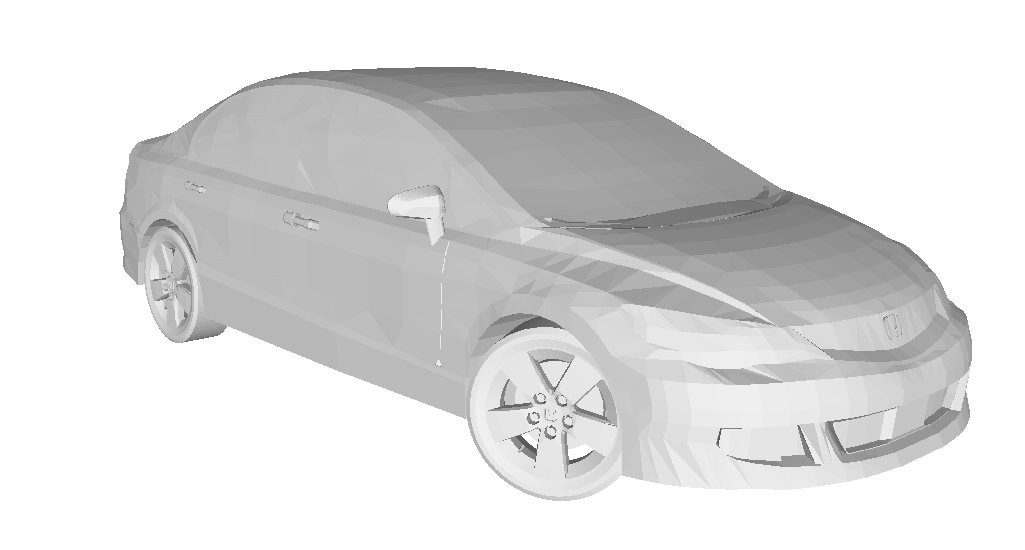} &
   \includegraphics[width=\reconfigwidthsupp]{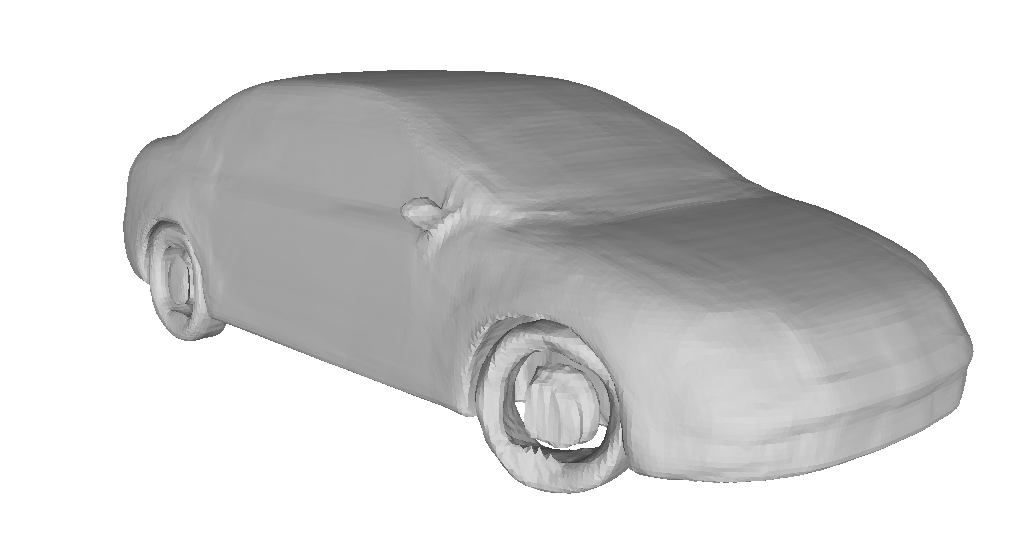}&
   \includegraphics[width=\reconfigwidthsupp]{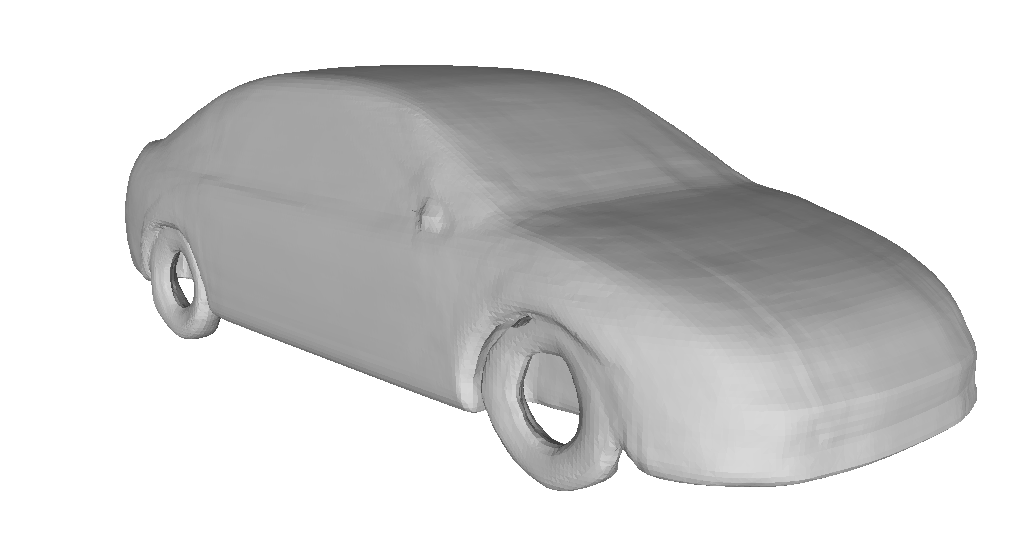} &
   \includegraphics[width=\reconfigwidthsupp]{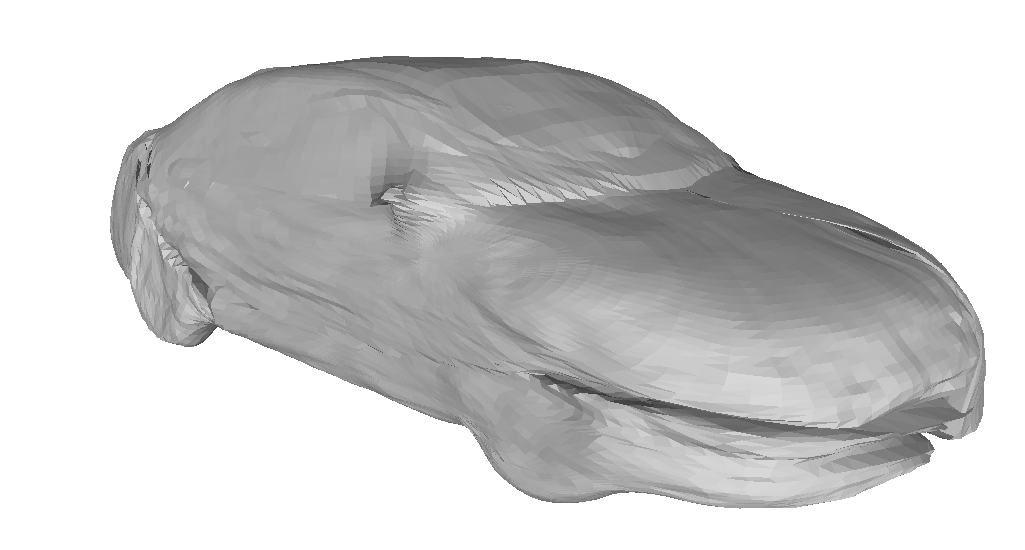}&
\includegraphics[width=\reconfigwidthsupp]{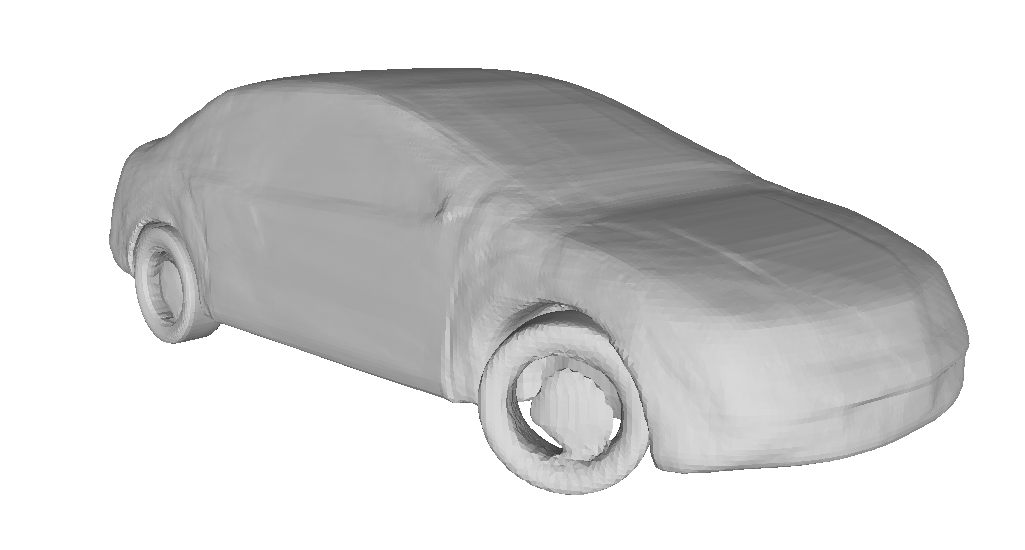}&
\includegraphics[width=\reconfigwidthsupp]{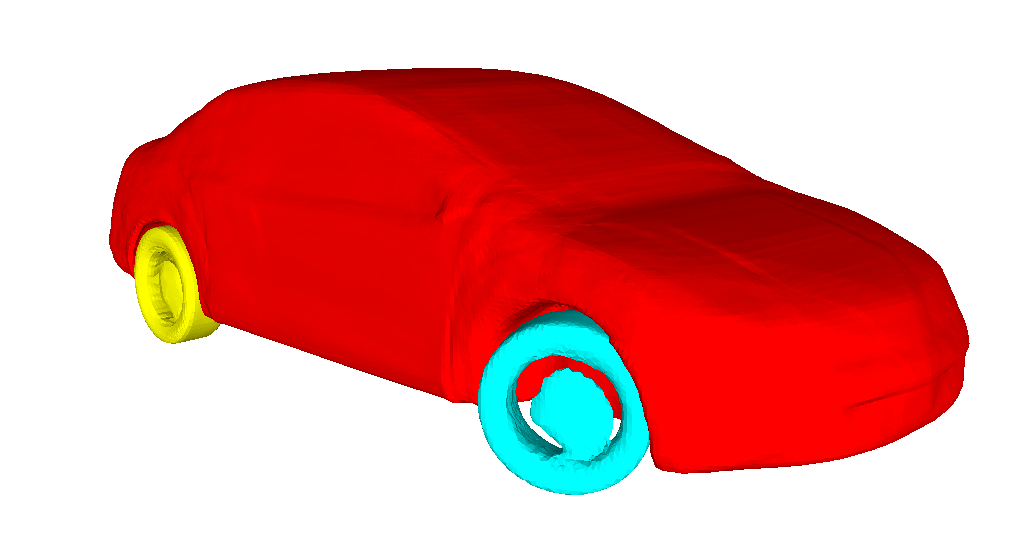}&
\includegraphics[width=\reconfigwidthsupp]{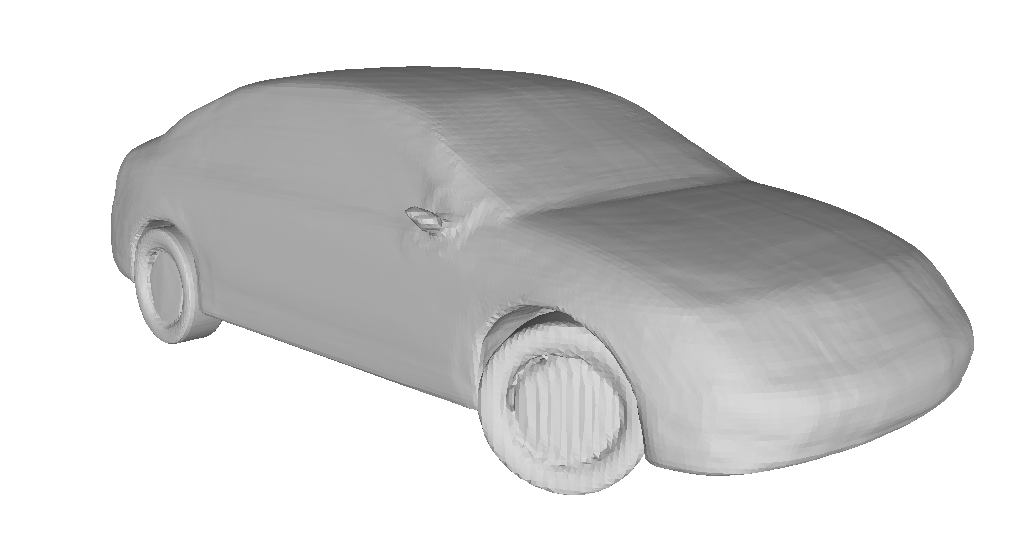}&
\includegraphics[width=\reconfigwidthsupp]{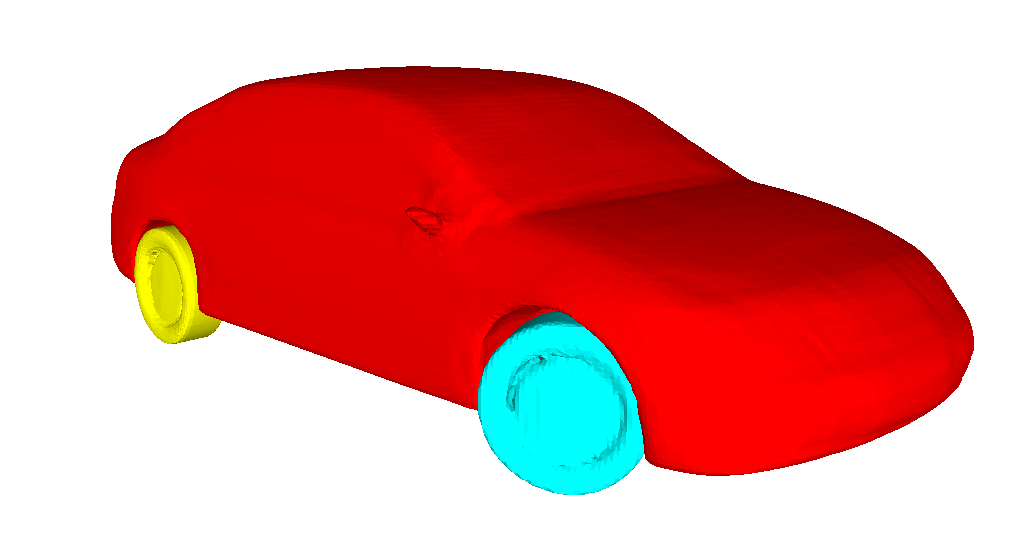}\\
   \includegraphics[width=\reconfigwidthsupp]{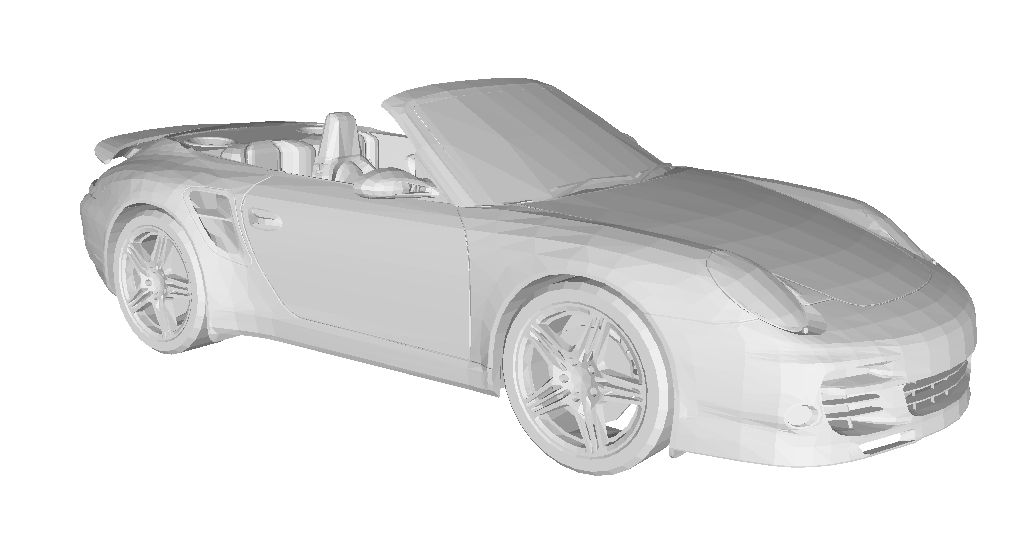} &
   \includegraphics[width=\reconfigwidthsupp]{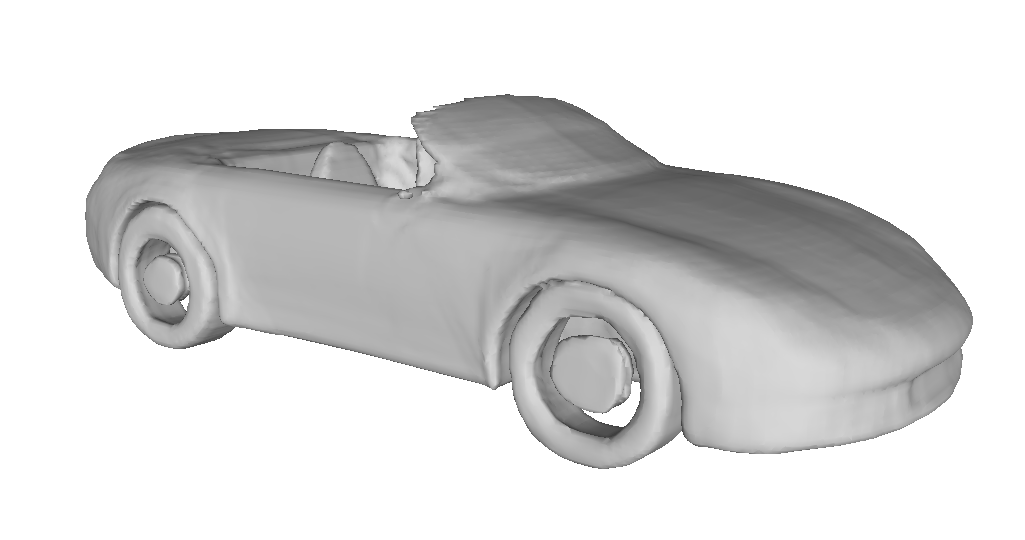}&
   \includegraphics[width=\reconfigwidthsupp]{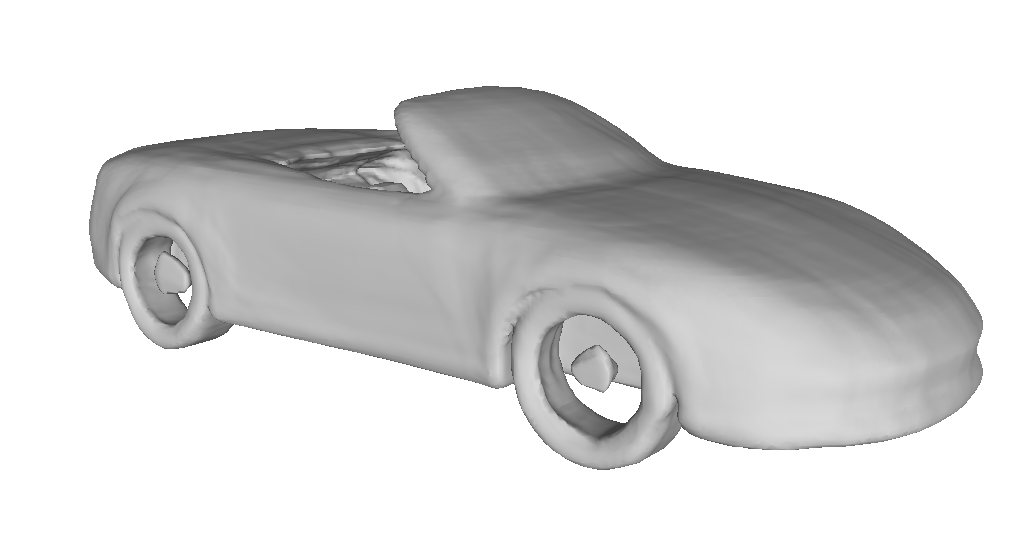} &
   \includegraphics[width=\reconfigwidthsupp]{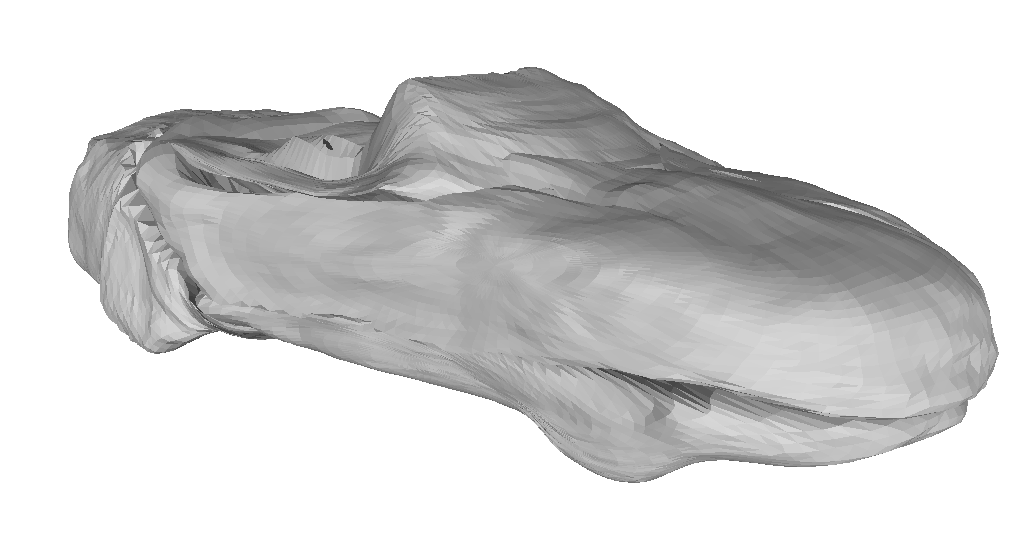}&
   \includegraphics[width=\reconfigwidthsupp]{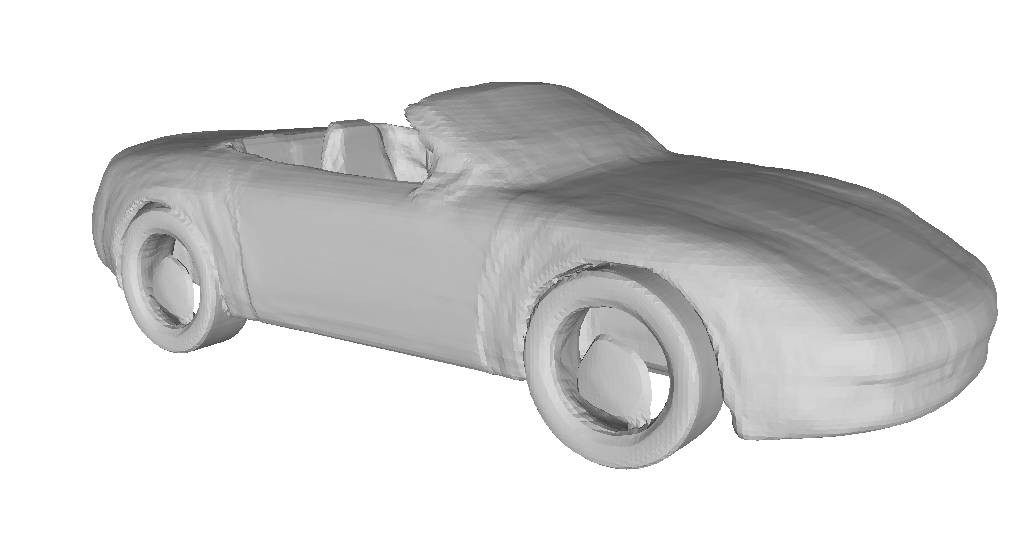}&
   \includegraphics[width=\reconfigwidthsupp]{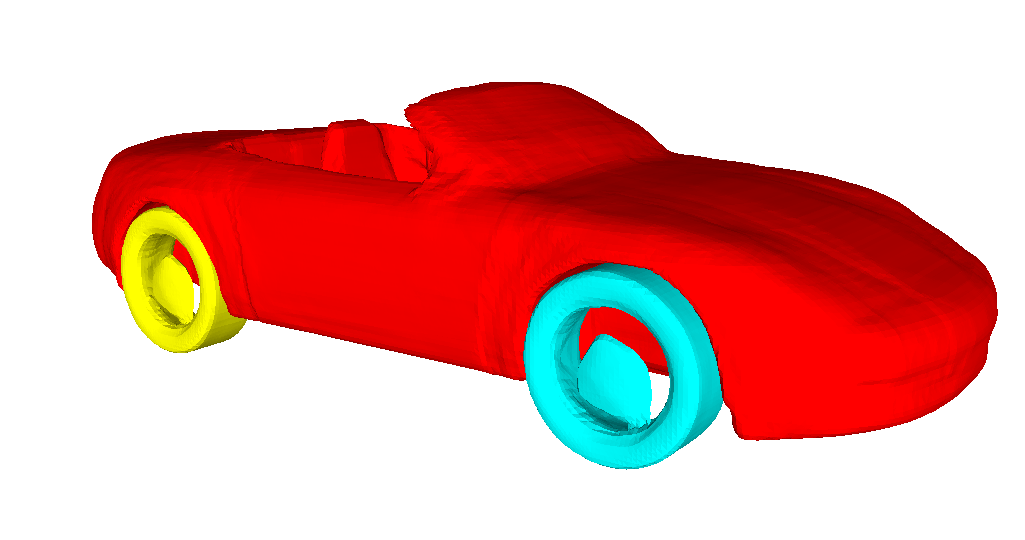}&
\includegraphics[width=\reconfigwidthsupp]{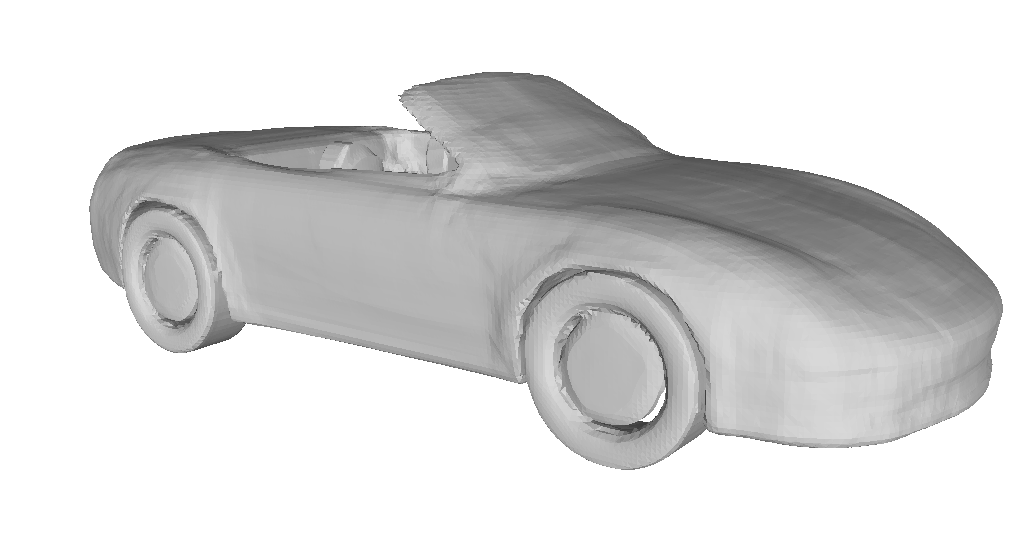}&
\includegraphics[width=\reconfigwidthsupp]{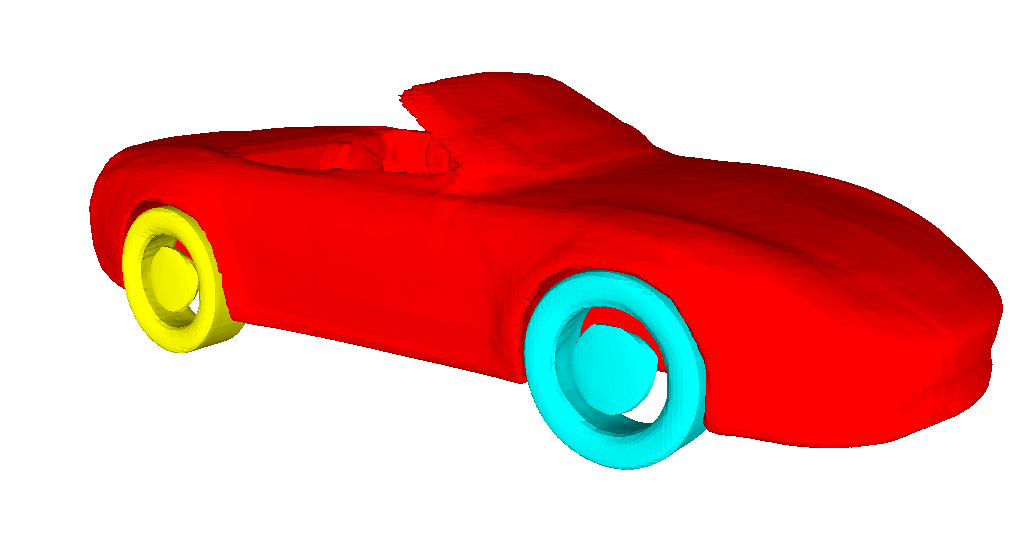}\\
   \includegraphics[width=\reconfigwidthsupp]{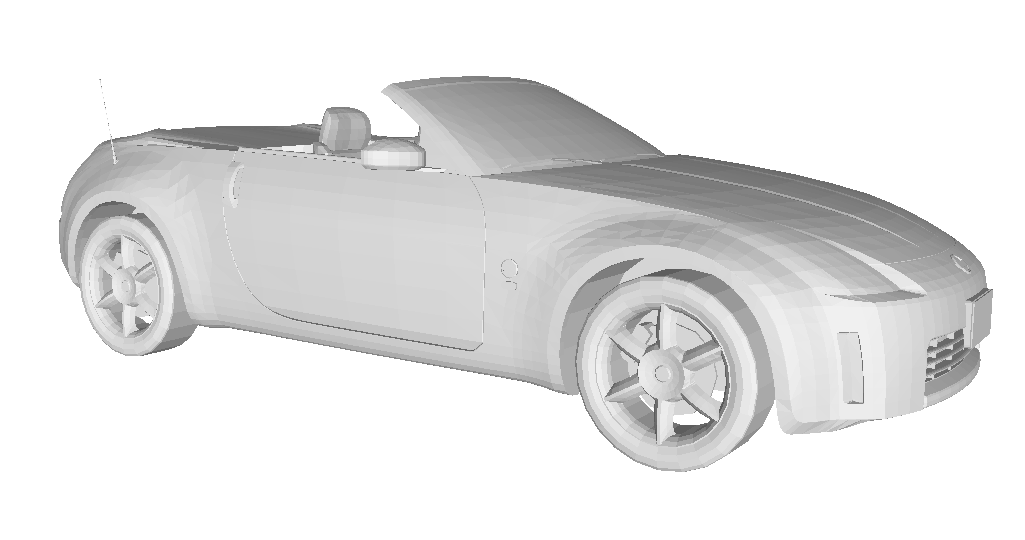} &
   \includegraphics[width=\reconfigwidthsupp]{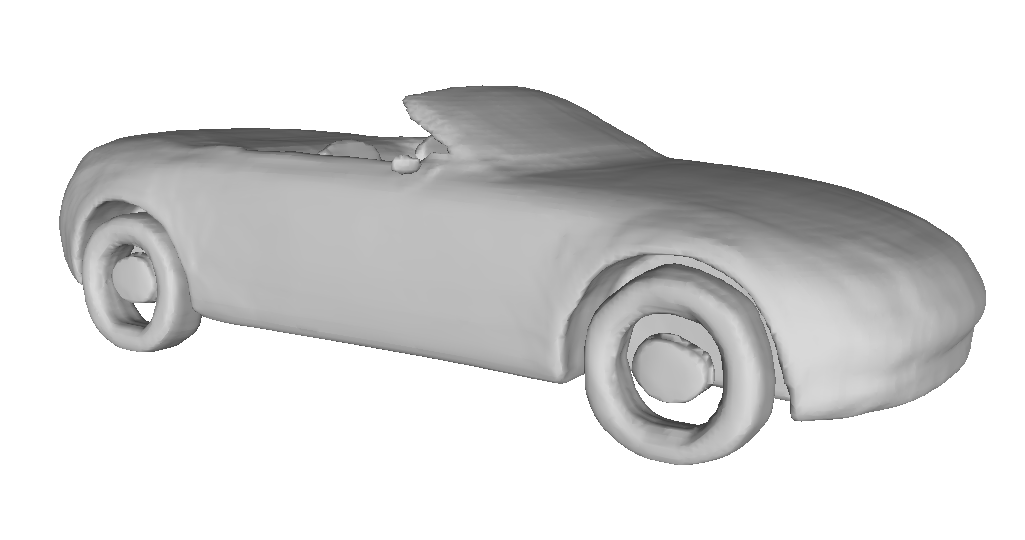}&
   \includegraphics[width=\reconfigwidthsupp]{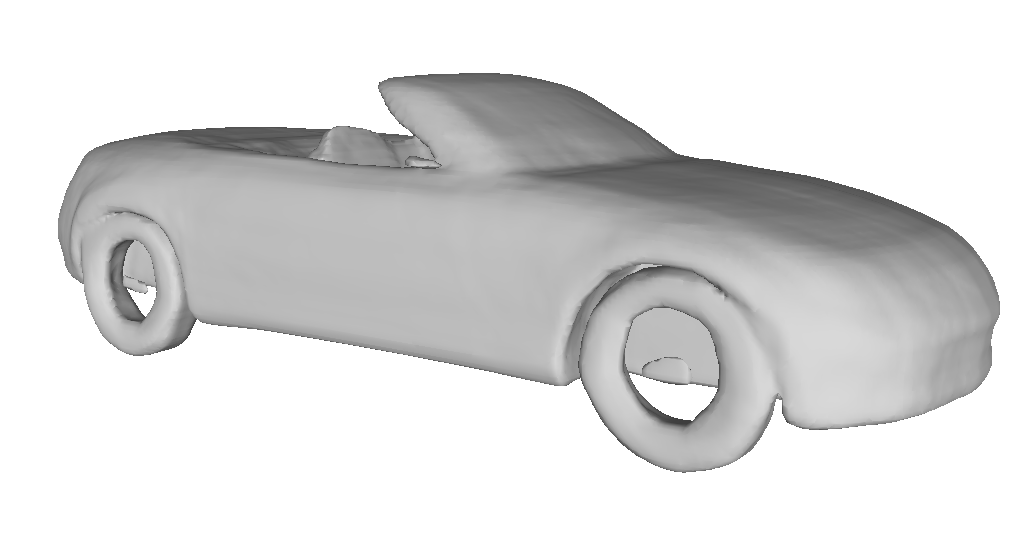} &
   \includegraphics[width=\reconfigwidthsupp]{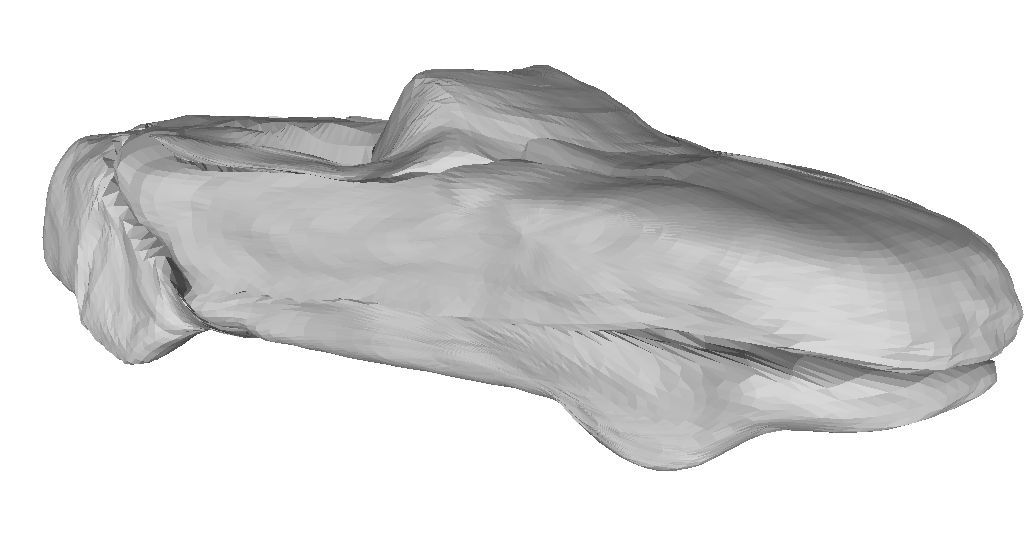}&
\includegraphics[width=\reconfigwidthsupp]{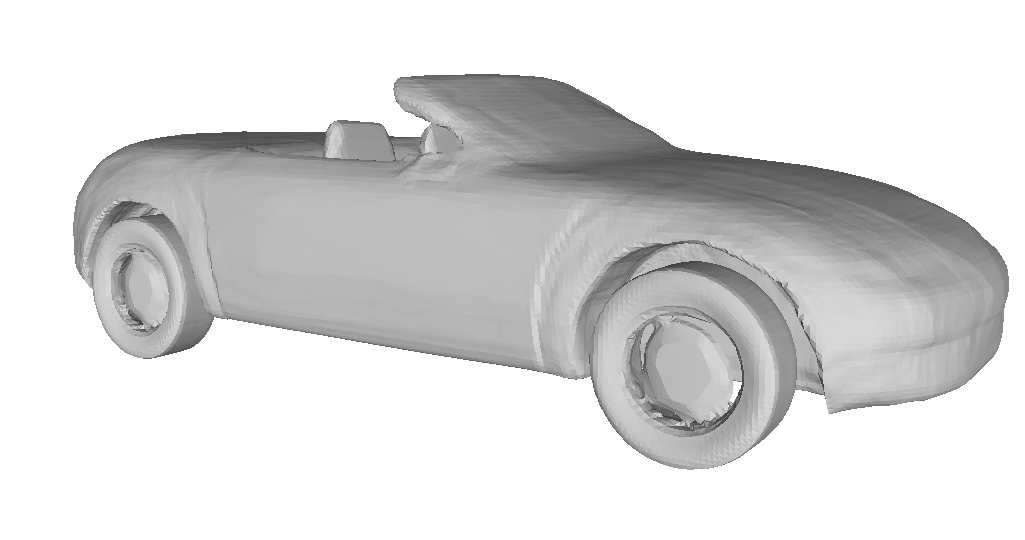}&
\includegraphics[width=\reconfigwidthsupp]{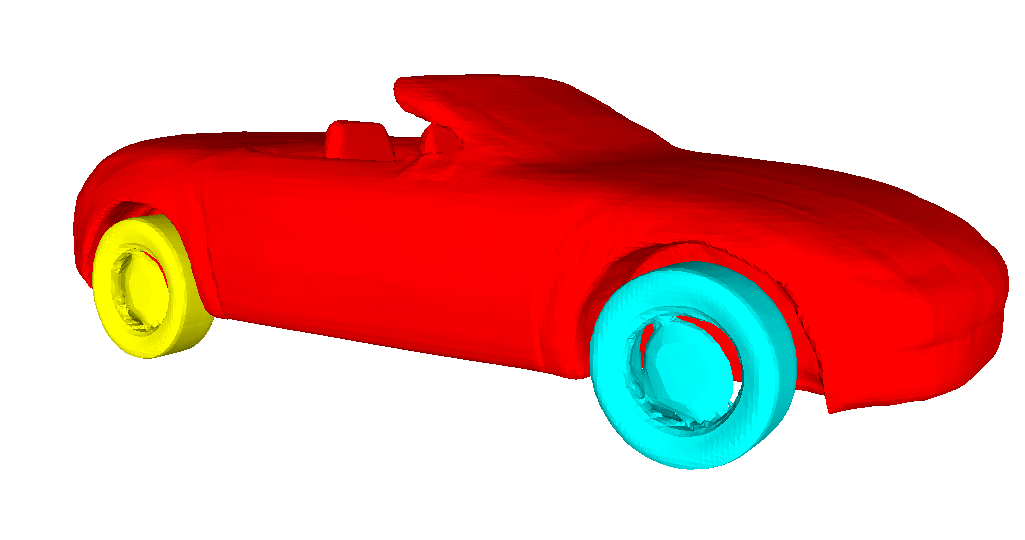}&
\includegraphics[width=\reconfigwidthsupp]{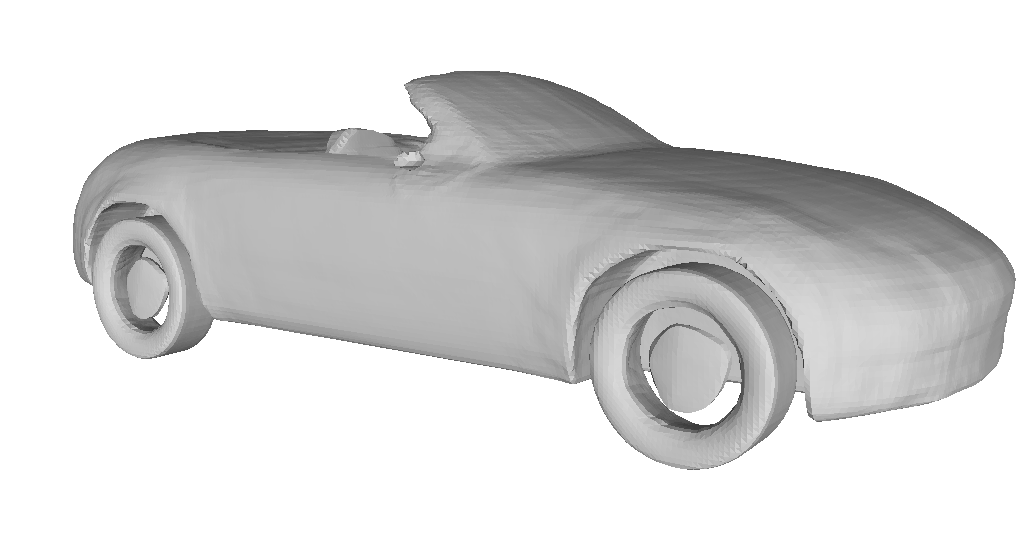}&
\includegraphics[width=\reconfigwidthsupp]{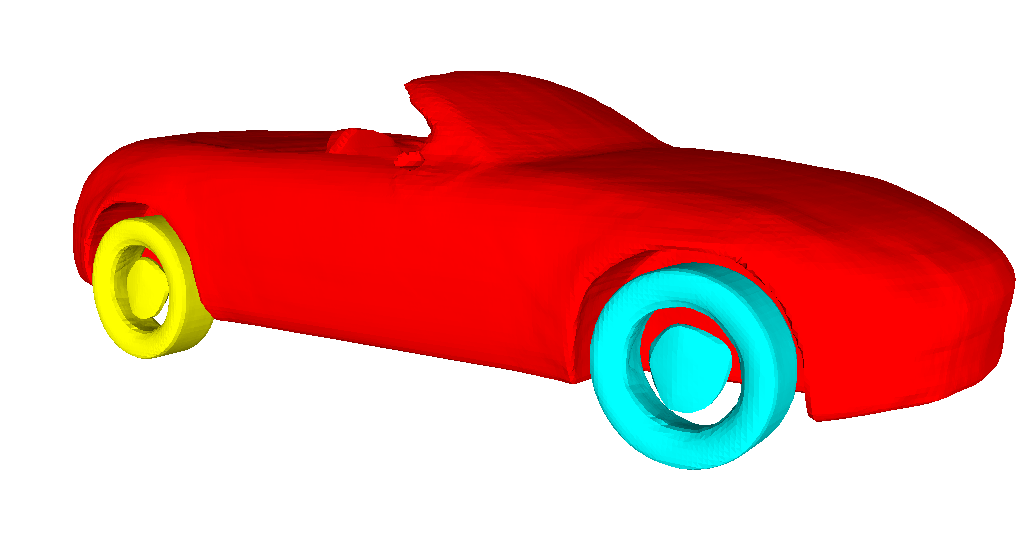}\\
    GT & \DeepS{} & \DualS{} & 
\NP{} & \multicolumn{2}{c}{\HS{}} & \multicolumn{2}{c}{\HSs{}} \\
  \end{tabular}
  \caption{{\it Additional test shape reconstruction on Cars.} {Our methods \HS{} and \HSs{} generate shapes of comparable quality while having advantages over competing methods such as the physical plausibility of the wheels. For our models, we also provide a version where parts are color-coded.}}
  \label{fig:car_recon_supple}
\end{figure*}
\setlength{\tabcolsep}{\mytabcolsepsupp}

\newlength{\reconfigwidthmixsupp}
\setlength{\reconfigwidthmixsupp}{0.09\textwidth}

\setlength\mytabcolsepsupp{\tabcolsep}
\setlength\tabcolsep{2pt}

\begin{figure*}[t]
  \centering
  \small  
\begin{tabular}{cccccccc}
   \includegraphics[width=\reconfigwidthmixsupp]{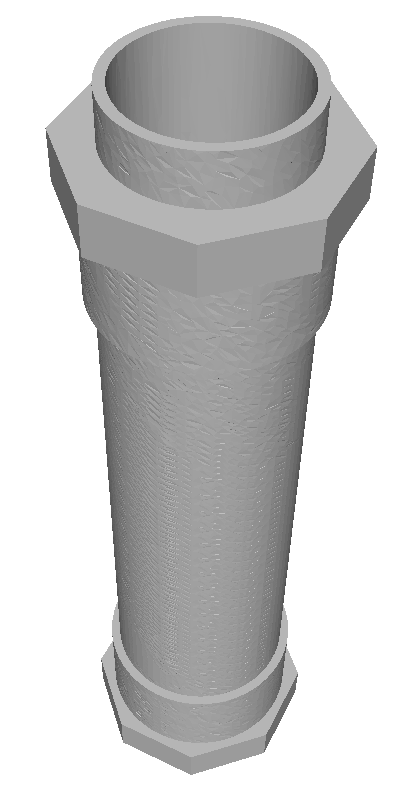} &
   \includegraphics[width=\reconfigwidthmixsupp]{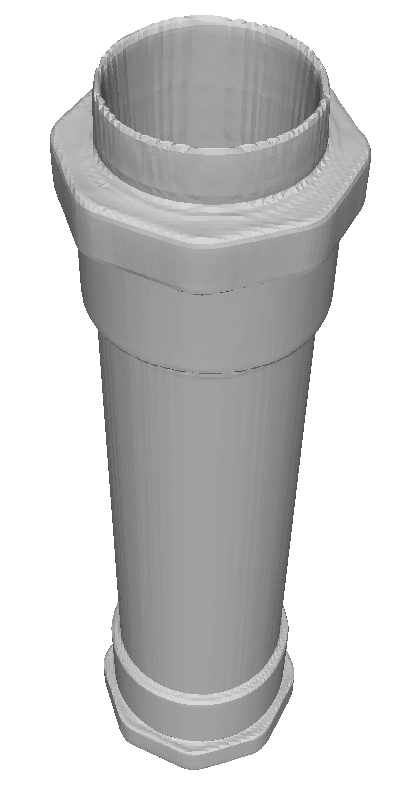}&
   \includegraphics[width=\reconfigwidthmixsupp]{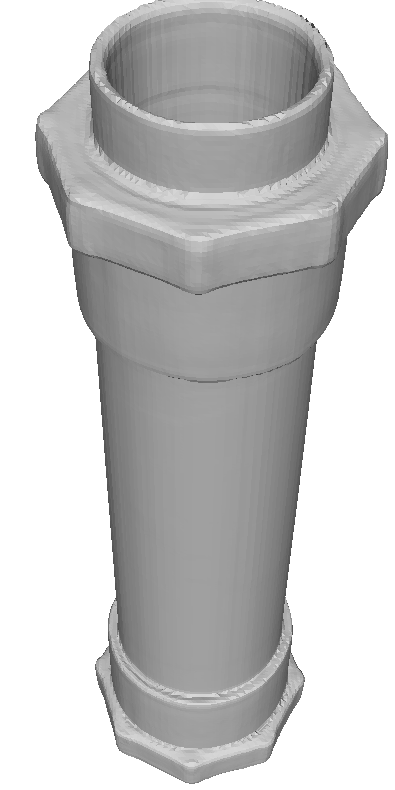} &
   \includegraphics[width=\reconfigwidthmixsupp]{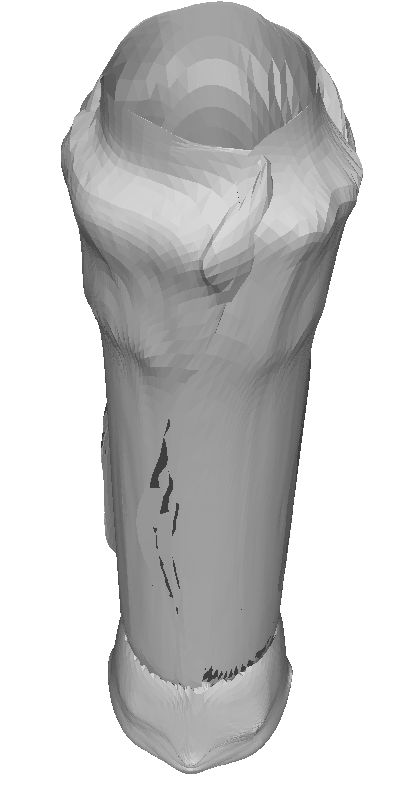}&
   \includegraphics[width=\reconfigwidthmixsupp]{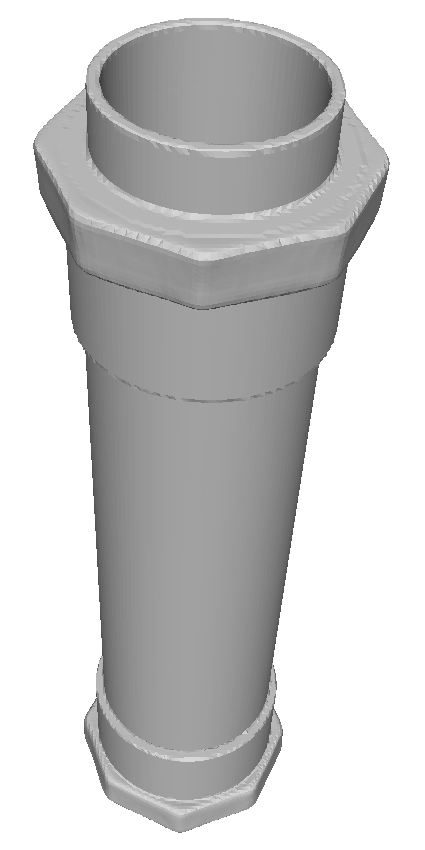}&
   \includegraphics[width=\reconfigwidthmixsupp]{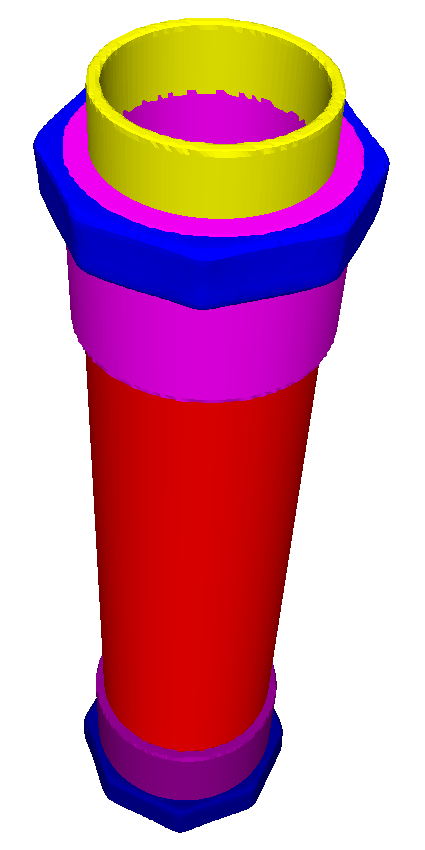}&
\includegraphics[width=\reconfigwidthmixsupp]{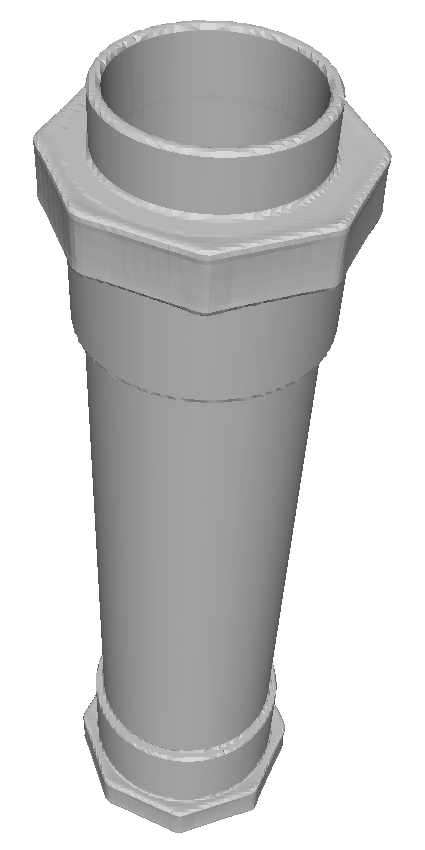}&
\includegraphics[width=\reconfigwidthmixsupp]{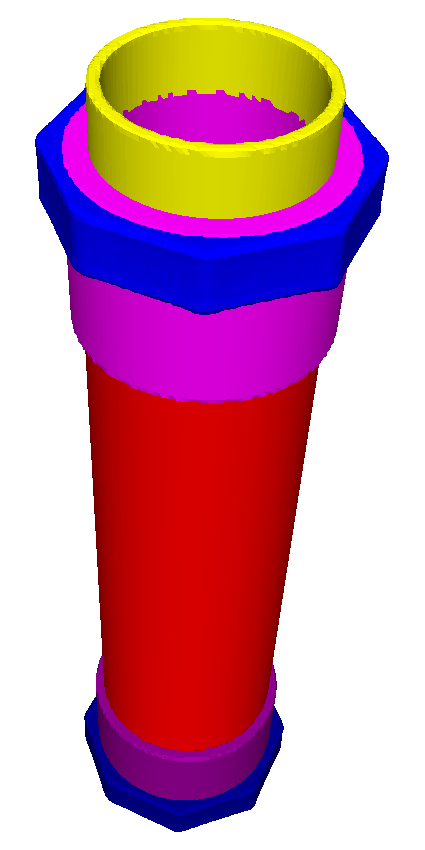}\\
\includegraphics[width=\reconfigwidthmixsupp]{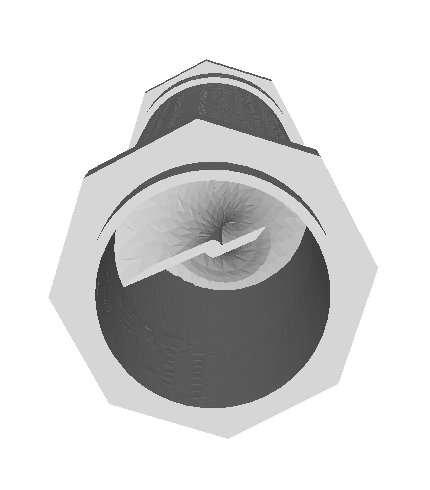} &
\includegraphics[width=\reconfigwidthmixsupp]{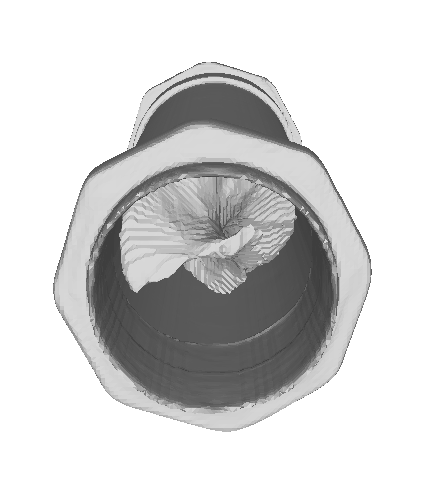}&
\includegraphics[width=\reconfigwidthmixsupp]{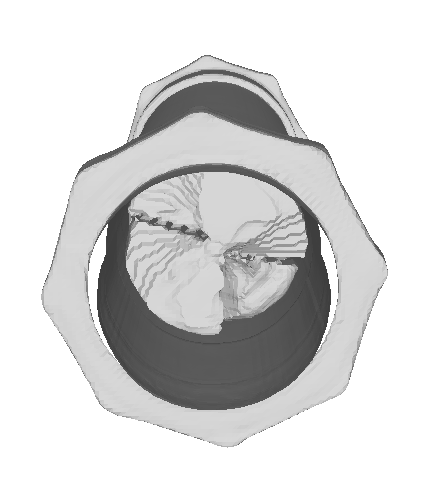} &
\includegraphics[width=\reconfigwidthmixsupp]{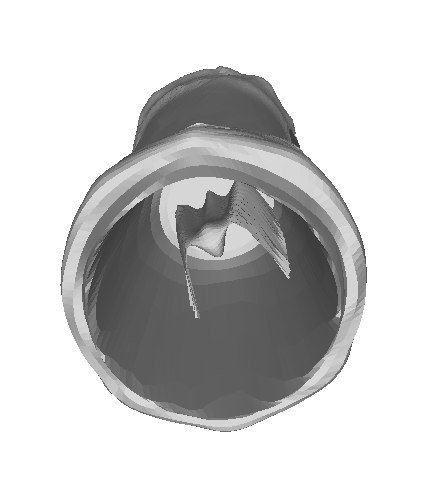}&
\includegraphics[width=\reconfigwidthmixsupp]{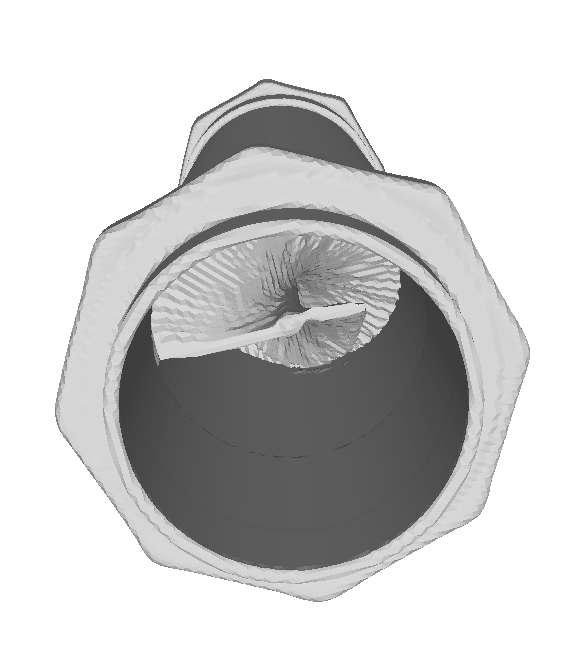}&
\includegraphics[width=\reconfigwidthmixsupp]{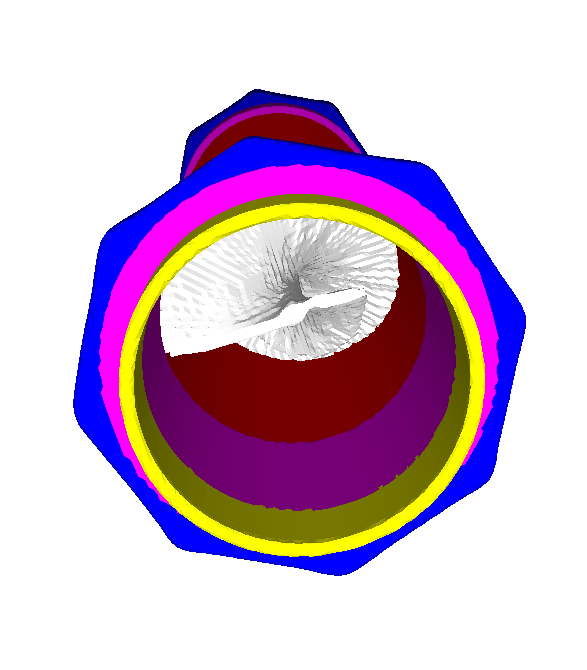}&
\includegraphics[width=\reconfigwidthmixsupp]{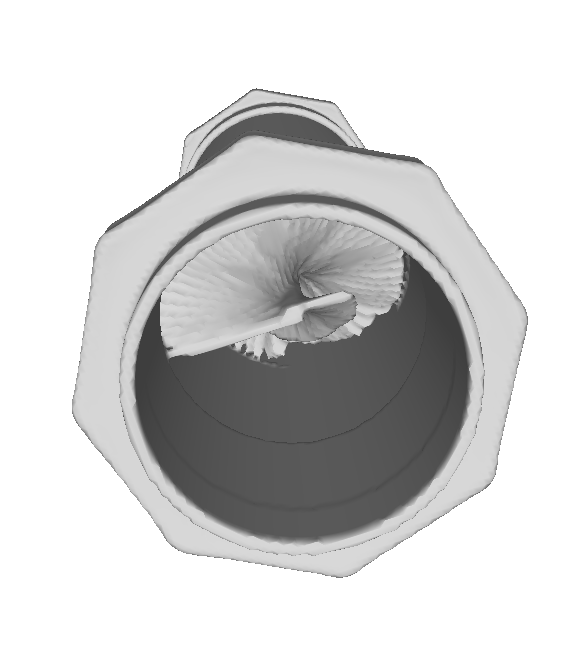}&
\includegraphics[width=\reconfigwidthmixsupp]{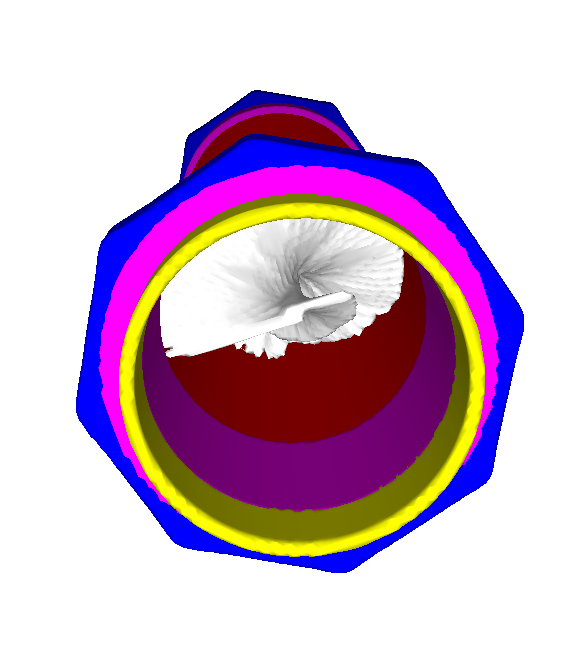}\\
   \includegraphics[width=\reconfigwidthmixsupp]{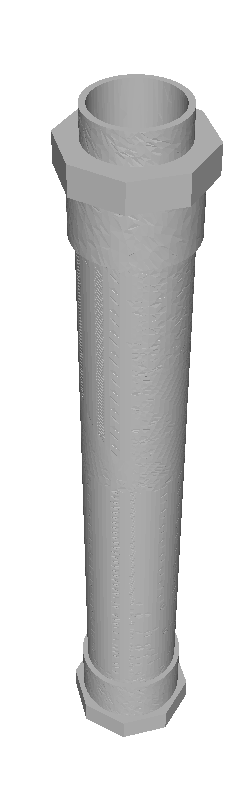} &
   \includegraphics[width=\reconfigwidthmixsupp]{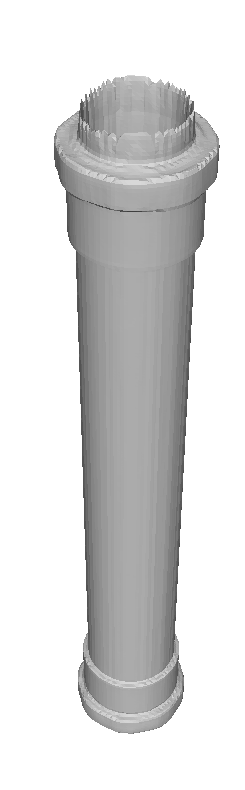}&
   \includegraphics[width=\reconfigwidthmixsupp]{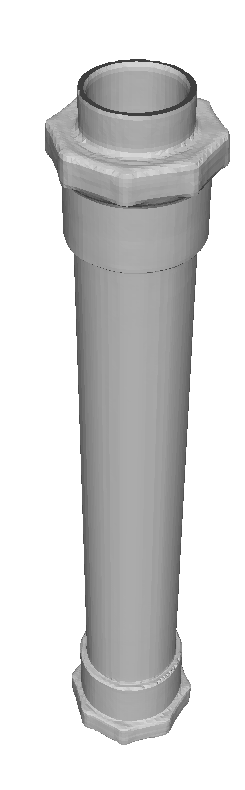} &
   \includegraphics[width=\reconfigwidthmixsupp]{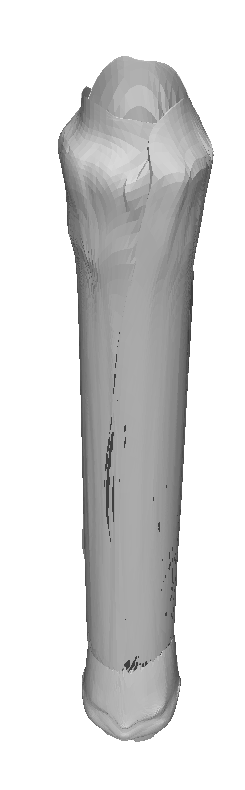}&
\includegraphics[width=\reconfigwidthmixsupp]{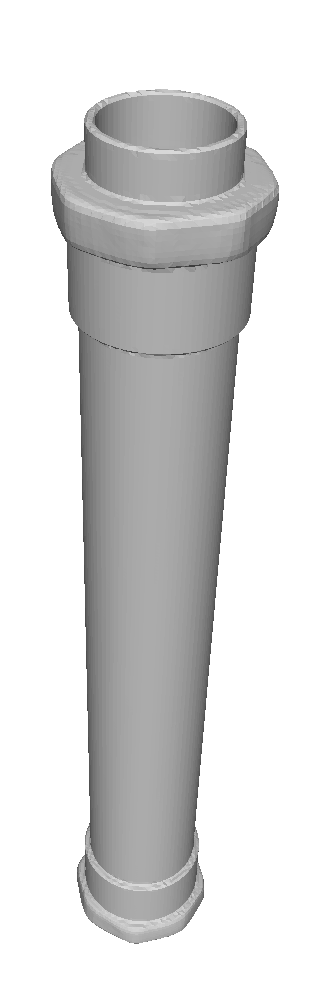}&
\includegraphics[width=\reconfigwidthmixsupp]{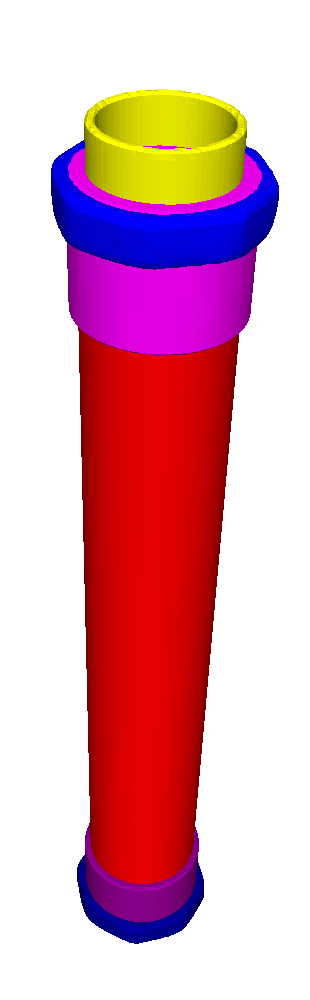}&
\includegraphics[width=\reconfigwidthmixsupp]{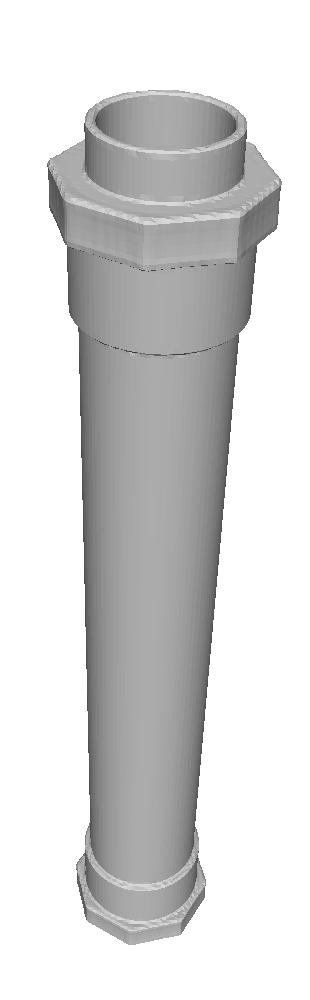}&
\includegraphics[width=\reconfigwidthmixsupp]{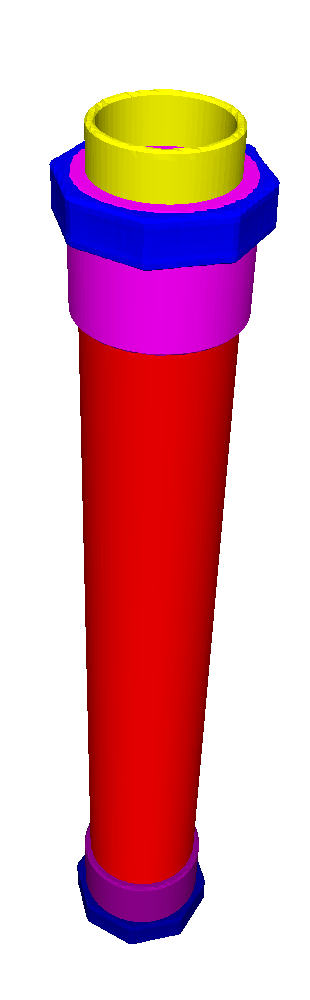}\\
   \includegraphics[width=\reconfigwidthmixsupp]{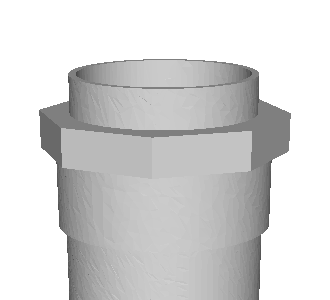} &
   \includegraphics[width=\reconfigwidthmixsupp]{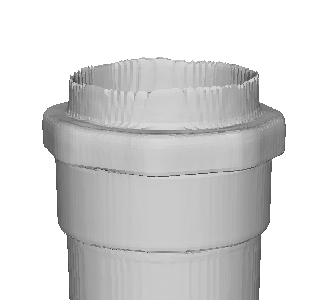}&
   \includegraphics[width=\reconfigwidthmixsupp]{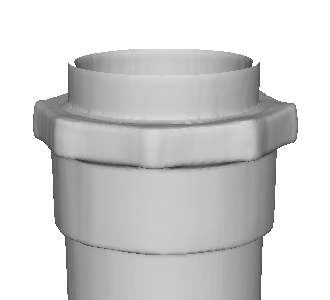} &
   \includegraphics[width=\reconfigwidthmixsupp]{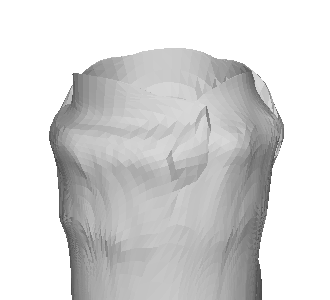}&
\includegraphics[width=\reconfigwidthmixsupp]{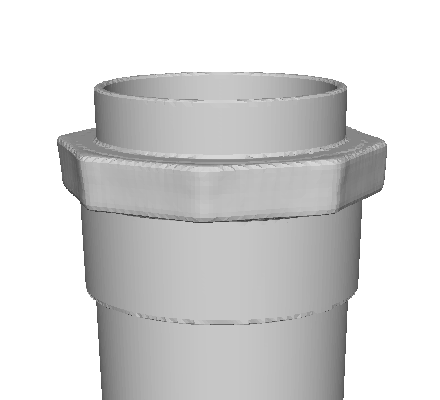}&
\includegraphics[width=\reconfigwidthmixsupp]{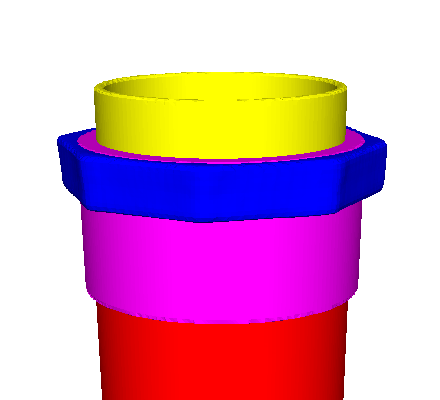}&
\includegraphics[width=\reconfigwidthmixsupp]{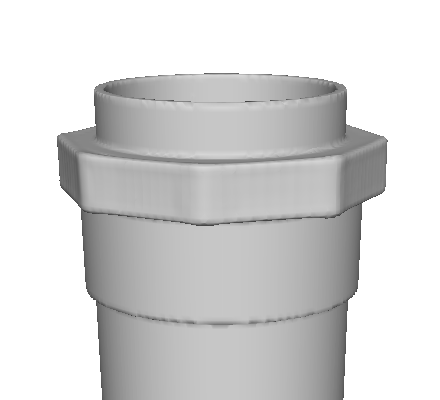}&
\includegraphics[width=\reconfigwidthmixsupp]{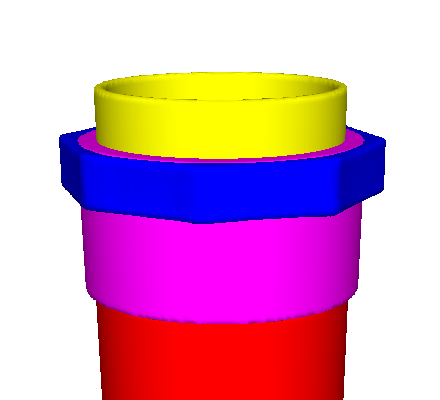}\\
    GT & \DeepS{} & \DualS{} & 
    \NP{}& \multicolumn{2}{c}{\HS{}} & \multicolumn{2}{c}{\HSs{}} \\
  \end{tabular}
  \caption{{\it Additional test shape reconstruction on Mixers.} {Our methods \HS{} and \HSs{} generate high quality surfaces with the help of geometric primitives, while the competing approaches suffer from local irregularities. For our models, we also provide a version where parts are color-coded.}}
  \label{fig:mixer_recon_supple}
\end{figure*}
\setlength{\tabcolsep}{\mytabcolsepsupp}

\subsection{Parametric shape manipulation}
\label{appendix:manip}

As discussed in Section~4.2 of the main paper, \HS{} enables to manipulate shapes in a parametric manner by directly editing the geometric parameters. For the \textit{shared} latent space approach, such manipulation consists in optimizing the latent vector with respect to the \textit{Latent Decoder} outputs, as described above.
We provide manipulation examples with both approaches in Fig.~\ref{fig:manip_all_supp}.


\newlength{\manipheightsupp}
\setlength{\manipheightsupp}{5.8cm}
\newcommand{\intercellmanipsupp}{\hspace{-0.5mm}}

\begin{figure*}[t]
\centering
\small  
\begin{tabular}{cc|cccc}\intercellmanipsupp
	\includegraphics[height=\manipheightsupp]{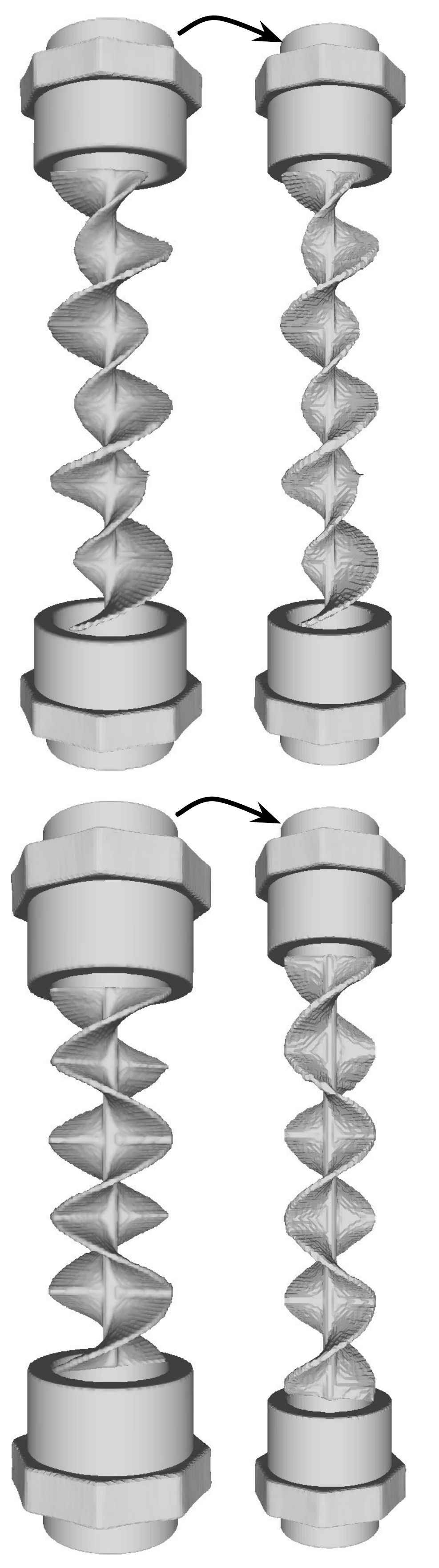} \intercellmanipsupp&\intercellmanipsupp
	\includegraphics[height=\manipheightsupp]{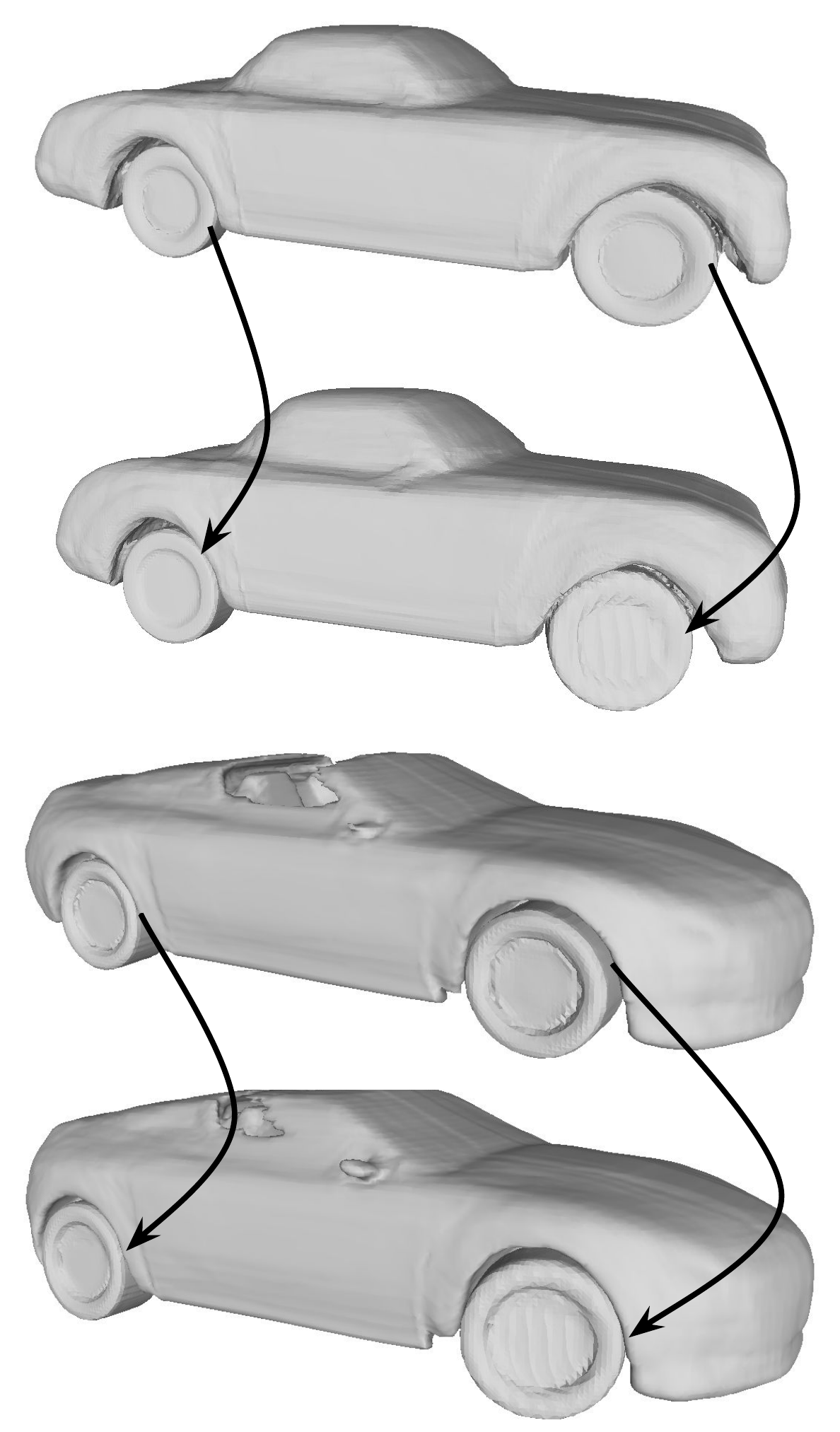} \intercellmanipsupp&\intercellmanipsupp
	\includegraphics[height=\manipheightsupp]{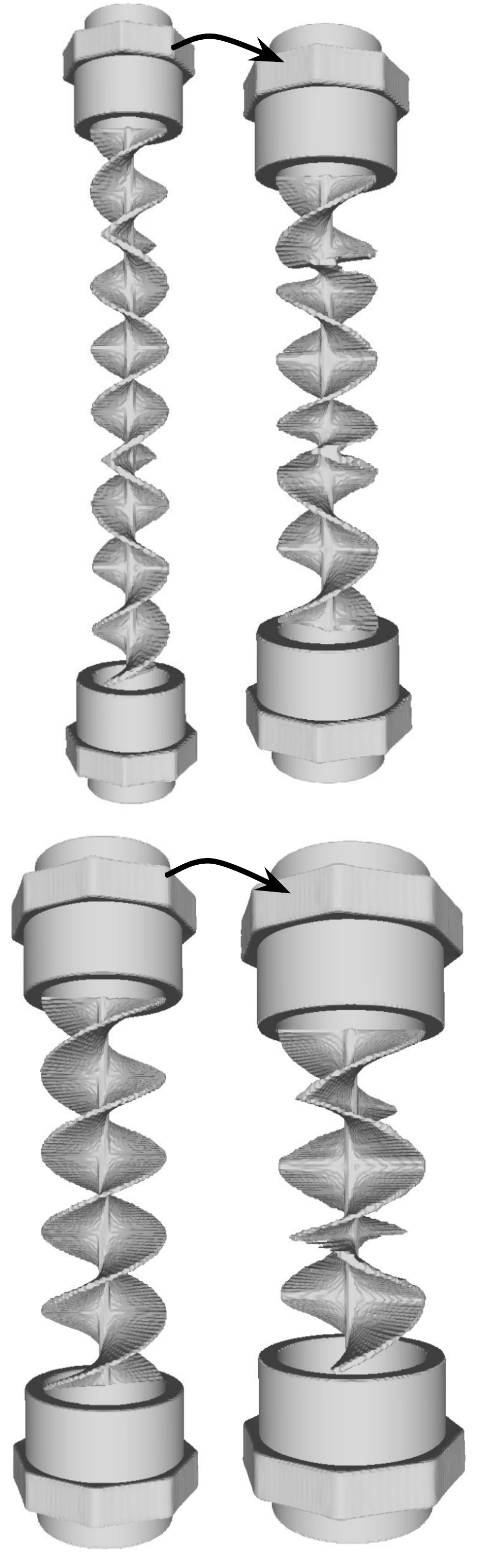} &
	\includegraphics[height=\manipheightsupp]{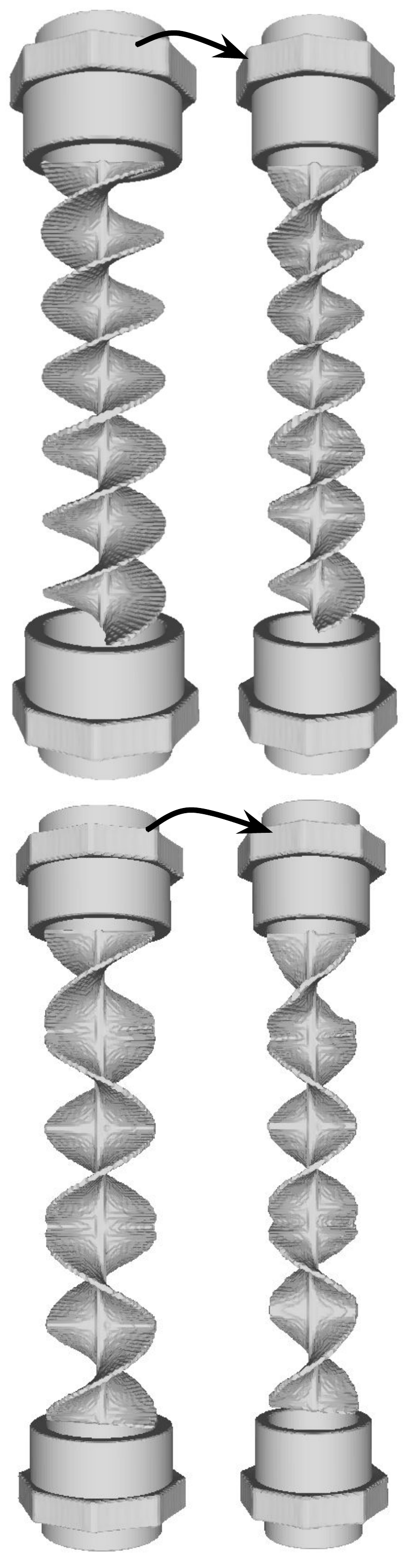} \intercellmanipsupp& \intercellmanipsupp
	\includegraphics[height=\manipheightsupp]{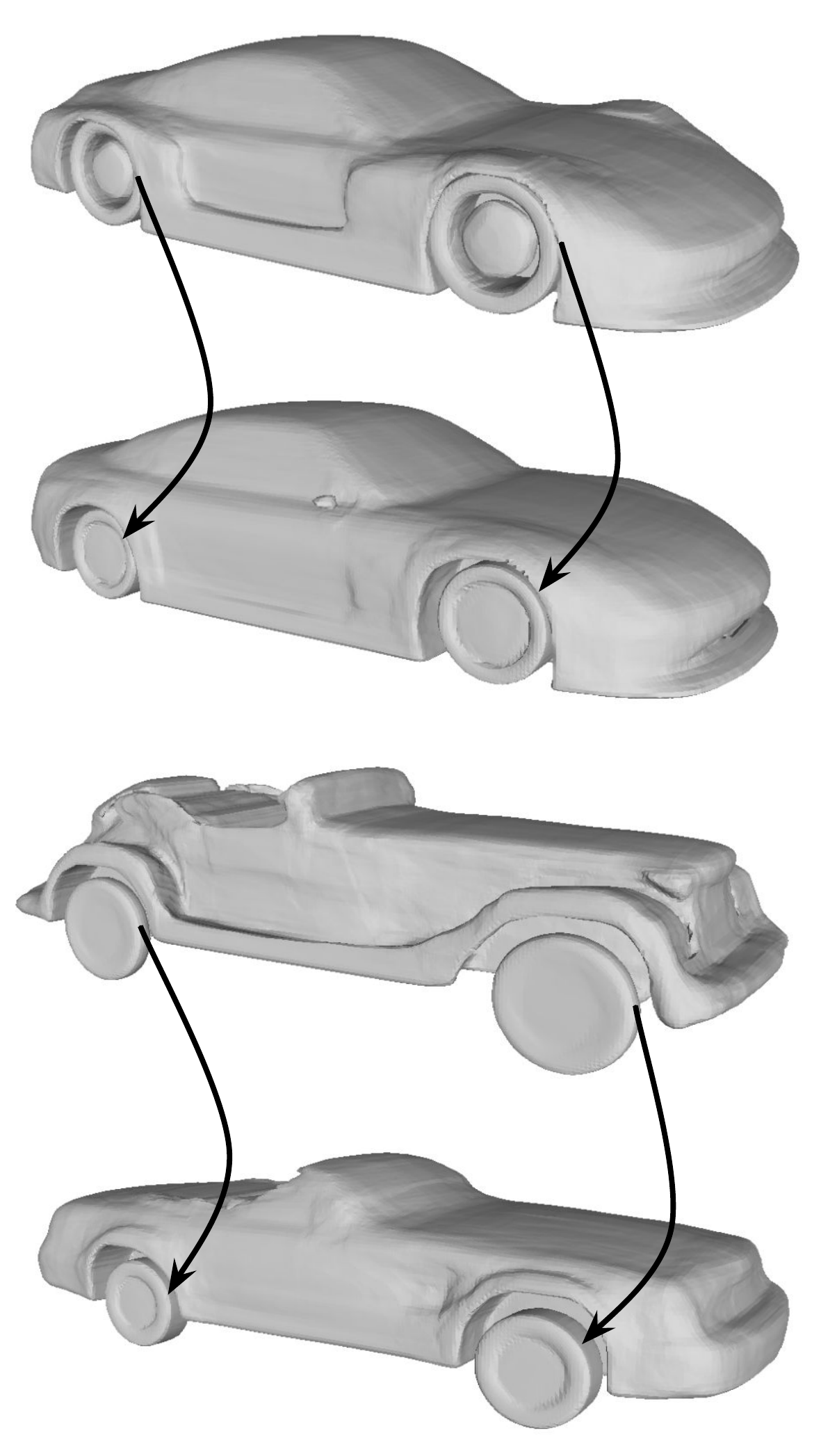} &
	\includegraphics[height=\manipheightsupp]{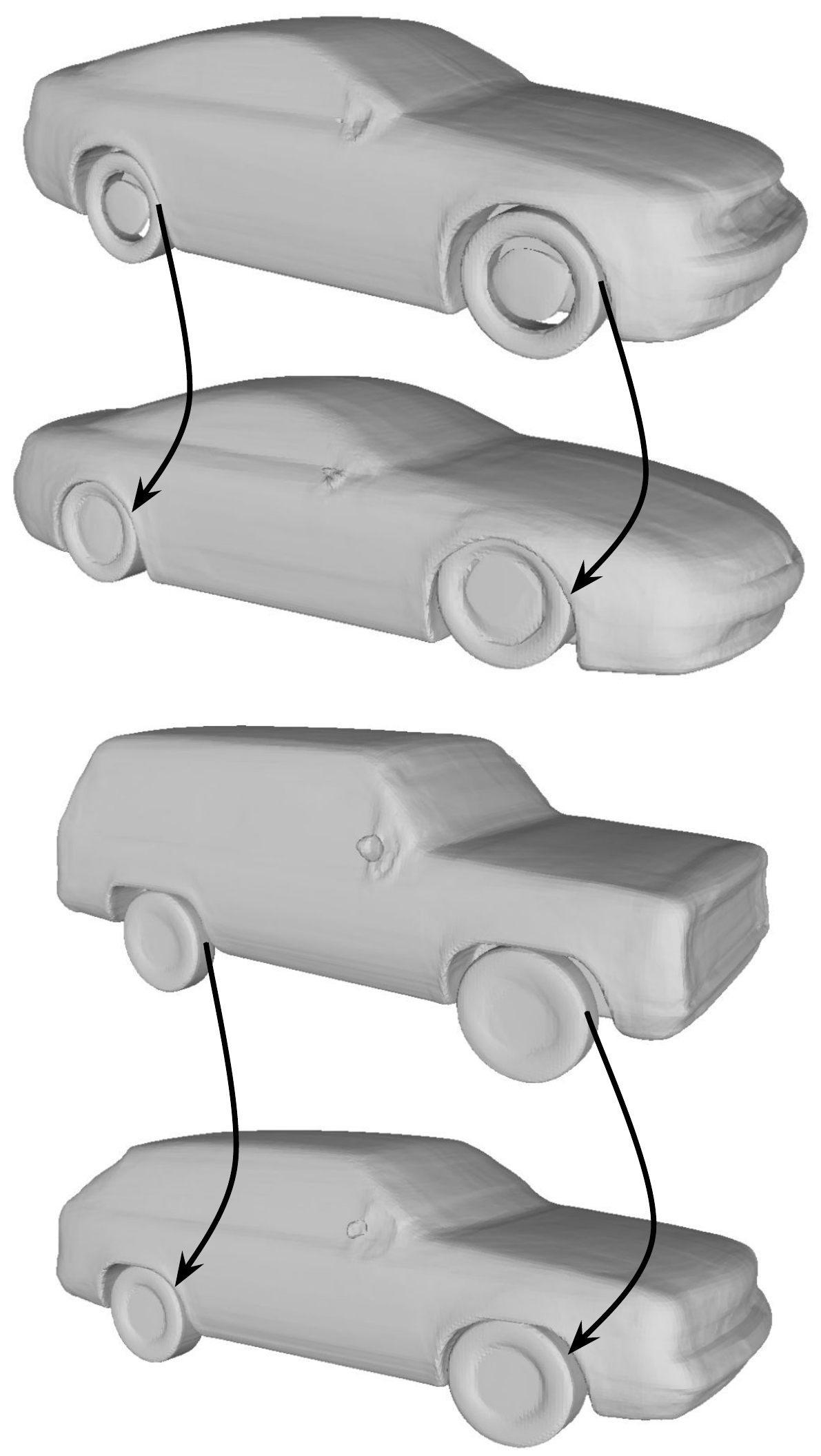} \intercellmanipsupp\\
	(a) \intercellmanipsupp&\intercellmanipsupp (b) \intercellmanipsupp& \multicolumn{2}{c}{(c)} & \multicolumn{2}{c}{(d)} \\
	\multicolumn{2}{c|}{\HS{}} & \multicolumn{4}{c}{\HSs{}} \\
\end{tabular}
\caption{\textit{Additional shape manipulation results.} In each example, we show a source object and the one obtained after manipulation, using \HS{} (a-b) and \HSs{} (c-d). (a-b) See Fig.~5 of the main text for explanation. (c) We edit the radius and height of the ring on the left, which causes proportionate changes in the rest of the parameters and the helix. (d) The parameters of the wheels are edited from the source car at the top, and the body adapts accordingly at the bottom.}
\label{fig:manip_all_supp}
\end{figure*}

\subsection{Single-image reconstruction}

Single-view reconstructions using the shared latent model \HSs{} are shown in Fig.~\ref{fig:image2car_supp}, while additional sketch reconstruction results with \HS{} are displayed in Fig.~\ref{fig:sketch2mixer_supp}.


\newlength{\imagetocarfigwidthsupp}
\setlength{\imagetocarfigwidthsupp}{0.3\columnwidth}

\setlength\mytabcolsepsupp{\tabcolsep}
\setlength\tabcolsep{2pt}

\begin{figure}[t]
	\centering
	\small  
	\begin{tabular}{ccc}
		\includegraphics[width=\imagetocarfigwidthsupp]{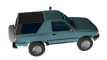} &
		\includegraphics[width=\imagetocarfigwidthsupp]{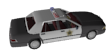} &
		\includegraphics[width=\imagetocarfigwidthsupp]{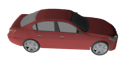} \\
		\includegraphics[width=\imagetocarfigwidthsupp]{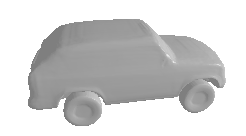} &
		\includegraphics[width=\imagetocarfigwidthsupp]{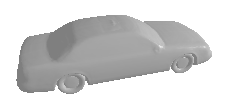} &
		\includegraphics[width=\imagetocarfigwidthsupp]{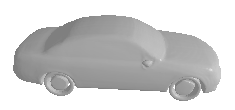} \\
		$r_\text{ours}=0.164$ & $r_\text{ours}=0.119$ & $r_\text{ours}=0.139$ \\		
		$r_\text{true}= 0.163$ & $r_\text{true}=0.113$ & $r_\text{true}=0.141$ \\
	\end{tabular}
	\caption{{\it Single view reconstruction using \HSs{}.} (\textit{Top row}) Input images. (\textit{Middle row}) Reconstructions using \HSs{}. (\textit{Bottom row}) The estimated wheel radius $r_\text{ours}$ and the ground truth $r_\text{true}$, as measured on the ground truth meshes.}
	\label{fig:image2car_supp}
\end{figure}
\setlength{\tabcolsep}{\mytabcolsepsupp}

\newlength{\sketchtomixerfigwidthsupp}
\setlength{\sketchtomixerfigwidthsupp}{0.13\columnwidth}
\newcommand{\intracellsupp}{\hspace{0mm}}
\newcommand{\intracellsuppbis}{\hspace{-2mm}}

\setlength\mytabcolsepsupp{\tabcolsep}
\setlength\tabcolsep{2pt}

\begin{figure}[t]
	\centering
	\small  
	\begin{tabular}{ccccc}
		\includegraphics[width=\sketchtomixerfigwidthsupp]{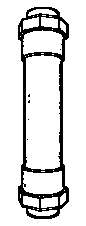} & \intracellsupp
		\includegraphics[width=\sketchtomixerfigwidthsupp]{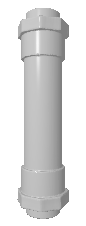} & \intracellsupp
		\includegraphics[width=\sketchtomixerfigwidthsupp]{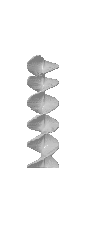} & \intracellsuppbis
		\includegraphics[width=\sketchtomixerfigwidthsupp]{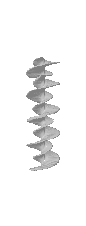} & \intracellsuppbis
		\includegraphics[width=\sketchtomixerfigwidthsupp]{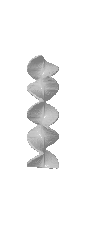} 
		\includegraphics[width=\sketchtomixerfigwidthsupp]{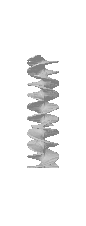} 
		\\
		
		\includegraphics[width=\sketchtomixerfigwidthsupp]{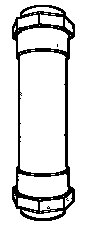} & \intracellsupp
		\includegraphics[width=\sketchtomixerfigwidthsupp]{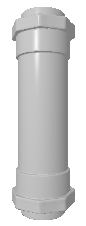} & \intracellsupp
		\includegraphics[width=\sketchtomixerfigwidthsupp]{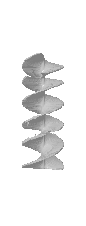} & \intracellsuppbis
		\includegraphics[width=\sketchtomixerfigwidthsupp]{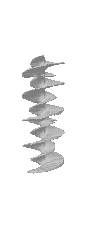} & \intracellsuppbis
		\includegraphics[width=\sketchtomixerfigwidthsupp]{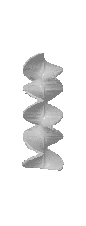} 
		\includegraphics[width=\sketchtomixerfigwidthsupp]{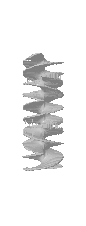} 
		\\
		
	\end{tabular}
	\caption{{{\it Additional reconstruction of static mixers from sketches.} (\textit{Left column}) Input sketches. (\textit{Second column}) Reconstructions using \HS{}.
	(\textit{Remaining columns}) Helices reconstructed by directly combining the tube parameters obtained from the sketch with four different latent vectors $\LV_{\text{generic}}$}.}
	\label{fig:sketch2mixer_supp}
\end{figure}
\setlength{\tabcolsep}{\mytabcolsepsupp}

\subsection{Chairs manipulation}
We report here an additional experiment on 4 legged chairs, using a subset of ShapeNet~\cite{Chang15} that considers those with four straight legs. We represent each leg as a geometric primitive, a cuboid, and the rest of the chair, the body, as a generic one, as shown in Fig.~\ref{fig:chair_supp}(a). Similarly to the cars, the part labels are taken from the noisy mesh annotations of~\cite{Kalogerakis17}. We end up with 649 chairs, 118 of which have part labels.
We report qualitative examples of parametric manipulation, as well as editing the latent vector of the generic primitive, in Fig.~\ref{fig:chair_supp}(b,c).


\newlength{\chairheight}
\setlength{\chairheight}{4.5cm}
\renewcommand{\intercellmanipsupp}{\hspace{0.mm}}

\begin{figure*}[t]
	\centering
	\small  
	\begin{tabular}{c|c|c}\intercellmanipsupp
		\includegraphics[height=\chairheight, trim={0 -4cm 0 -4cm}]{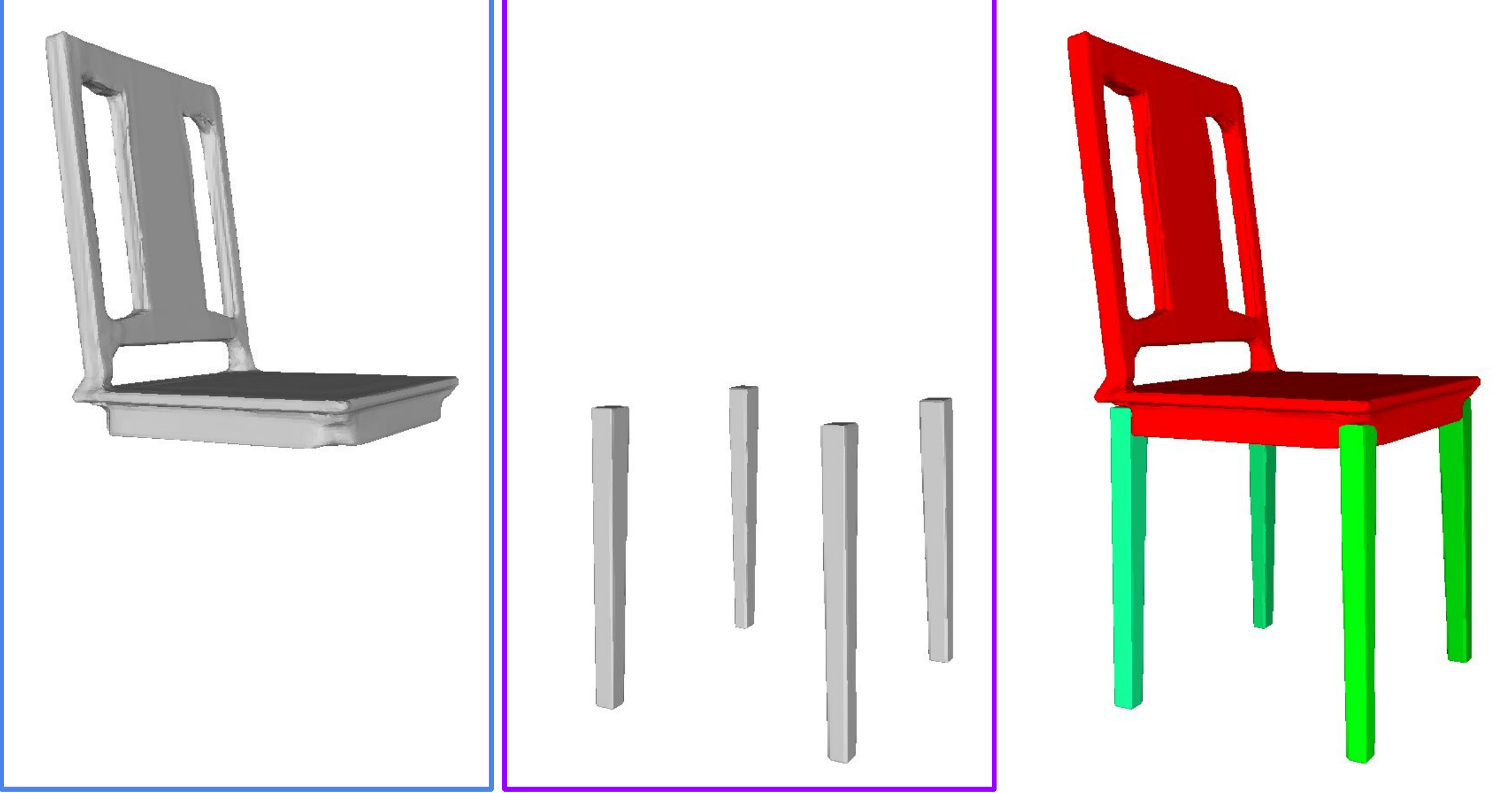} \intercellmanipsupp&\intercellmanipsupp
		\includegraphics[height=\chairheight]{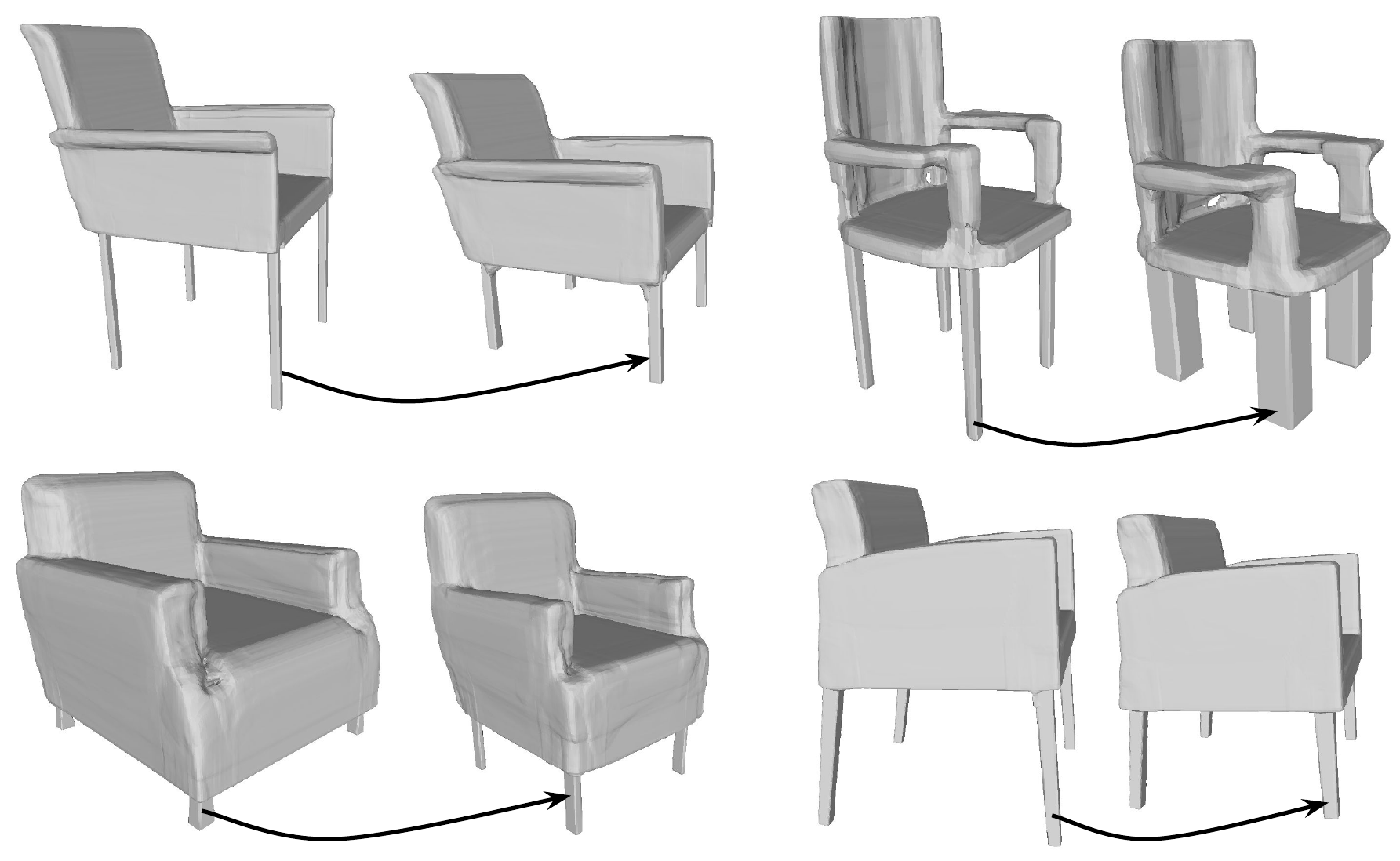} \intercellmanipsupp&\intercellmanipsupp
		\includegraphics[height=\chairheight]{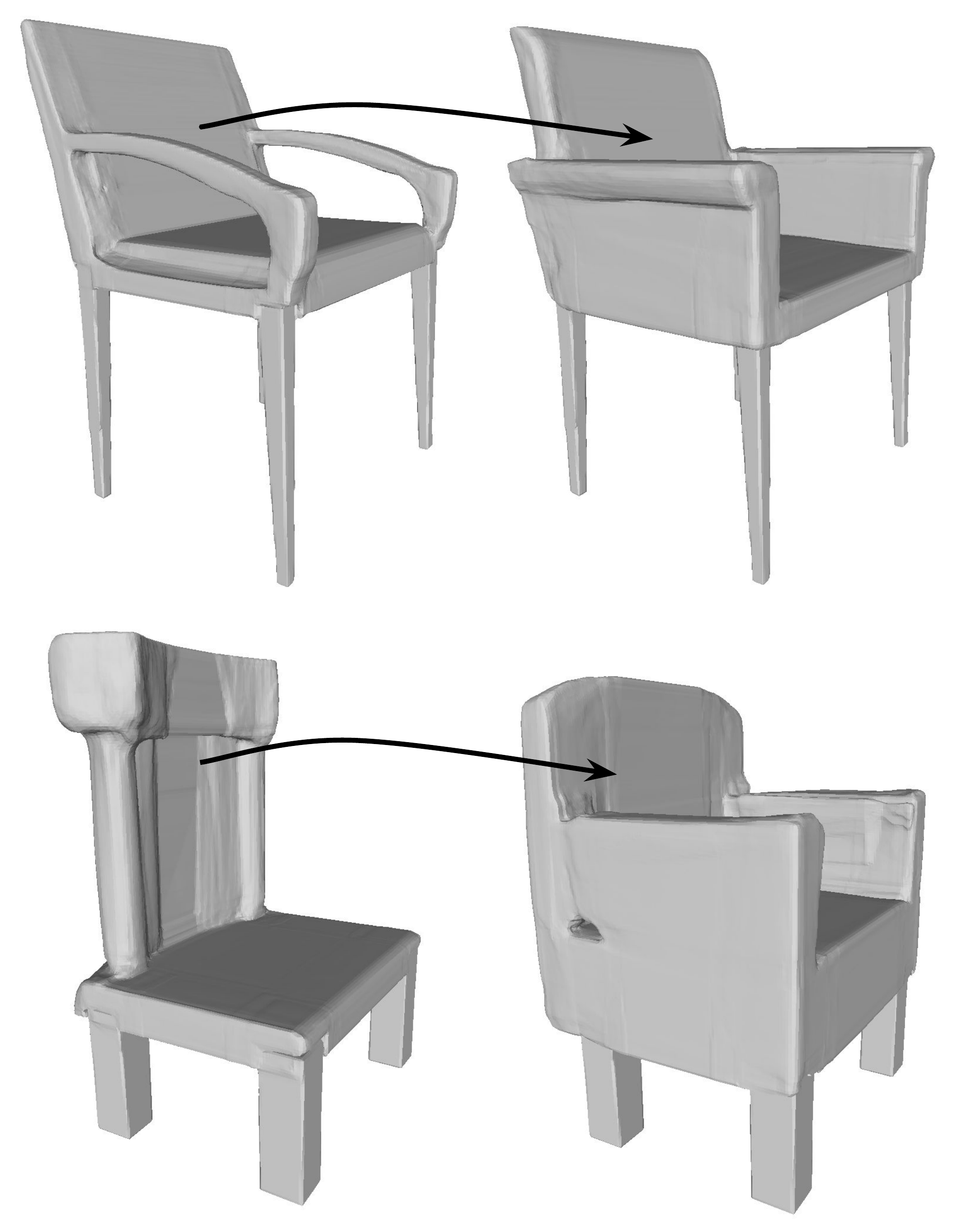} \\
	(a) \intercellmanipsupp&\intercellmanipsupp (b) \intercellmanipsupp&\intercellmanipsupp (c) \\
	\end{tabular}
	\caption{\textit{Results on Chairs using \HS{}.} (a) We depict chairs as the combination of the main body (sitting area, back and armrest) and four straight legs. The body is represented using a \textit{Generic-SDF} (red), while the legs are modeled by \textit{Geometric-SDFs} (green) of cuboids. (b) Parametric manipulation of the legs, the rest of the chair adapts accordingly. (c) The latent $\LV_{\text{generic}}$ is edited with constant leg parameters to obtain a new chair that fits the same legs.}
	\label{fig:chair_supp}
\end{figure*}

\subsection{\HSs{} for drag optimization}

\begin{figure}[t]
  \centering
  \small  
  \begin{tabular}{c}
  \includegraphics[width=0.7\columnwidth]{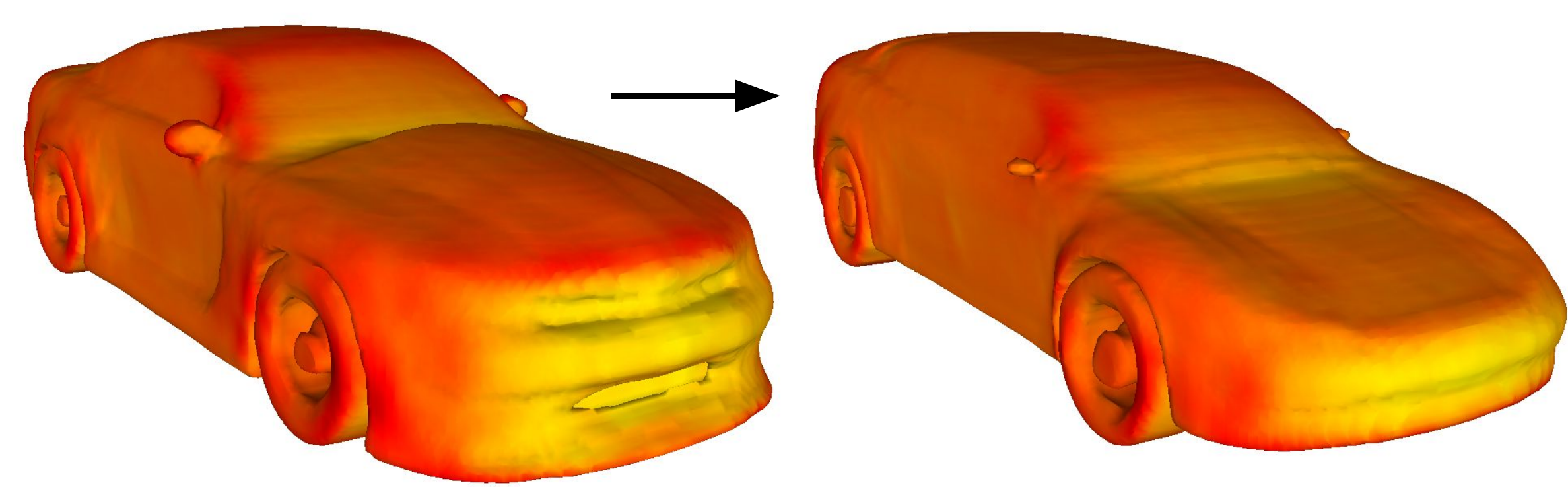}\\
  \includegraphics[width=0.7\columnwidth]{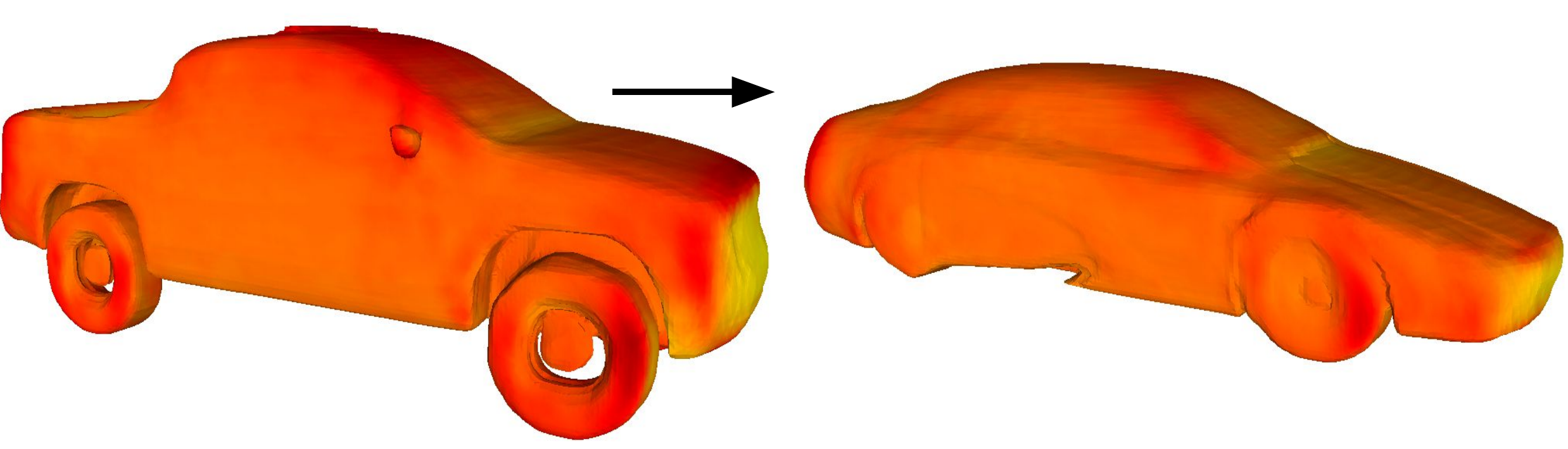}\\
  \DeepS{} \\ \\
  \includegraphics[width=0.7\columnwidth]{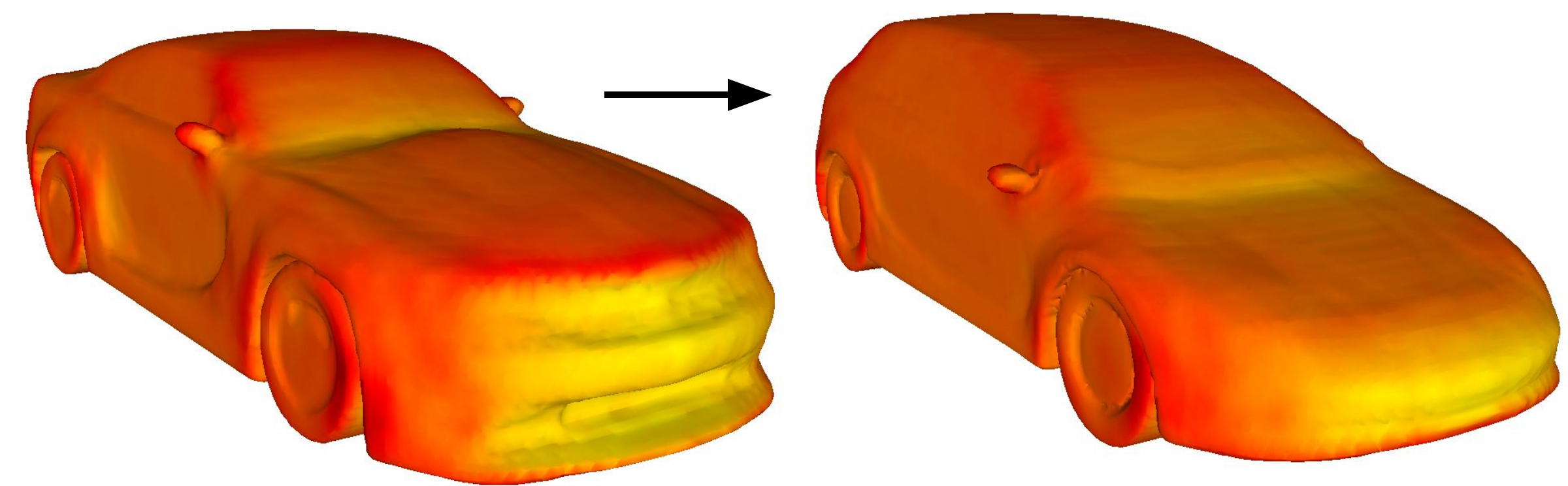}\\
  \includegraphics[width=0.7\columnwidth]{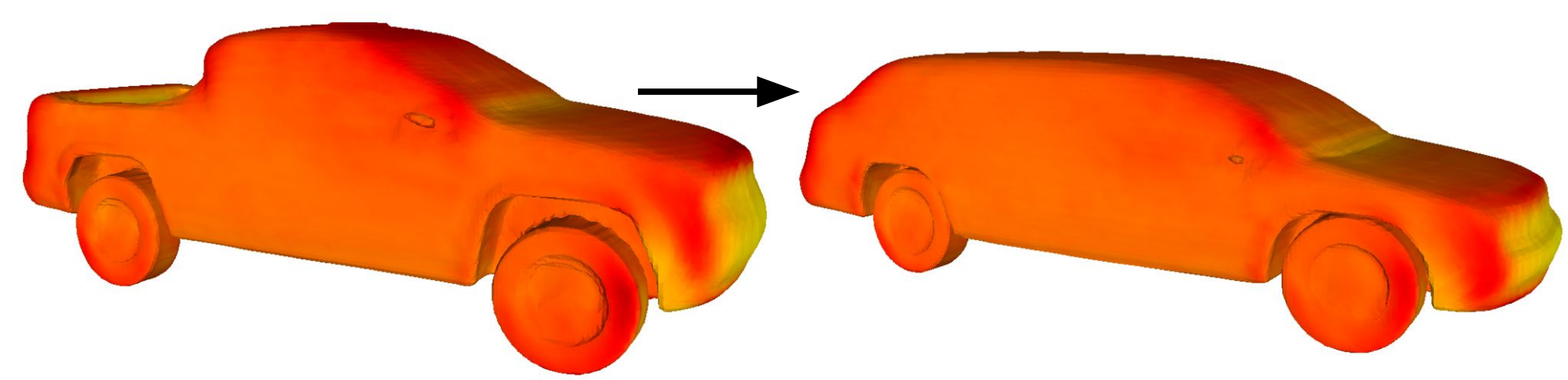}\\
  \HSs{} \\
  \end{tabular}
  \caption{\textit{Drag minimization.} We optimize two shapes using \DeepS{} and \HSs{} to minimize the drag they incur. In all cases, the initial shape is on the left and the drag-optimized one on the right. They are color coded from yellow for high pressure to red for low pressure. The top results are relatively similar between the methods. For the bottom ones, \HSs{} preserves the separation between wheel and body, whereas \DeepS{} does not and creates an undesirable artifact near the back of the optimized car.}
  \label{fig:drag_opt}
\end{figure}

The tight coupling between parts in a shared latent space also lends itself well to automated shape optimization as proposed in~\cite{Baque18, Remelli20b}. Fig.~\ref{fig:drag_opt} depicts two cars optimized for drag reduction in this manner. The resulting shapes are those of plausible cars whose wheels are still circular and do not touch the car body because \HSs{} imposes these constraints, unlike \DeepS{}.



\section{\HS{} within a \DualS{} framework}
\label{appendix:dualhs}

\begin{figure*}[t]
  \centering
  \small  
\begin{tabular}{c}
   \includegraphics[width=0.9\textwidth]{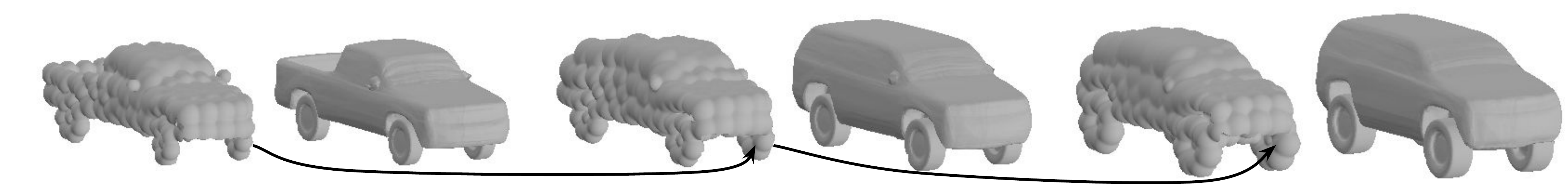} \\
    \includegraphics[width=0.9\textwidth]{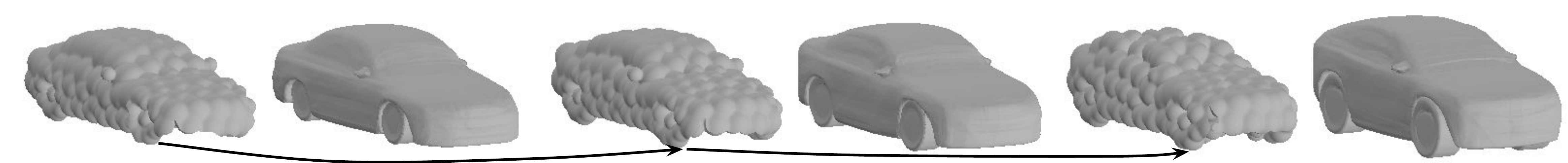} \\
    \includegraphics[width=0.45\textwidth]{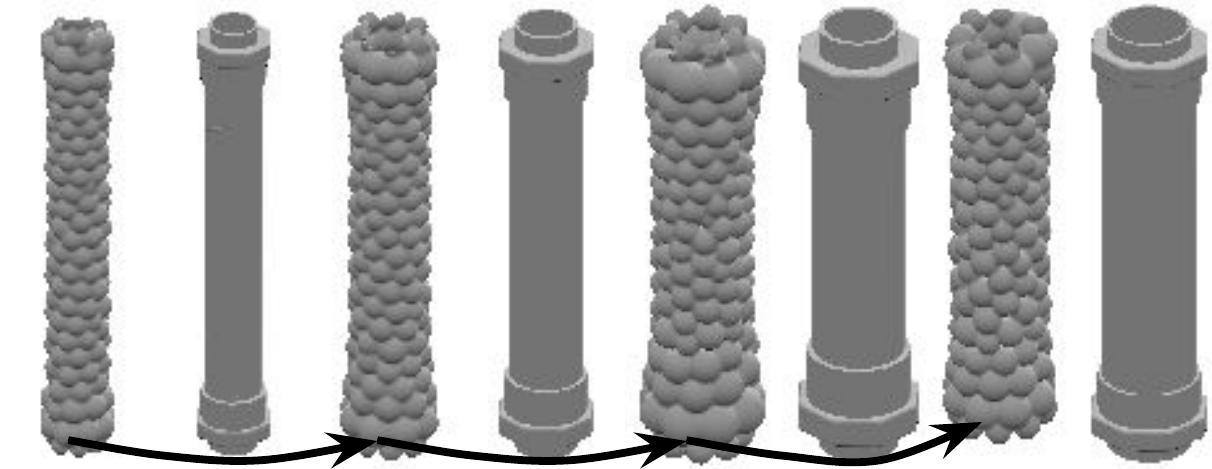} \\
  \end{tabular}
  \caption{{\it Shape manipulation using \DualS{}+\HSs{}.} {We progressively change the size of one sphere (connected by the arrows) and display the resulting shapes obtained using \DualS+\HSs{}. Our approach preserves the plausibility of the car wheels and prevents local surface distortions of the mixers.}}
  \label{fig:dualsdf_man}
  \end{figure*}

The parametric edition of our approach is not opposed to \DualS{} manipulation, but rather they can be complementary. Indeed, our approach can be integrated into the \DualS{} framework by replacing its high-fidelity representation by \HS{}, resulting in a new approach that we dub \DualS{}+\HS{}. For simplicity, we use the shared representation \HSs{} proposed above.
\DualS{} interactive manipulation involves changing the parameters of one of the spheres in the coarse level representation and then adapting the latent code accordingly. We can adapt \HSs{} shared latent code in the exact same way. As show in Fig.~\ref{fig:dualsdf_man}, this gives us the same real-time editing capability as in \DualS{}, in addition to our parametric manipulation, with the added benefit that we retain the physical plausibility of the car wheels or circumventing the local surface distortions of mixers.


\end{document}